\documentclass[%
 reprint,
 superscriptaddress,
 onecolumn,
nofootinbib,
 amsmath,amssymb,
 aps,
rmp,
longbibliography,
]{revtex4-2}

\usepackage{amsmath,amssymb, amsthm, mathtools}
\usepackage{url}
\usepackage{bm}
\usepackage{braket}
\usepackage{xcolor}
\definecolor{crimson}{RGB}{165,28,48}
\usepackage[colorlinks=true,allcolors=crimson]{hyperref}
\usepackage{microtype}

\usepackage{subfigure}

\usepackage{tikz}
\usetikzlibrary{calc} 

\theoremstyle{definition}
\newtheorem{example}{Example}

\let\sectionsymbol\S
\let\caron\v

\def\W{\bm W}
\def\w{\bm w}
\def\x{\bm x}

\def\M{\bm M}

\def\G{\bm G}
\def\K{\bm K}
\def\Sh{\Sh}

\def\v{\bm v}
\def\y{\bm y}

\def\A{\bm A}
\def\B{\bm B}
\def\O{\bm O}
\def\F{\bm F}
\def\P{\bm P}
\def\C{\bm C}
\def\X{\bm X}
\def\Z{\bm Z}

\def\S{\bm \Sigma}

\def\R{\bm R}
\def\df{\mathrm{df}}
\def\tf{\mathrm{tf}}
\def\Sh{\hat {\bm \Sigma}}
\def\e{\bm \epsilon}
\def\Sig{\bm \Sigma}

\def\min{\mathrm{min}}

\DeclareMathOperator*{\argmin}{arg\,min}
\DeclareMathOperator{\Tr}{Tr}
\DeclareMathOperator{\tr}{tr}

\usepackage{autonum}

\makeatletter
\let\cat@comma@active\@empty
\makeatother

\begin{document}

\title[Scaling and renormalization in high-dimensional regression]{Scaling and renormalization in high-dimensional regression}

\author{Alexander Atanasov}
\email{atanasov@g.harvard.edu}
\affiliation{Department of Physics, Harvard University, Cambridge, MA}
\affiliation{Center for Brain Science, Harvard University, Cambridge, MA}

\author{Jacob A. Zavatone-Veth}
\email{jzavatoneveth@fas.harvard.edu}
\affiliation{Department of Physics, Harvard University, Cambridge, MA}
\affiliation{Center for Brain Science, Harvard University, Cambridge, MA}
\affiliation{John A. Paulson School of Engineering and Applied Sciences, Harvard University, Cambridge, MA}
\affiliation{Society of Fellows, Harvard University, Cambridge, MA}

\author{Cengiz Pehlevan}
\email{cpehlevan@seas.harvard.edu}
\affiliation{Center for Brain Science, Harvard University, Cambridge, MA}
\affiliation{John A. Paulson School of Engineering and Applied Sciences, Harvard University, Cambridge, MA}
\affiliation{Kempner Institute for the Study of Natural and Artificial Intelligence, Harvard University, Cambridge, MA}

\date{\today}%

\begin{abstract}
From benign overfitting in overparameterized models to rich power-law scalings in performance, simple ridge regression displays surprising behaviors sometimes thought to be limited to deep neural networks. This balance of phenomenological richness with analytical tractability makes ridge regression the model system of choice in high-dimensional machine learning. In this paper, we present a unifying perspective on recent results on ridge regression using the basic tools of random matrix theory and free probability, aimed at readers with backgrounds in physics and deep learning. We highlight the fact that statistical fluctuations in empirical covariance matrices can be absorbed into a renormalization of the ridge parameter. This `deterministic equivalence' allows us to obtain analytic formulas for the training and generalization errors in a few lines of algebra by leveraging the properties of the $S$-transform of free probability. From these precise asymptotics, we can easily identify sources of power-law scaling in model performance. In all models, the $S$-transform corresponds to the train-test generalization gap, and yields an analogue of the generalized-cross-validation estimator. Using these techniques, we derive fine-grained bias-variance decompositions for a very general class of random feature models with structured covariates. This allows us to discover a scaling regime for random feature models where the variance due to the features limits performance in the overparameterized setting. We also demonstrate how anisotropic weight structure in random feature models can limit performance and lead to nontrivial exponents for finite-width corrections in the overparameterized setting. Our results extend and provide a unifying perspective on earlier models of neural scaling laws.
\end{abstract}

\maketitle

\tableofcontents

\clearpage

\section{Introduction}

The remarkable successes of deep learning confront us with many puzzles \cite{belkin2019reconciling,zhang2021understanding,kaplan2020scaling}. In particular, the study of ``neural scaling laws'' in deep learning has drawn wide attention. As dataset sizes and compute capabilities have increased, remarkably regular power law trends have been observed in the performance of large language, vision, and multimodal models \cite{hestness2017deep, kaplan2020scaling, bachmann2024scaling}. The exponents of these power laws determine how dataset and model size should be jointly scaled in order to achieve optimal performance for a given compute budget \cite{hoffmann2022training}. As a result, these scaling laws play an important role in modern deep learning practice, and serve to drive the state of the art performance across a variety of models. Therefore, understanding what determines these exponents is a key question for which one might hope to develop basic theoretical insights. 

Since the 1970s, statistical physicists have played a prominent role in the quest to understand learning in neural networks \cite{watkin1993rule,engel2001statistical}. The observation of scaling laws in deep learning is particularly interesting from the perspective of statistical physics, where the identification of scaling exponents as principal quantities of study led to major breakthroughs in the field  \cite{widom1965equation, kadanoff1966scaling, kadanoff1967static}. Especially key was the development of renormalization as a central method for the study of scaling properties in complex systems \cite{wilson1971renormalization, wilson1971renormalization2}.\footnote{See \citet{wilson1974renormalization} for an early review and \citet{cardy1996scaling} for an introduction.} Though we by no means intend to draw a clear historical analogy, it is important to emphasize the crucial role that the study of analytically tractable model systems played in the development of the general theory.

One can therefore ask whether there is simple setting of an information processing system where such power law behavior in performance as a function of dataset size and model size can be studied analytically. Recent papers, often using mathematical methods from statistical physics, have shown that high dimensional least squares regression from various feature spaces is one such example. These settings include linear regression \cite{krogh1992generalization, dicker2016minimax,dobriban2018prediction, nakkiran2019more, advani2020high, hastie2022surprises}, kernel regression \cite{sollich1998learning, sollich2002learning, bordelon2020spectrum, canatar2021spectral, spigler2020asymptotic, simon2023eigenlearning, loureiro2021learning}, and random feature models \cite{hastie2022surprises, louart2018random, mei2022generalization, adlam2020neural, d2020double, d2020triple, loureiro2021learning, bahri2021explaining, zavatone2023learning,dhifallah2020precise, hu2022universality, maloney2022solvable, bach2024high}. For these models, sharp asymptotic characterizations of training and generalization performance can be derived in limits where the feature space dimension and number of training data points jointly tend to infinity.

Here, we pursue an alternative approach to deriving these sharp asymptotics, based in random matrix theory and specifically making use of the $S$-transform of free probability \cite{voiculescu1992free}. This approach makes explicit the central role played by the randomness of sample covariance matrices.
Through this lens, a variety of phenomena including sample-wise and model-wise double descent \cite{belkin2019reconciling, nakkiran2021deep, nakkiran2019more, d2020triple}, scaling and bottleneck behavior \cite{bahri2021explaining, atanasov2022onset}, and the analysis of sources of variance for trained networks \cite{adlam2020understanding, d2020double}, can be seen as natural consequences of a basic renormalization phenomenon. This approach also yields a simple interpretation of the self-consistent equations that determine the generalization error across a wide variety of solvable models. 

We highlight how one can derive these phenomena across a variety of settings from a set of three basic principles:
\begin{enumerate}
    
    \item \textbf{Gaussian Universality}
    \nopagebreak \\
    When the number of dimensions in a ridge regression problem scales linearly with the number of data points, the training and generalization error are asymptotically identical to the error obtained by replacing the true data with Gaussian data of matched mean and covariance. This phenomenon is also referred to as Gaussian equivalence \cite{hastie2022surprises,montanari2022universality,hu2022universality,misiakiewicz2024non}.     
   
    \item \textbf{Deterministic Equivalence} 

    When calculating average case training and generalization error, one must average over the random choice of finite training set. In particular, this will involve averaging over the empirical covariance matrix of the sample of data. In recent years, several authors have shown how one can replace the (data-dependent, random) sample covariance with the (deterministic) population covariance within relevant algebraic expressions \cite{potters2020first,bun2016rotation}. Such a replacement is known as a \textbf{deterministic equivalence}. This allows one to easily perform the necessary averages and precisely characterize average case training and generalization error.

    \item \textbf{The $S$-transform}

    The $S$-transform allows one to characterize the spectral properties of a product of two matrices \cite{potters2020first}. An empirical covariance can be viewed as a multiplicative noise applied to the ``ground truth'' population covariance. In our settings, this noise is usually due to either a finite choice of training set or a finite set of random features that the data is passed through. The $S$-transform then gives us the method to replace expressions involving the empirical covariance with the deterministic equivalent involving only the population covariance. When this replacement is made, the ridge is rescaled (or more properly \textbf{renormalized}) to a new value. The renormalized ridge is given directly by multiplying the original ridge by the $S$-transform of the noise.

\end{enumerate}
These first two principles have been highlighted by several important recent works, which we review. Our primary focus is on the third point. By making use of basic properties of the $S$-transform, one can recover results previously obtained using replica, cavity, or linear pencil derivations in a few lines of algebra. The appearance of the $S$-transform also highlights that multiplicative noise on the covariance is at the heart of all overfitting and scaling phenomena in linear models.

\subsection{Review of Neural Scaling Laws} \label{sec:nsl}

In this section, we will review the phenomenology of neural scaling laws as well as the solvable models that seek to explain how data and task structure determine scaling behavior. We focus initially on observations of scaling laws in large language models, as those observations have substantially motivated recent theoretical interest. These initial observations focused on models using the Transformer network architecture \cite{vaswani2017attention}, which underpins modern language models like OpenAI's GPT series \cite{radford2018improving,radford2019language,achiam2023gpt} or DeepSeek's R1 \cite{deepseekai2025}. For a very recent review of the scaling laws literature in the context of language models, see \citet{anwar2024foundational}.  

To fix notation, let $\mathcal L(N, T)$ be the performance of a model with $N$ parameters trained on $T$ sample datapoints (usually referred to as ``tokens'' in the language modeling context).\footnote{We will often take $N$ to be the hidden layer width of the random feature models we study. Here it denotes the number of parameters. In deep networks trained end-to-end these quantities do not coincide, but in random feature models they are equal; see Section \ref{sec:LRF_motivation} for details.} We will be interested in characterizing the scaling properties of $\mathcal L$ as either $N$ or $T$ increase. For either of the parameters, its scaling law will vary depending on whether it is the bottlenecking parameter or not.

The existence of power-law scalings in language model performance with model and dataset size was highlighted in early empirical work \cite{hestness2017deep, rosenfeld2019constructive} (see also \citet{ahmad1988scaling} for extremely early work). \citet{kaplan2020scaling} performed an extensive empirical study of scaling laws in language modeling tasks and proposed the following scaling \textit{Ansatz} for $\mathcal L$:
\begin{equation}
    \mathcal L(N, T) = \left[ \left(\frac{N_c}{N} \right)^{\alpha_N / \alpha_T} + \frac{T_c}{T} \right]^{\alpha_T}.
\end{equation}
Here $N_c, T_c$ are constants and $\alpha_N, \alpha_T$ are scaling exponents, all of which must be fit to data. As $T \to \infty$ at fixed $N$ we see a scaling law going as $N^{-\alpha_N}$. Similarly as $N \to \infty$ at fixed $T$ we get a scaling law going as $T^{-\alpha_T}$. For trained Transformer language models, experimental estimates of both $\alpha_N$ and $\alpha_T$ are rather small, of order less than $0.1$.

More recently, \citet{hoffmann2022training} have proposed alternative scaling \emph{Ans\"atze} that can serve as better fits to data. This accounts for the fact that the entropy of text is nonzero and so the cross-entropy loss between natural and model-generated text should not vanish even in the $N, T \to \infty$ limit. They write:
\begin{equation}
    \mathcal L(N, T) = E + N^{-\alpha_N} + T^{-\alpha_T},
\end{equation}
where $E$ corresponds to the entropy of natural text. Again, as $N \to \infty$ (resp $T \to \infty$) this loss has power law scaling with the other parameter. \citet{besiroglu2024chinchilla} have performed a detailed replication attempt of the results of \citet{hoffmann2022training}, finding different estimates for the scaling exponents. 

These observations regarding scaling laws for language models have been refined and extended by a host of papers over the past few years \citep{ghorbani2021scaling, hernandez2021scaling, hernandez2022scaling, gordon2021data, muennighoff2024scaling,anwar2024foundational}. Moreover, scaling laws for non-language tasks \citep{zhai2022scaling, alabdulmohsin2024getting} and non-Transformer architectures \citep{bachmann2024scaling} have been investigated in other works.\footnote{More general parametric fits of the occasionally ``broken'' power law behavior observed in practice have been investigated in \citet{caballero2022broken}.}

Many attempts to build solvable models for how scaling laws arise in neural network training and generalization focus on learned functions that are linear in the set of trainable weights.\footnote{There are additional ways of thinking about models of scaling laws that don't fall into the framework of linear models, including \citet{hutter2021learning, sharma2022scaling, michaud2024quantization,arora2023theory}.} This means $f(\x) = \w \cdot {\bm \phi}(\x)$ for some $N$-dimensional vector of features $\bm \phi(\x)$, with $N$ possibly infinite.  The features themselves may also be random. Such models are called \textbf{linear models} and include kernel methods and random feature models. When the weights are learned via ridge regression on a fixed dataset of $P$ examples, one can compute the exact asymptotic behavior for the generalization performance of the model. The crucial simplification which enables precise asymptotic study of these linear models is Gaussian universality, which has been studied both for kernel methods with deterministic kernels \cite{dietrich1999statistical, bordelon2020spectrum,canatar2021spectral,mei2022hypercontractivity,xiao2022precise,misiakiewicz2022spectrum,hu2022sharp,dubova2023universality,loureiro2021learning,cui2021generalization,spigler2020asymptotic} and for random feature models \cite{pennington2017nonlinear,hu2022universality,dandi2023universality,pesce2023limits,montanari2022universality,mei2022generalization,mei2022hypercontractivity,loureiro2021learning,louart2018random,schroder2023deterministic,schroder2024asymptotics,adlam2020neural,adlam2020understanding,d2020double,d2020triple,hastie2022surprises}. One can adapt these methods to study the dynamics of high-dimensional linear models trained with stochastic gradient descent (SGD) \cite{ali2019earlystopping,paquette2021sgdlarge,paquette2022exactsgd,bordelon2021learning,lee2022momentum,bordelon2024dynamical}.

One motivation for the study of such linear models is that neural networks in the \textbf{neural tangent kernel} (NTK) parameterization converge to kernel methods in the infinite width limit \cite{jacot2020kernel, lee2019wide}.\footnote{See \citet{misiakiewicz2023six} for a recent review of NTKs and linearized networks.} Kernel methods have a long history, as their convex objective function has allowed for a tractable theory to be developed, see \citet{scholkopf2002learning, williams2006gaussian} for accessible introductions. Even at finite width, networks can be parameterized so that they still behave as linear models by using the output rescaling introduced in \citet{chizat2019lazy}. This is called the \textbf{lazy limit} of neural network training. It is also known as the \textbf{linearized regime}, since the network's training dynamics match that of its linearization in parameter space \cite{liu2021linearity}. Finite-width lazy networks behave like random feature approximations to the infinite-width neural tangent kernel \cite{adlam2020neural, ghorbani2021linearized}. By developing a better perspective on the kernel regime, one hopes to inform the analysis of neural networks that learn features \cite{belkin2018understand, fort2020deep, atanasov2021neural}.

What determines the scaling exponent in linear models? Considering possible scaling laws in $N$ and $P$, \citet{bahri2021explaining} provide a useful distinction between the scaling of generalization error with respect to whichever of $N$ and $P$ acts as a bottleneck (\textit{i.e.}, the smaller of the two when they are well-separated), and the scaling with respect to the other, non-bottlenecking parameters. The former type of scaling they term \textbf{resolution-limited} and the latter type they term \textbf{variance-limited}. 

Bahri \textit{et al.} argue that variance-limiting scaling of the non-bottlenecking parameter leads to a trivial exponent of $1$ and a power-law decay to an asymptote determined by the bottleneck parameter. In the underparameterized case $P \gg N$, one can interpret the $1/P$ corrections as coming from the finite-dataset variance of the final predictor as in classical statistics \cite{cramer1999mathematical,fahrmeir1985glm}. In the overparameterized case $N \gg P$, one can interpret the $1/N$ corrections as coming from the finite-width variance in the neural tangent kernel, as observed in \citet{geiger2020scaling} and calculated in several recent works \cite{dyer2019asymptotics, roberts2022principles, atanasov2022onset, bordelon2023fluctuations, aitken2020asymptotics, zavatone2022contrasting, zv2022asymptotics}. We will refer to all power laws with exponent $1$ as trivial scaling.

The resolution-limited scalings are generally nontrivial, irrespective of whether the model is over- or under-parameterized \citep{kaplan2020scaling}. In linear models, these nontrivial exponents can be estimated using the \textbf{source-capacity formalism}, which stipulates particular power law decays for the feature covariance eigenspectrum (the capacity exponent) and the coefficients of the target vector in the covariance eigenbasis (the source exponent) \cite{caponnetto2007optimal,cui2021generalization,caponnetto2005fast}. Given source-capacity conditions on the data, one can calculate the resulting power-law exponent for the generalization error of kernel ridge regression \cite{caponnetto2007optimal,cui2021generalization,caponnetto2005fast,bahri2021explaining,bordelon2020spectrum, canatar2021spectral,spigler2020asymptotic}. We reproduce this analysis in Section \ref{sec:kernel_scaling}. 

It is important to stress that the resolution-limited and variance-limited scalings are \textit{not} different scaling regimes. In both the overparameterized and underparameterized setting, there will always be a bottlenecking parameter with resolution-limited scaling exponents and non-bottlenecking parameters with variance-limited scaling exponents. The resolution-limited scaling exponents will depend on additional details of the dataset and model. These details will determine which \textbf{scaling regime} the model is in. We characterize the different scaling regimes for linear and kernel regression using the source and capacity formalism of \citet{cui2021generalization} in Equation \eqref{eq:LR_Eg_scaling_full}. We extend the source-capacity analysis to linear random feature models in Equations \eqref{eq:LRF_all_scalings1} and \eqref{eq:LRF_all_scalings2}, expanding on results on single-layer linear random feature models by \citet{maloney2022solvable}. 

Even when the number of parameters is much greater than the number of data points, the effects of finite model size can limit the scaling of the test error as one increases the number of data points. In particular, variance in the predictor due to the randomness over initializations can limit the scaling exponent \cite{atanasov2022onset}. This worse scaling can manifest itself long before the number of data points is comparable to the number of parameters, or even before it is comparable to the width.\footnote{$P = \text{width}$ can also be viewed as a separate double descent peak \cite{adlam2020neural}.} Here, we will show this occurs across a variety of random feature models with and without feature noise, and corresponds to a \textit{variance-dominated} scaling regime.

\subsection{Overview and Contributions}

The goals of this paper are twofold. First, we aim to provide an accessible introduction to the relevant random matrix theory necessary to obtain the results of prior models of neural scaling laws, double descent, and random feature regression \cite{bahri2021explaining, wei2022more, bordelon2020spectrum, canatar2021spectral, maloney2022solvable, mei2018mean, adlam2020neural, advani2020high, atanasov2022onset, spigler2020asymptotic, simon2023eigenlearning, d2020double, d2020triple, mel2021anisotropic, mel2021theory, cui2021generalization, pillaud2018statistical, cui2023error, zavatone2022contrasting, zavatone2023learning, jacot2020implicit, mei2022generalization, louart2018random, hastie2022surprises, loureiro2021learning, adlam2020understanding}. By using the $S$-transform, the results across a wide variety of the literature can be obtained in a straightforward and parsimonious manner. 
Second, by applying these techniques, we provide novel characterizations of the scaling regimes and the sources of variance that drive them across a wide variety of random feature models. We emphasize that all of these results could be derived using alternative techniques. However, the formalism used here makes it particularly easy to derive results for many different linear models in a unified manner. 

In \sectionsymbol\ref{sec:rmt}, we give a brief introduction of the key ideas in random matrix theory necessary for the derivations that follow.  We motivate this by considering empirical covariance matrices. We highlight that one can view a given empirical covariance as a multiplicatively noised version of the ``true'' population covariance. We define the resolvent and the Stiltjes transform, and then introduce the $R$ and $S$-transforms of free probability and their relevant properties. Self-contained derivations of the key properties of the $R$- and $S$-transforms are given in Appendix \ref{sec:diagrams}. Moreover, for completeness, we explicitly calculate the $R$ and $S$ transforms for a variety of random matrix ensembles that will be useful for us in Appendix \ref{sec:S_examples}. By using the basic properties of these transforms, we are able to bootstrap their algebraic form without needing to directly compute any resolvents. \sectionsymbol\ref{sec:renormalization} details the connection between the random matrix theory results introduced in \sectionsymbol\ref{sec:rmt} and renormalization in physical theories.

In \sectionsymbol\ref{sec:LR}, we apply these results to study learning curves in linear and kernel ridge regression. We efficiently recover the exact asymptotics of training and generalization error computed in previous works \cite{hastie2022surprises, bordelon2020spectrum, canatar2021spectral, simon2023eigenlearning, loureiro2021learning,dobriban2018prediction}. We can understand the key parameter $\kappa$ (sometimes called the signal capture threshold) as a multiplicatively renormalized ridge parameter $\lambda$. The multiplicative constant is precisely given by the $S$-transform of the multiplicative noise. Through this, non-monotonicities in the generalization error can be interpreted as renormalization effects \cite{canatar2021spectral, mel2021theory}. We further note that the square of the $S$-transform gives the ratio between out-of-sample and in-sample errors. By estimating the $S$-transform using only training data, one can arrive at prior results on out-of-sample risk estimation \cite{golub1979generalized, jacot2020kernel, wei2022more} also known as generalized cross-validation. We then provide exact formulas for the bias-variance decomposition of linear and kernel regression, reproducing the results of \citet{canatar2021spectral}. Finally, we derive the resolution-limited scaling exponents in terms of the source and capacity exponents of the dataset \cite{bordelon2020spectrum, cui2021generalization, caponnetto2005fast, caponnetto2007optimal}. We highlight how label noise and nonzero ridge can lead to different scaling regimes for the resolution-limited exponents, as explored in \cite{cui2021generalization}.

Sections \ref{sec:lrf} and \ref{sec:NLRF} contain the main novel technical contributions.
In \sectionsymbol\ref{sec:lrf} we apply the $S$-transform to obtain the generalization error of a variety of linear random feature models. This is the simplest setting where both the dataset size and the model size appear jointly in the scaling properties of the model. We derive the training and generalization error for \textit{any class} of random features, as long as the features are relatively free of the empirical covariance. We apply this to recover many previously known formulas for generalization error for specific random feature models \cite{bach2024high,zavatone2022contrasting,zavatone2023learning,gerace2020generalisation,loureiro2021learning,maloney2022solvable}, and obtain novel generalization formulas for the case of orthogonal projections. We obtain novel formulas for the fine-grained bias variance decomposition in the case of structured input data. These decompositions yield an equivalence between infinite ensembles of linear random feature models and linear regression with rescaled ridge. Aspects of this have been explored in past works \cite{lejeune2020implicit,yao2021minipatch,patil2024asymptotically}. We also find that adding structure to the weights can affect the exponents of the finite-width corrections in the overparameterized regime, giving a nontrivial variance-limited scaling. Fast-decaying weight spectra can lead to variance over initializations even when the width is infinite.  We recover the target-averaged scaling laws discussed in \citet{bahri2021explaining, maloney2022solvable}, and extend them to settings where the target labels are more general. Using our fine-grained bias-variance decompositions, we find a new scaling regime where finite-width effects can substantially impact performance even in the overparameterized setting. The bias-variance decomposition further allows us to characterize all scaling regimes of linear random feature models. To our knowledge, a characterization of these scaling regimes has not been previously obtained. 

In \sectionsymbol\ref{sec:NLRF} we extend these results to the setting of a random feature model with additive feature noise. This arises in the study of nonlinear random feature models via Gaussian equivalence, as studied in \citet{pennington2017nonlinear,hu2022universality,dandi2023universality,pesce2023limits,montanari2022universality,mei2022generalization,mei2022hypercontractivity,loureiro2021learning,louart2018random,schroder2023deterministic,adlam2020neural,adlam2020understanding,d2020double,d2020triple}. There, the effect of nonlinearity can be treated as independent additive noise on the features. Models with additive noise have also been used to study the limiting effects of finite-width fluctuations of the empirical NTK in \citet{atanasov2022onset}. We recover results on nonlinear random feature models \cite{mei2022generalization, adlam2020neural, mel2021anisotropic}. The formulas simplify substantially, leading us to note a surprising connection to linear random feature models. We derive novel formulas for the bias-variance decomposition when the input covariates are anisotropic and apply this to provide a characterization of the scaling regimes in this setting as well.

\subsection{Code Availability}

The following public repository 
\begin{center}
\href{https://github.com/Pehlevan-Group/S_transform}{https://github.com/Pehlevan-Group/S\_transform}    
\end{center}
contains the code necessary to reproduce all figures in this paper. Readers interested in the numerics may wish to follow along with these interactive Python notebooks.

\clearpage

\section{Random Matrix Models of Empirical Covariance Matrices}\label{sec:rmt}

Here we give a relatively brief overview of the key concepts from random matrix theory necessary to understand the derivations that follow. A basic knowledge of probability and linear algebra is sufficient. For a modern introduction to random matrix theory aimed at a broad technical audience, we recommend the recent text of \citet{potters2020first}. 

\subsection{Motivation: Empirical Covariance Matrices}

In many fields involving the analysis of large-scale data, ranging from neuroscience to finance to signal processing, many useful statistical observations depend on the covariance matrix of a given dataset. Concretely, consider a dataset of $P$ observations $\{\x_\mu\}_{\mu = 1}^P$, which we will take to be independent and identically distributed (i.i.d.) throughout this paper. Each $\x_\mu \in \mathbb R^N$ consists of $N$ features $[\x_\mu]_{i=1}^N$ and is drawn from the distribution $p(\x)$. For simplicity, we will assume all features are mean zero. The Greek $\mu$  will label the data points while the Roman $i$ will label the features. 

Given this, the \textbf{design matrix} $\X \in \mathbb R^{P \times N}$ has $\x_\mu^{\top}$ in its $\mu$-th row. The \textbf{empirical covariance} (also called the \textbf{sample covariance}) of this dataset is given by
\begin{equation}
    \hat \S \equiv \frac{1}{P} \X^\top \X \in \mathbb R^{N \times N}.
\end{equation}
The matrix $\Sh$ is a \textbf{random matrix}; that is, a matrix whose entries are random variables. 

Defining the ground truth covariance of the data (also called the \textbf{population covariance}) as $\S \equiv \mathbb E_{\x \sim p(\x)} [\x \x^\top]$, we get that $\hat \S \to \S$ as $P \to \infty$ for fixed $N$. This is the regime of classical statistics (see, e.g. \citet{hastie2009elements} for an overview). In the modern regime of machine learning, however, one frequently encounters situations where $P, N$ are both large and of the same scale, or even where $N \gg P$. For example, in deep learning, the activations of a given layer can exist in a several thousand dimensional space, leading to a setting where $P \sim N$. In kernel regression, the space of features is often infinite-dimensional.

In this work, we will be most interested in problems where a target $y$, which is a function of $\x$ is to be predicted via linear or ridge regression. Given a training set of $\X \in \mathbb R^{P \times N}$ and corresponding set of labels $\{y_\mu\}_{\mu=1}^P$, we will consider finding weights that minimize the ridge-regularized least squares error:
\begin{equation}\label{eq:ridge_loss}
    \hat \w = \argmin_{\w} \frac{1}{P} \sum_{\mu=1}^P (\x_\mu^\top \w - y_\mu)^2  + \lambda \|\w\|^2.
\end{equation}
The solution to this regression problem is given by:
\begin{equation}
    \hat \w = (\Sh + \lambda \mathbf I)^{-1} \frac{1}{P}\X^\top \y.
\end{equation}
Both the empirical feature-label correlation $\frac 1P \X^\top \y$ and the empirical covariance $\Sh$ appear in this formula. The role of the empirical covariance will be especially important. Understanding the properties of $\Sh$ in this proportional limit is a rich topic of study that belongs in the field of \textbf{random matrix theory} (RMT). 

In what follows, we will give some examples of random matrices. When the $\x_\mu$ are all drawn from a high-dimensional Gaussian distribution, their empirical covariance will be distributed as a \textbf{Wishart} random matrix. Many aspects of these matrices can be easily characterized in the limit where $N, P \to \infty$ with fixed ratio $q = N/P$, known as the proportional high-dimensional limit. Here, $q$ is called the \textbf{overparameterization ratio}. Moreover, a wide variety of covariance matrices that do not come from Gaussian data will have covariances that effectively converge to Wishart matrices in the proportional limit. If one is only interested in properties involving the covariance, one can replace the dataset with a high dimensional Gaussian of matching covariance. This phenomenon is known as \textbf{Gaussian universality} or \textbf{Gaussian equivalence}.  

\subsection{Examples of Random Matrices}\label{sec:RMT_examples}

\begin{example}[White Wishart Matrices]
    In the case where $\x_\mu$ are all drawn i.i.d. from a Gaussian with population covariance $\S$ equal to the identity, $\x_\mu \sim \mathcal N(0, \mathbf I)$, the empirical covariance is said to be drawn from a \textbf{white Wishart} ensemble. In particular, it is an $N$-dimensional Wishart matrix with $P$ degrees of freedom and scale matrix $P^{-1} \mathbf{I}$. This is also known as an isotropic or unstructured Wishart matrix . 
\end{example}

\begin{example}[Structured Wishart Matrices and Multiplicative Noise]
    When $\x_\mu$ are drawn from a Gaussian with population covariance $\S \neq \mathbf I$, then $\S$ is called a structured covariance and $\Sh$ is called a \textbf{structured} Wishart. This is also known as the anisotropic or colored case. 

    Any such $\X$ can be written as $\tilde \X \sqrt{\S}$ where the entries of $\tilde \X$ are i.i.d. as $\mathcal N(0, 1)$ and $\sqrt{\S}$ is the principal square root of $\S$. Then, one can write the empirical covariance as $\Sh = \sqrt{\S} \W \sqrt{\S}$, where $\W = \frac{1}{P} \tilde \X^\top \tilde \X$ is distributed as a white Wishart. In this sense, Wishart matrices can be understood as noisy version of the population covariance $\S$, where the noise process is given by multiplication with a white Wishart.
\end{example}

\begin{example}[Wigner Matrices as Additive Noise]\label{eg:Wigner}
Consider the setting where we are given a symmetric matrix $\A$ (possibly a covariance) that has additive noise applied to each entry. This is usually given by taking $\A$ and adding a symmetric random matrix with Gaussian entries to it. Such additive noise is observed, for example, as a leading-order correction to the empirical covariance $\hat \S$ in $1/P$ at large $P$.  This is the regime of classical statistics, which deals with corrections to the empirical covariance due to large but finite $P$ when $N$ is held fixed. For Gaussian data, the central limit theorem implies that at large $P$ one can asymptotically approximate $\Sh = \S + \frac{1}{\sqrt{P}} \sqrt{\S} \Z  \sqrt{\S} + O(P^{-1})$ \cite{neudecker1990variance}. Here $\Z$ is an unstructured \textbf{Wigner matrix}. We show this at the end of Section \ref{sec:white_wish}.

An unstructured Wigner matrix can be generated as follows: Take $\bm X \in \mathbb R^{N \times N}$ to be a random matrix with i.i.d. Gaussian entries such that $[\X]_{ij} \sim \mathcal N(0, \frac{\sigma^2}{N})$. The symmetrized random matrix $\bm X^\top + \bm X$ is known as a \textbf{Wigner} random matrix. This construction has the property that because $\X$ is drawn from a rotationally symmetric distribution, so is $\X + \X^\top$. We will not deal with Wigner matrices very often, but they are the most well-known example of random matrices. The limiting $N \to \infty$ spectral density of a Wigner matrix is the famed \textbf{semicircle law}. 
\end{example}

\begin{example}[Random Projection]
Consider a random $N$-dimensional subspace\footnote{We get this subspace by starting with the subspace spanned by the first $N$ basis vectors and rotating it by a random orthogonal matrix $\O$, chosen with respect to Haar measure on the orthogonal group.} of $\mathbb R^D$. The projection operator $\P$ that takes each vector in $\mathbb R^{D}$ and maps it to its orthogonal projection in this $N$-dimensional subspace is symmetric and satisfies $\P^2 = \P$. It is also a random matrix with the property that its eigenvalues are either zero or one. 
\end{example}

\subsection{The Spectral Density and the Resolvent}

In what follows, we will consider only symmetric matrices $\A$. The eigenvalues are therefore real and the eigenvectors form an orthogonal basis by the  spectral theorem. It will be convenient to adopt the following shorthand for the normalized trace of an $N \times N$ matrix:
\begin{equation}
    \tr [\cdot] \equiv \frac{1}{N} \Tr [\cdot] . 
\end{equation}

We will be primarily interested in quantities related to the spectral structure of a given random matrix $\bm A \in \mathbb R^{N \times N}$ in the limit of $N \to \infty$. At finite $N$, the \textbf{spectral density} of a given random matrix $\A$ with eigenvalues $\{\lambda_i\}_{i=1}^N$ is given by:
\begin{equation}
    \rho_{\A}(\lambda) := \frac{1}{N} \sum_{i = 1}^N \delta(\lambda - \lambda_i).
\end{equation}
In the limit of $N \to \infty$, $\rho_{\A}$ tends to a limiting distribution, which can have both a continuous ``bulk'' and countably many isolated outliers depending on the ensemble from which $\bm A$ was drawn. 

Another quantity of interest is the \textbf{matrix resolvent}:
\begin{equation}
    \bm G_{\bm A}(z) = (z \mathbf I - \bm A)^{-1}.
\end{equation}
This object has the property that its poles correspond to the eigenvalues of $\A$, and the residues are the outer products of the corresponding eigenvectors. The normalized trace of this quantity---also known as the \textbf{Stiltjes Transform} of $\rho_{\A}$ or sometimes just the \textbf{resolvent} of $\A$---is directly related to the spectral density $\rho_{\A}$:
\begin{equation}
    g_{\bm A}(z) \equiv \tr \left[ (z \mathbf I - \bm A)^{-1} \right] = \frac{1}{N} \sum_{i=1}^N \frac{1}{z - \lambda_i} = \int \frac{\rho_{\A}(\lambda) d\lambda}{z - \lambda}.
\end{equation}
Expanding $g_{\A}(z)$ in a power series in $1/z$, one gets coefficients equal to the normalized traces $\tr[A^k]$. This means $g_{\A}(z)$ behaves like a moment generating function for the spectral distribution of $\A$.

From this resolvent, one can recover the spectral density using the \textbf{inverse Stiltjes transform}:
\begin{equation}\label{eq:inv_stiltjes}
    \rho_{\A}(\lambda) = \lim_{\epsilon \to 0^+} \frac{1}{\pi} \mathrm{Im} [g_{\A}(\lambda - i \epsilon)],
\end{equation}
where the notation implies that $\epsilon$ tends to $0$ from above.

Crucially, for all of the random matrices that we will study, the Stiltjes transform $g_{\A}(z)$ \textbf{concentrates} over $\A$ as $N \to \infty$. A quantity $\mathcal O_{\A}$ is said to concentrate if it becomes independent of the specific choice of $\A$ in the ensemble. That is, as $N \to \infty$, $\mathcal O_{\A}$ approaches a finite deterministic quantity.\footnote{Technically speaking, we only assume that $\mathcal{O}_{\A}$ converges in probability to a deterministic limit.} This means that for sufficiently large matrices, we can replace this quantity with its average value. A consequence of this concentration is that the spectral density itself concentrates. That is, the eigenspectrum of a very large random matrix drawn from a well-behaved (e.g. a Wigner or Wishart) ensemble will have an eigenvalue density that is essentially deterministic. For a precise characterization and proof of the conditions under which resolvents and their associated eigenspectra will concentrate, see \citet{tao2023topics} or \citet{potters2020first}. 

A second type of moment-generating function encountered is defined as:
\begin{equation}
    \bm T_{\bm A}(z) = \bm A (z \mathbf I - \bm A)^{-1}.
\end{equation}
Its corresponding normalized trace, sometimes called the $t$-transform, is given by
\begin{equation}
    t_{\bm A}(z) = \tr\left[\bm A (z \mathbf I -\bm A)^{-1} \right].
\end{equation}
The matrix identity $\mathbf{I} + \bm A (z \mathbf I - \bm A)^{-1} = z (z \mathbf{I} - \A)^{-1}$ relates the $t$-transform to the resolvent:
\begin{equation}\label{eq:tg_rel}
\begin{aligned}
   \bm T_{\A}(z) &= z \G_{\A}(z) - \mathbf I, \quad \G_{\A}(z) = \frac{1}{z} \left(\bm T_{\A}(z) + \mathbf I \right),\\
   t_{\bm A}(z) &= z g_{\bm A}(z) - 1, \quad g_{\bm A}(z) = \frac{t_{\bm A}(z) + 1}{z}.
\end{aligned}
\end{equation}

\subsection{Degrees of Freedom}

Both $g_{\bm A}$ and $t_{\bm A}$ enter naturally in the calculations of training and generalization error that we will perform. In all such cases, however, they enter only after being evaluated at a negative value of $z$, e.g. $z = - \lambda$ for some $\lambda > 0$. As we will see in Section \ref{sec:LR}, this negative value is related to the ridge parameter of the regression. To simplify the final results in this paper, we therefore define the following auxiliary generating functions:
\begin{align}
    \mathrm{df}_{\bm A}^1(\lambda) &\equiv  \tr \left[\bm A (\bm A+\lambda \mathbf I)^{-1}\right] = -t_{\bm A}(-\lambda), \label{eq:dof1}\\
    \mathrm{df}_{\bm A}^2(\lambda) &\equiv  \tr \left[\bm A^2 (\bm A+\lambda \mathbf I)^{-2}\right] = \partial_{\lambda}(- \lambda t_{\bm A}(-\lambda)). \label{eq:dof2}  
\end{align}
These are the first and second \textbf{degrees of freedom} of the matrix $\A$. When $\A$ is understood from context, they will also be written as $\mathrm{df}_1$ and $\mathrm{df}_2$. The first of these appears prominently in statistics when defining the effective degrees of freedom of a linear estimator, see for example section 7.6 of \citet{hastie2009elements} and \citet{hastie2022surprises}. The notation has also been used extensively in a recent paper on linear random feature models by \citet{bach2024high}.

For some intuition about what $\df_1, \df_2$ measure, we will consider the concrete example of a high-dimensional Gaussian with covariance $\S \in \mathbb R^{N \times N}$. The eigenvalues $\eta_{k}$ of $\S$ will appear in the principal component analysis of this Gaussian. Frequently, one is interested in the \textit{effective dimensionality} of such an object. In order to calculate this, we define a scale of resolution $\lambda$. Eigenvalues greater than $\lambda$ will tend to be counted as increasing the dimensionality whereas eigenvalues smaller than $\lambda$ will tend to be be ignored. Rather than a sharp threshold at $\lambda$, we instead consider a softer such measure of dimensionality given by:
\begin{equation}
    \mathrm{dim}_1(\lambda) \equiv \sum_{k} \frac{\eta_k}{\lambda + \eta_k}.
\end{equation}
Here, if $\eta_k \gg \lambda$ then the term will contribute to the sum with a value close to $1$. On the other hand, if $\eta_k \ll \lambda$, then the term will enter the sum with a value close to zero, and not contribute substantially. A sharper but still analytic measure of dimensionality would involve raising each term to some power $p > 1$:
\begin{equation}
    \mathrm{dim}_p(\lambda) \equiv \sum_{k} \left(\frac{\eta_k}{\lambda + \eta_k}\right)^p.
\end{equation}
We see that $\df_1, \df_2$ correspond exactly to $\frac1N \mathrm{dim}_1$ and $\frac1N \dim_2$. These notions of dimensionality will appear naturally in the context of ridge regression. In fact, they are the only notions of dimensionality that turn out to matter in this context. Given that both $\df_1, \df_2$ are bounded to be between $0$ and $1$, one can also view them as the ``fraction of eigenvalues resolved'' at a given scale $\lambda$. 

Similarly, when there is a ``teacher'' vector $\bar \w$ that we want to weight the degrees of freedom by, we will define the following quantities by analogy to $\df_1, \df_2$:
\begin{align}
    \tf^1_{\A, \bar \w}(\lambda) &= \bar \w^\top \A (\A + \lambda \mathbf I)^{-1} \bar \w,\\
    \tf^2_{\A, \bar \w}(\lambda) &= \bar \w^\top \A^2 (\A + \lambda \mathbf I)^{-2} \bar \w.
\end{align}
When $\A, \bar \w$ are understood, we will similarly write these as just $\tf_1$ and $\tf_2$. In the case where we average $\tf_1, \tf_2$ over an isotropic distribution of $\bar \w$ (\textit{i.e.}, such that $\mathbb{E}[\bar{\w}\bar{\w}^{\top}] = \mathbf{I}/d$), we recover $\df_1, \df_2$ respectively. These formulae are also related to quantities used in \citet{hastie2022surprises, mel2021anisotropic,zavatone2023learning, bach2024high}. 

The following identities will be particularly useful to us:
\begin{align}
   &\frac{d}{d\lambda} (\lambda \df_1) = \df_2, \label{eq:df1df2_rel1}\\
    \frac{d\, \df_1}{d \log \lambda}  &=  \lambda \frac{d\, \df_1}{d\lambda} = \df_2 - \df_1, \label{eq:df1df2_rel2}\\
     \frac{d \log \df_1}{d \log \lambda} &=\frac{\lambda}{\df_1} \frac{d\, \df_1}{d\lambda} =\frac{\df_2 - \df_1}{\df_1} \label{eq:df1df2_rel3}.
\end{align}
The $\tf$ functions satisfy the same relationships between themselves. 

Finally we have an upper bound on $\df_2$ by:
\begin{equation}\label{eq:df2_bound}
\begin{aligned}
    \df_2 = \df_1 - &\lambda \tr[\A (\A + \lambda \mathbf I)^{-2}] \leq  \df_1 - \frac{\lambda}{\| \A \|_{op}} \df_2  \\
    \Rightarrow \df_2 &\leq \frac{\df_1}{1+\lambda / \| \A \|_{op}}
\end{aligned}
\end{equation}
where $\| \A \|_{op}$ is the maximal eigenvalue of $\A$. 

\subsection{Addition and Multiplication of Random Matrices}

We now summarize the key random matrix theory results that we will use in this paper. These results have their origins in the theory of \textbf{free probability}, which is concerned with the study of non-commutative random variables that satisfy a technical condition known as \textbf{freedom}. This theory is extremely general and powerful, and there are many excellent introductory texts \cite{mingo2017free,nica2006lectures,potters2020first,voiculescu1997free}. 

However, we will only be concerned with the application of free probability theory to particular classes of large random matrices. For our purposes, it suffices to say that a pair of $N \times N$ random matrices $(\A,\B)$ are jointly (asymptotically) free as $N \to \infty$ if they are ``randomly rotated'' with respect to one another. That is, $(\A,\B)$ is equal in distribution to $(\A,\O \B \O^{\top})$ for any randomly-chosen rotation matrix $\O$.\footnote{Here by ``randomly chosen'' we mean uniformly distributed with respect to the Haar measure on the orthogonal group of $N \times N$ matrices $O(N)$.} For the interested reader, we give a general definition of freedom in Appendix \ref{sec:diagrams}. Moreover, we give self-contained proofs for the key random matrix theory results we will use in Appendices \ref{sec:diagrams} and \ref{sec:S_examples}.

\subsubsection{\texorpdfstring{$R$-transform}{R-transform}}

Consider two large $N$-dimensional random matrices $\A, \B$ whose spectra $\rho_{\A}(\lambda), \rho_{\B}(\lambda)$ are known. One may ask what can be said about the spectrum of the sum $\A + \B$. It turns out that under certain assumptions on $\A, \B$, this question can be answered straightforwardly using the \textbf{$R$-transform} of free probability theory \cite{voiculescu1992free}. We define the $R$-transform of a matrix $\A$ by
\begin{equation}
    g_{\bm A}(z) = \frac{1}{z - R_{\bm A}(g_{\bm A})}.
\end{equation}
Note that $R_{\bm A}$ depends explicitly on the resolvent $g_{\bm A}$, not on $z$. 

For free random matrices $\A$ and $\B$, the $R$-transform satisfies the remarkable property that it is additive: 
\begin{equation}\label{eq:R_prop}
    R_{\bm A + \bm B}(g) = R_{\bm A}(g) + R_{\bm B}(g).
\end{equation}
Thus, one can easily determine the $R$-transform of the sum, which in turn enables computation of the resolvent and then the limiting spectral density. 

\subsubsection{\texorpdfstring{$S$-transform}{S-transform}}
Just as one is interested in the eigenvalues of a sum of two random matrices $\A, \B \in \mathbb R^{N \times N}$, one is also frequently interested in the spectrum of their product. In general, if $\A$ and $\B$ are symmmetric, then $\A \B$ will not be symmetric. However both $\A \B$ and $\B \A$ will share the same nonzero eigenspectrum. Further, if we define the symmetrized or \textbf{free product} by
\begin{equation}
\bm A * \bm B := \bm A^{1/2} \bm B \bm A^{1/2},
\end{equation}
we see that $\A \B$, $\B \A$, $\A*\B$, and $\B * \A$  will all share the same non-zero spectrum. We use this symmetrized product to ensure $\bm A * \bm B$ remains symmetric. 

Just as for sums of matrices, assuming $\A$ and $\B$ are free of one another, there is another transform that allows one to calculate the spectral properties of their product given individual knowledge of the spectra of $\A$ and $\B$. This is the \textbf{$S$-transform} of free probability theory \cite{voiculescu1992free}, which is defined by the solution of the equation 
\begin{equation}
    t_{\bm A}(z) =\frac{1}{z S_{\bm A}(t_{\bm A}) - 1}.
\end{equation}
Equivalently, defining $\zeta_{\bm A}(t)$ as the functional inverse of $t_{\bm A}$ (satisfying $\zeta_{\bm A}(t_{\bm A}(z))= z$), we can write:
\begin{equation}
    S_{\bm A}(t) = \frac{t+1}{t \zeta_{\bm A}(t)}.
\end{equation}
The $S$-transform has the important property that when $\A$ and $\bm B$ are free of one another:
\begin{equation}\label{eq:S_prop}
    S_{\bm A * \bm B}(t) = S_{\bm A}(t) S_{\bm B}(t).
\end{equation}
This is the main result that we will utilize to derive many of the formulas that follow. Finally, because $\df_{\A}^1(\lambda) = - t_{\A}(-\lambda)$ we will also write $S_{\A}(t) = S_{\A}(-\df_1)$ in many of the applications of this equation.

\subsubsection{Subordination Relations and Strong Deterministic Equivalence}\label{sec:subordination}

The properties of the $R$- and $S$-transforms reviewed above allow one to determine the traced resolvents of sums or products of random matrices, and thus determine their limiting density of eigenvalues. This leaves open the question of whether one can get useful information about the limiting properties of \textit{eigenvectors} of sums or products of random matrices. The fact that this question can be systematically answered in the affirmative is one of the key developments of modern random matrix theory \cite{potters2020first}. 

The key concept underlying this advance is the idea of \textbf{strong deterministic equivalence}, which intuitively speaking states that certain random matrices can be replaced by deterministic matrices if one promises only to query them in sufficiently nice ways. More precisely, given a sequence of random $N \times N$ matrices $\A$ and deterministic $N \times N$ matrices $\B$, we say that $\B$ is a deterministic equivalent for $\A$ if $\tr(\A\M) / \tr(\B\M) \to 1$ in probability as $N \to \infty$ for any $N \times N$ test matrix $\M$ of bounded operator norm. In this case, we write $\A \simeq \B$. One could also strengthen this condition to $\tr(\A \M) \to \tr(\B\M)$ in probability, but following \citet{bach2024high} we prefer to work with ratios as it is convenient not to worry too much about overall normalization. Moreover, one can also allow $\B$ to be a random matrix, and prove deterministic equivalences that average out only some of the randomness in $\A$. This will be important for many of our derivations 

Using the concept of strong deterministic equivalence, one can extend the identities encountered above for the traced resolvents of sums and products of random matrices to their un-traced counterparts. This leads to the key equivalences
\begin{align}\label{eq:subordination1}
    \mathbb E_{\B} \bm G_{\A + \B}(z) &\simeq \G_{\A} (z- R_{\B}(g_{\A + \B}(z)))\\ \mathbb E_{\B} \bm T_{\A \B}(z) &\simeq \bm T_{\A} (z S_{\B}(t_{\A \B}(z))),\label{eq:subordination2}
\end{align}
where we take $\B$ to be free of $\A$. These are called \textbf{subordination relations} for the $R$ and $S$ transforms respectively. Note that after multiplying Equation \eqref{eq:subordination2} by $\A^{-1/2}$ and $\A^{1/2}$ on the left and right respectively and making use of the pushthrough identity, \eqref{eq:pushthrough}, we obtain its symmetrized analogue:
\begin{equation}
    \mathbb E_{\B} \bm T_{\A * \B}(z) \simeq \bm T_{\A} (z S_{\B}(t_{\A \B}(z))). \label{eq:subordination3}
\end{equation}
Here, $t_{\A*\B} = t_{\A \B}$ since the nonzero eigenvalues are the same for both matrices. Note on the right hand side there is no need to take an expectation over $\B$ because $R_{\B}, S_{\B}, g_{\A+\B}, t_{\A \B}$ all concentrate. 

If we take the trace of Equations \eqref{eq:subordination1} and \eqref{eq:subordination2} and use that $g_{\A+\B} = [z-R_{\A+\B}(g_{\A+\B}(z))]^{-1}$ and $t_{\A \B} = (z S_{\A \B} - 1)^{-1}$ we get:
\begin{align}
    R_{\A+\B}(g_{\A+\B}(z)) &= R_{\B}(g_{\A+\B}(z)) + R_{\A}(g_{\A}(z - R_{\B}(g_{\A}(z))))\\
    &= R_{\A}(g_{\A+\B}(z)) + R_{\B}(g_{\A+\B}(z)), \nonumber
    \\
    S_{\A \B}(t(z)) &= S_{\B}(t_{\A \B}(z)) S_{\A}( t_{\A}(z S_{\B}(t(z))) ) \\
    &= S_{\A}(t_{\A \B}(z)) S_{\B}(t_{\A \B}(z)). \nonumber
\end{align}
These are the familiar $R$ and $S$ transform properties. We thus see that Equations \eqref{eq:subordination1} and \eqref{eq:subordination2} are stronger forms of these two properties.

Viewing $\B$ as additive or multiplicative noise, one can directly interpret these subordination relations. Equation \eqref{eq:subordination1} states that the resolvent of an additively noised matrix is equal to the resolvent of the clean matrix with a shifted value of $z$. The shift is given by the $R$-transform. Equation \eqref{eq:subordination2} states that $\bm T$ of a multiplicatively noised matrix is equal to $\bm T$ of the clean matrix with a rescaled value of $z$. This rescaling is given by the $S$-transform. As we discuss in Section \ref{sec:renormalization}, these are in a precise sense \textit{renormalization} effects as encountered in statistical field theories. 

These subordination relations have been derived using a myriad of techniques in prior works. In Appendix \ref{sec:diagrams}, we give a self-contained diagrammatic derivation of these subordination relations for general orthogonally-invariant ensembles, which is to our knowledge novel. For a derivation using the replica trick and the Harish-Chandra-Itzhakson-Zuber integral, we direct the interested reader to Appendix B of the work of \citet{bun2016rotation}. \citet{burda2011multiplication} gave a different diagrammatic derivation based on viewing the random matrices as perturbative corrections to a Wigner matrix. For simpler derivations of strong $S$-transform subordination in the special case where one of the random matrices is Wishart, see \citet{bach2024high} or \citet{atanasov2024risk} for proofs using the cavity method and diagrams, respectively. Regardless of which proof one prefers, what is important is that the subordination relations can be broadly applied while treating the details of the derivation as a black box.

\subsubsection{\texorpdfstring{Summary of $R$- and $S$-transform identities}{Summary of R and S transform identities}}

There are a few identities that will be helpful for us in our derivations. Firstly, a trivial consequence of the additivity of $R$ is that
\begin{equation}\label{eq:Radd}
    R_{\A + J \mathbf I}(g) = J + R_{\A}(g).
\end{equation}
Further we can get a multiplicative identity for $R$ by noting that for a fixed constant $\alpha$
\begin{equation}\label{eq:R_scaling}
    g_{\alpha \A}(z) = \alpha^{-1} g_{\A}(z/\alpha) \Rightarrow z_{\alpha \A} (g) = \alpha z_{\A} (\alpha g) \Rightarrow R_{\alpha \A}(g) = \alpha R_{\A}(\alpha g).
\end{equation}
Here we have let $z_{\bm A}(g)$ be the funtional inverse of $g_{\bm A}(z)$. 

We can also get a multiplicative identity for $S$. Consider $t_{\alpha \A}(z)$. We see that
\begin{equation}\label{eq:S_scaling}
    t_{\alpha \A}(z) = t_{\A}(z/\alpha) \Rightarrow \zeta_{\alpha \A}(t) = \alpha \zeta_{\A}(t) \Rightarrow S_{\alpha \A}(t) =  \frac{t+1}{t \alpha \zeta_{\A} (t)} = \alpha^{-1} S_{\A}(t).
\end{equation}
One can relate $g_{\A}, t_{\A}, R_{\A}, S_{\A}$ in the following two equations:
\begin{align}
    g_{\bm A}(z) &= \frac{t_{\bm A}(z)+1}{z} = t_{\bm A}(z) S_{\bm A}(t_{\bm A}(z)),\\
    t_{\bm A}(z) &= z g_{\bm A} - 1 = g_{\bm A}(z) R_{\bm A}(g_{\bm A}(z)).
\end{align}
Combining the above two equations also gives a relationship between the $R$ and $S$ transforms:
\begin{align}
    S_{\A}(t) &= \frac{1}{R_{\A}(t S_{\bm A}(t))}\label{eq:SRinv},\\
    R_{\A}(g) &= \frac{1}{S_{\A}(g R_{\bm A}(g))}\label{eq:RSinv}.
\end{align}

\subsection{Application: Empirical Covariances}\label{sec:emp_covariance}

The $S$-transform is especially useful when studying empirical covariance matrices. When $\Sh$ is drawn from a structured Wishart we have seen that we can write it as the free product of $\S$ with a white Wishart:
\begin{equation}
    \Sh = \S^{1/2} \bm W \S^{1/2}.
\end{equation}
The $S$-transform relation then yields:
\begin{align}
    t_{\Sh}(z) = \frac{1}{z  S_{\Sh}(t_{\Sh}) - 1} &= \frac{1}{z S_{\bm W}( t_{\Sh}) S_{\S}( t_{\Sh}) - 1} = t_{\S}(z S_{\bm W}( t_{\Sh})).
\end{align}
Taking $\lambda := -z, \kappa := - z S_{\bm W}( t_{\Sh}(z))$ gives the key deterministic equivalence
\begin{equation}\label{eq:df_equiv}
    \boxed{ \mathrm{df}_{\Sh}^1 (\lambda) \simeq \mathrm{df}_{\S}^1 (\kappa), \quad \kappa = \lambda S_{\W}(-\df_1). }
\end{equation}
This equivalence implies that one can evaluate $S_{\W} = S_{\W}(-\df_1)$ using either $\df_1 = \df^1_{\Sh}(\lambda)$ or $\df_1 = \df^1_{\S}(\kappa)$.\footnote{Throughout this paper, we use the shorthand $\df_1$. Because of Equation \eqref{eq:df_equiv}, in the large $N, P$ limit that we work in, there is no confusion as to whether this is $\df_{\Sh}^1(\lambda)$ or $\df_{\S}^1(\kappa)$. Both of these quantities are asymptotically equal in this limit.} Because $\df_{\Sh}^1(\lambda)$ enters prominently in all generalization error formulas encountered in this paper, this equation will play a key role in the derivations that follow. 

This equation relates the degrees of freedom (as in equation \eqref{eq:dof1}) of the empirical covariance at a given ridge to the degrees of freedom of the true covariance with a \textbf{renormalized ridge} $\kappa$ (see Section \ref{sec:renormalization} for discussion of why this terminology is justified). Because $S_{\W}$ has a simple analytic form as derived in \ref{sec:white_wish}, one can write a self-consistent equation for $\kappa$, giving.
\begin{align}
    \kappa =  \lambda S_{\W}(-\df_1) = \frac{\lambda}{1 - \frac{N}{P} \df_1} . \label{eq:kappa_def}
\end{align}
Again, one can evaluate $\df_1$ either as $\df^1_{\Sh}(\lambda)$ or $\df^1_{\S}(\kappa)$.
The first way gives an estimate of $\kappa$ from the empirical data of $\Sh$ alone, while the second way yields an analytic self-consistent equation for $\kappa$ in terms of the true population covariance $\S$. As noted in the prelude, Equation \eqref{eq:df_equiv} extends to the strong deterministic equivalence
\begin{equation}\label{eq:strong_T}
\boxed{
    \Sh (\Sh + \lambda \mathbf I)^{-1} \simeq \S (\S + \kappa \mathbf I)^{-1}.
    }
\end{equation}

Using the relationship \eqref{eq:tg_rel} between the $t$-transform and the resolvent, \eqref{eq:df_equiv} and \eqref{eq:strong_T} extend to deterministic equivalences for the resolvents of Wishart matrices:
\begin{align}
    \tr (  (\Sh + \lambda \mathbf I)^{-1} )  &\simeq \frac{\kappa}{\lambda}  \tr (  (\S + \kappa \mathbf I)^{-1} ), \label{eq:df_equiv2}
    \\
    (\Sh + \lambda \mathbf I)^{-1} &\simeq \frac{\kappa}{\lambda} (\S + \kappa \mathbf I)^{-1}. \label{eq:strong_G}
\end{align}
Equations \eqref{eq:strong_T} and \eqref{eq:strong_G} are true when $\Sh$ is the free product of $\S$ with \textit{any} rotation-invariant multiplicative noise matrix $\M$, not just a white Wishart. In particular, writing $\Sh = \S^{1/2} \M \S^{1/2}$
\begin{equation}\label{eq:all_det_equiv}
    \Sh (\Sh + \lambda \mathbf I)^{-1} \simeq \S (\S + \kappa \mathbf I)^{-1}, \quad  (\Sh + \lambda \mathbf I)^{-1} \simeq \frac{\kappa}{\lambda} (\S + \kappa \mathbf I)^{-1}, \quad \kappa = \lambda S_{\M}.
\end{equation}

In all of the above equations, $\kappa$ can be interpreted in several ways:
\begin{enumerate}
    \item It is the original ridge $\lambda$, renormalized by $S_{\W}$ coming from the multiplicative noise of the high-dimensional covariance. Even in the ridgeless limit, $\kappa$ remains nonzero provided that $S_{\W}$ picks up a pole. We will study this in Sections \ref{sec:iso_lin_reg} and \ref{sec:double_desc}; see also \citet{hastie2022surprises,kobak2020optimal,wu2020optimal} for some early discussions of this effect. In fact, the poles of the $S$-transform will be in correspondence with the different ridgeless regimes of a given model, as we show in Section \ref{sec:LRF_ridgeless}. 
    \item It is the \textbf{signal capture threshold}, or equivalently the \textbf{resolution}. Eigenvalues larger than $\kappa$ will correspond to modes that are all learned, while eigenvalues smaller than $\kappa$ will not be learned. We will demonstrate this in Equation \eqref{eq:kernel_Eg} in Section \ref{sec:LR_derivation}.
\end{enumerate}
As $P$ gets larger, the fluctuations of the high dimensional covariance are suppressed and $S_{\W}$ becomes smaller. Consequently, $\kappa$ becomes smaller and the resolution improves. We will see in \ref{sec:kernel_scaling} that for covariances with power law structure, where the $k$th eigenvalue of $\S$ decays $k^{-\alpha}$ that the resolution improves as $\kappa \sim P^{-\alpha}$. $\alpha$ is called the \textbf{capacity} exponent of the data manifold \cite{caponnetto2005fast, caponnetto2007optimal,pillaud2018statistical,  cui2023error, steinwart2009optimal, cui2021generalization}. Large $\alpha$ implies most of the spread of the data is in the first few principal components, leading to effective low dimensionality. Smaller $\alpha$ imply the data is higher dimensional and thus the curse of dimensionality has a stronger effect. Consequently, the resolution $\kappa$ gets finer-grained at a slower rate in $P$. This is at the heart of all resolution-limited scalings.

\subsection{Why is this renormalization?}\label{sec:renormalization}

The use of the term \textit{renormalized} here is intentional, as this is an exact example of a renormalization phenomenon. For one, the diagrammatic picture as discussed in Appendix \ref{sec:diagrams} as well as \citet{maloney2022solvable, burda2011multiplication} mirrors the treatment of self-energy diagrams in renormalized perturbation theory. Here, because of the nature of the problem, the perturbative treatment is exact. 

The change from $\lambda$ to $\kappa$ is exactly due to $\kappa$ absorbing the contributions of the statistical fluctuations when we go from $\Sh$ to $\S$. This is analogous to how a renormalized mass term absorbs the quantum or thermal fluctuations in standard field theory. The $S$-transform exactly accounts for the multiplicative rescaling of $\lambda$ due to these fluctuations. In this setting the resolvents $\bm T$ and $\bm G$ play the roles of Green's functions. 

In the limit of $\lambda \to 0$, one finds that $\kappa$ can remain nonzero. This happens in overparameterized settings, as appear in Sections \ref{sec:LR}, \ref{sec:lrf}, \ref{sec:NLRF} and also in bottlenecked settings, as appear in Sections \ref{sec:lrf}, \ref{sec:NLRF}. Moreover, this nonzero $\kappa$ is precisely what causes models without explicit regularization to undergo double descent. $\kappa$ can be thought of as the implicit regularization that the model sees. In statistical and quantum field theory, a similar effect also occurs. There, a theory that is scale free (i.e. massless) at the classical level can pick up a scale (i.e. mass) after fluctuations are accounted for.  A commonly given example of this effect is in $\phi^4$ theory  \cite{peskin2018introduction, zinn2021quantum}. This is to say that double descent in unregularized ridge regression has the same underlying mechanism as the ``radiative mass generation'' in statistical and quantum field theory.

Finally, one might ask whether there is a notion of ``renormalization group flow'' in this setting, wherein only some fluctuations are integrated out while others remain \cite{peskin2018introduction, zinn2021quantum}. The deterministic equivalences that we have written down specifically integrate out all fluctuations in order to yield the deterministic quantities that are most useful in precisely characterizing asymptotic properties of the learned weights, and of train and test risks. More generally, denoting $\Sh_{P} \in \mathbb R^{N\times N}$ as an empirical covariance with $P$ datapoints, one has a set of equivalences 
\begin{equation}
    \Sh_P (\Sh_P + \lambda)^{-1} \simeq \Sh_{P'} (\Sh_{P'} + \lambda')^{-1} \simeq \S (\S + \kappa)^{-1}.
\end{equation}
Here $\lambda' = \lambda S_{\W_{N/P}} / S_{\W_{N/P'}}$ while $\kappa =  \lambda S_{\W_{N/P}}$, where $\W_{q}$ is a white Wishart matrix with overparameterization ratio $q$ and $S_{\W}$ is the corresponding $S$-transform. The population covariance $\S$ corresponds to $\Sh_{\infty}$. We thus see that varying $P$ gives a ``flow'' between covariances of different amount of data. Strictly speaking, we should take the joint limit $N, P \to \infty$ and view the overparameterization ratio $q$ as varying. After accounting for the renormalization of the ridge, this gives an equivalence between the corresponding Green's functions. In Appendix \ref{sec:white_wish}, we give a derivation of the $S$-transform of a Wishart matrix based on this idea of partially integrating out data. 

\clearpage 

\section{Linear and Kernel Ridge Regression}\label{sec:LR}

In this section, we will use the random matrix technology developed thus far to compute sharp asymptotics for the training and generalization error in linear ridge regression in the limit of dataset size $P$ and input dimension $N$ going to infinity jointly with fixed ratio, as in \citet{advani2020high, hastie2022surprises, dicker2016minimax,dobriban2018prediction, krogh1992generalization}. We will assume that the data is distributed according to a high-dimensional Gaussian. In the proportional limit, this assumption is not restrictive due to the phenomenon of Gaussian equivalence, which states that the generalization error for models with suitably-distributed non-Gaussian covariates will coincide with that of a Gaussian model with matched first and second moments. We will provide a more detailed discussion of Gaussian equivalence in Section \ref{sec:kernel}. We will further show how these results naturally give the formulae for the generalization error of kernel ridge regression as studied in \citet{bordelon2020spectrum, spigler2020asymptotic, canatar2021spectral}.

As a technical note: Although the formulas presented hold only in the limit of $N, P \to \infty$ with fixed ratio, we will keep $P$, $N$ explicit in this and subsequent sections. This notational choice is based on the fact that we will view all expressions as the leading order term in an asymptotic series in $1/P$ and $1/N$. The subleading finite $N, P$ contributions can in principle be calculated through finite $N, P$ corrections to the spectrum of the covariance together with adding crossing diagrams in the derivation of Appendix \ref{sec:diagrams}. The latter is given by the genus expansion in the full Weingarten formula \cite{weingarten1978asymptotic}. In this sense, the deterministic equivalence $\simeq$ will be taken to mean that these quantities are equal after neglecting the higher order terms in the series. In practice, we find excellent agreement from just the leading term. 

\subsection{Linear Regression with Structured Gaussian Covariates}

We begin by defining our statistical model for training data, along the way fixing notation that will be used throughout the paper. We consider $P$ data points $\x_\mu \in \mathbb R^N$, which we assume to be drawn i.i.d. from a $N$-dimensional Gaussian distribution with zero mean and covariance $\S$:
\begin{align}
    \x_\mu \underset{\mathrm{i.i.d.}}{\sim} \mathcal N(\bm{0}, \S).
\end{align}
We generate labels $y_\mu$ corresponding to each $\x_\mu$ by
\begin{align}
    y_\mu = \bar \w \cdot \x_\mu + \epsilon_\mu, 
\end{align}
where $\bar \w \in \mathbb{R}^{N}$ is the \textbf{signal} or \textbf{teacher weights} and $\epsilon_\mu$ is \textbf{label noise} which models variability in $y_{\mu}$ conditional on $\x_{\mu}$. Unless stated otherwise, we assume that $\bar{\w}$ is deterministic. We take the noise to be independent and Gaussian:
\begin{align}
    \epsilon_{\mu} \underset{\mathrm{i.i.d.}}{\sim} \mathcal{N}(0,\sigma_{\epsilon}^{2}) .
\end{align}
Collecting the covariates into a design matrix $\X \in \mathbb R^{P \times N}$ with $\X_{\mu i} = [\x_\mu]_i$, the labels into a vector $\y \in \mathbb{R}^{P}$, and the label noises into a vector $\e \in \mathbb{R}^{P}$, our statistical model can therefore be summarized as
\begin{align}
    \y = \X \bar{\w} + \e .
\end{align}
For brevity, we denote our data model by $\mathcal{D}$, and write $\mathbb{E}_{\mathcal{D}}[\cdot] = \mathbb{E}_{\X,\e}[\cdot]$.
We will take the eigenvalues of $\S$ and the norm of $\bar{\w}$ to be of order unity with respect to $N$. 

We will consider ridge regression with as in Equation \eqref{eq:ridge_loss}. The weights of the ridge regression estimator are then given by
\begin{equation}
\begin{aligned}
    \hat \w &= (\X^\top \X + P \lambda \mathbf I)^{-1} \X^\top \y\\
    \Rightarrow \bar {\bm w} - \hat \w &= P \lambda (\X^\top \X + P \lambda \mathbf I)^{-1} \bar {\bm w} - (\X^\top \X + P \lambda \mathbf I)^{-1} \X^\top \bm \epsilon\\
    &= \lambda (\Sh + \lambda \mathbf I)^{-1} \bar {\bm w} - \frac1P (\Sh + \lambda \mathbf I)^{-1} \X^\top \bm \epsilon.
\end{aligned}
\end{equation}
Here, we have taken $\Sh := \frac{1}{P} \X^\top \X \in \mathbb R^{N \times N}$ to be the empirical covariance obtained from sampling $P$ datapoints. As $P \to \infty$ we have $\Sh \to \S$ and $\mathbb E_{\Sh} \Sh = \S$. On a held out identically distributed test point $\x'$ (\textit{i.e.} $\mathbb E[\x' {\x'}^\top] = \S$)  we calculate the average generalization error: 
\begin{equation}\label{eq:LR_Eg_signal_noise}
\begin{aligned}
E_g &=\mathbb E_{\mathcal D, \x'} \|{\x'}^\top  \hat \w-  {\x'}^\top \bar {\bm w} \|^2\\
&= \mathbb E_{\Sh, \bm \epsilon} [(\bar {\bm w} - \hat \w)^\top \S  (\bar {\bm w} -  \hat \w) ]\\
&= \lambda^2 \mathbb E_{\Sh} [\bar{ \bm w }^\top (\Sh + \lambda \mathbf I)^{-1} \S (\Sh + \lambda \mathbf I)^{-1} \bar {\bm w} ] + \frac{\sigma_\epsilon^2}{P} \mathbb E_{\Sh} \mathrm{Tr}[\Sh (\Sh + \lambda \mathbf I)^{-1} \S (\Sh + \lambda \mathbf I)^{-1}]\\
&= \underbrace{-\lambda^2 \partial_{J} \bar \w^\top  \mathbb E_{\Sh} \left[ (\Sh + J \S + \lambda \mathbf I)^{-1} \right] \bar \w \big|_{J=0}}_{\text{Signal}} + \underbrace{\frac{\sigma_\epsilon^2}{P} \partial_\lambda \mathbb E_{\Sh} \left[ \lambda  \mathrm{Tr}\left[\S (\Sh + \lambda \mathbf I)^{-1}\right] \right]}_{\text{Noise}}.
\end{aligned}
\end{equation}
We get two terms. The first, which we call the \textit{signal} term, involves $\bar \w$ directly. The other, which we call the \textit{noise} term is proportional to $\sigma_\epsilon^2$ and independent of $\bar \w$. Both of these terms have been written in terms of matrix resolvents in the last line. We will now perform the average over the data in both of these terms using the methods developed in the prior section. 

To evaluate the noise term, we will simply need the deterministic equivalence stated in Equation \eqref{eq:strong_G}. For the signal term, we need the equation for the $S$-transform of a shifted Wishart matrix obtained in Section \ref{sec:shifted_wish} as well as the deterministic equivalence between resolvents for general noise structure given by Equation \eqref{eq:all_det_equiv}.

\subsection{Derivation}\label{sec:LR_derivation}

We evaluate the noise term first. There, using the deterministic equivalence \eqref{eq:strong_G} of the resolvent we have that
\begin{equation}
    \mathrm{Noise} \simeq \frac{\sigma_\epsilon^2}{P} \partial_\lambda \left[ \kappa \mathrm{Tr}\left[\S ( \S + \kappa \mathbf I)^{-1}\right] \right] = \sigma_\epsilon^2 \frac{d \kappa}{d \lambda} \frac{N}{P}  \partial_{\kappa} [\kappa \df_1(\kappa)] = \sigma_\epsilon^2 \frac{d \kappa}{d \lambda} \frac{N}{P} \df_2(\kappa)
\end{equation}
where we have used Equation \eqref{eq:df1df2_rel1} in the last equality to relate $\df_1$ to $\df_2$. Recalling that $t_{\A} = - \df^{1}_{\A}$ for any matrix $\A$ we can write $\kappa = S_{\W} \lambda $ as:
\begin{equation}\label{eq:kappa_lr_defn}
    \kappa = \frac{\lambda}{1 - \frac{N}{P} \df^1_{\S}(\kappa)}.
\end{equation}
Adopting the shorthand $\df_1 = \df^1_{\S}(\kappa)$, This lets us evaluate $\kappa$ and its derivative:
\begin{equation}
    \kappa (1 - \frac{N}{P} \df_1(\kappa)) = \lambda \Rightarrow \frac{d \lambda}{d \kappa} = 1 - \frac{N}{P} \df_2(\kappa).
\end{equation}
By defining the quantity
\begin{equation}
    \gamma \equiv \frac{N}{P} \mathrm{df}_2(\kappa) =  \frac{1}{P} \Tr[\S^2 (\S + \kappa \mathbf I)^{-2}]
\end{equation}
we get that 
\begin{equation}
    \mathrm{Noise} = \sigma_\epsilon^2 \frac{\gamma}{1-\gamma}.
\end{equation}

For the signal term, we need to calculate a deterministic equivalent for the resolvent $(\lambda + \Sh + J \S)^{-1}$. The trick is to realize that $\Sh + J \S$ can be written as the free product of $\S$ with a shifted white Wishart matrix. That is, $\Sh + J \S= \S^{1/2} (\W + J \mathbf I) \S^{1/2}$. Then, using Equation \eqref{eq:all_det_equiv}:
\begin{equation}
    (\Sh + J \S + \lambda)^{-1} \simeq \frac{\kappa_J}{\lambda} (\S + \kappa_J \mathbf I  )^{-1}, \quad \kappa_J = S_{\W + J \bm I} \lambda.
\end{equation}
The signal term then becomes:
\begin{equation} \label{eq:eval_signal1}
    \mathrm{Signal} \simeq -\lambda  \partial_J [\kappa_J \bar \w^\top (\S + \kappa_J \mathbf I)^{-1}  \bar \w ] \big|_{J=0} = - \lambda \frac{d \kappa_J}{d J} \Big|_{J=0} \bar \w^\top \S (\S + \kappa \mathbf I)^{-2} \bar \w.
\end{equation}
We have calculated the shifted Wishart $S$-transform $S_{\W + J \mathbf I}$ in Section \ref{sec:shifted_wish}. There, using Equation \eqref{eq:partial_shifted_wish}, we have at leading order in $J$ that
\begin{equation} \label{eq:eval_signal2}
\begin{aligned}
    \kappa_J \left(1 - \frac{N}{P} \df_1(\kappa_J) + J \frac{\kappa_J}{\lambda} \right) = \lambda \Rightarrow -\frac{d \kappa_J}{d J}\Big|_{J = 0} = \frac{\kappa^2/\lambda}{1-\gamma}.
\end{aligned}
\end{equation}
This gives the full generalization error:
\begin{equation}\label{eq:lr_eg}
\boxed{
    E_g \simeq -\frac{\kappa^2 \tf_{\S, \bar \w}'(\kappa)}{1-\gamma} + \sigma_\epsilon^2 \frac{\gamma}{1-\gamma}.
    }
\end{equation}
Letting $\eta_{i}$ be the eigenvalues of the covariance matrix $\S$, this can be written as:
\begin{equation}\label{eq:kernel_Eg}
    E_g \simeq \frac{\kappa^2}{1-\gamma} \sum_{k=1}^N \frac{\eta_k \bar w_k^2 }{(\kappa + \eta_k)^2}  + \sigma_\epsilon^2 \frac{\gamma}{1-\gamma}.
\end{equation}
This result recovers the sharp asymptotics for linear ridge regression obtained with various methods in prior works, including \cite{hastie2022surprises, bordelon2020spectrum, canatar2021spectral}. As noted in Section \ref{sec:emp_covariance}, modes with $\eta_k \gg \kappa$ are learned while modes with $\eta_k \ll \kappa$ are not yet learned.  This result has also recently found various applications in the context of neuroscience \cite{canatar2024spectral,bordelon2022population}.  

Equation \eqref{eq:kernel_Eg} is sometimes referred to as an \textbf{omniscient risk estimate}. This is because it requires exact knowledge of the spectrum of $\S$, the scale of $\sigma_\epsilon^2$, and the form of $\bar \w$ in order to calculate this. In statistical learning, it is strongly preferable to be able to build such an estimator out of the training data alone, without having to know all the details of the distribution of $\x$ and the data generating process for $y$.

As we will show in Section \ref{sec:KARE}, one can estimate the out-of-sample risk from \textit{only} the training error and $S$. Because of the key property that $S$ can be calculated solely in terms of the sample covariance and the original ``bare'' ridge $\lambda$, namely $S = (1 - q \df^1_{\hat \S} (\lambda))^{-1}$, we obtain a way to estimate the out-of-sample risk using in-sample data alone. This has been obtained in prior works \cite{craven1978smoothing, golub1979generalized, jacot2020kernel, wei2022more} under the name of \textbf{kernel alignment risk estimator} (KARE) or \textbf{generalized cross-validation} (GCV).

\subsection{Example: Isotropic Linear Regression}\label{sec:iso_lin_reg}

\begin{figure}[t]
    \centering
    \includegraphics[width=4in]{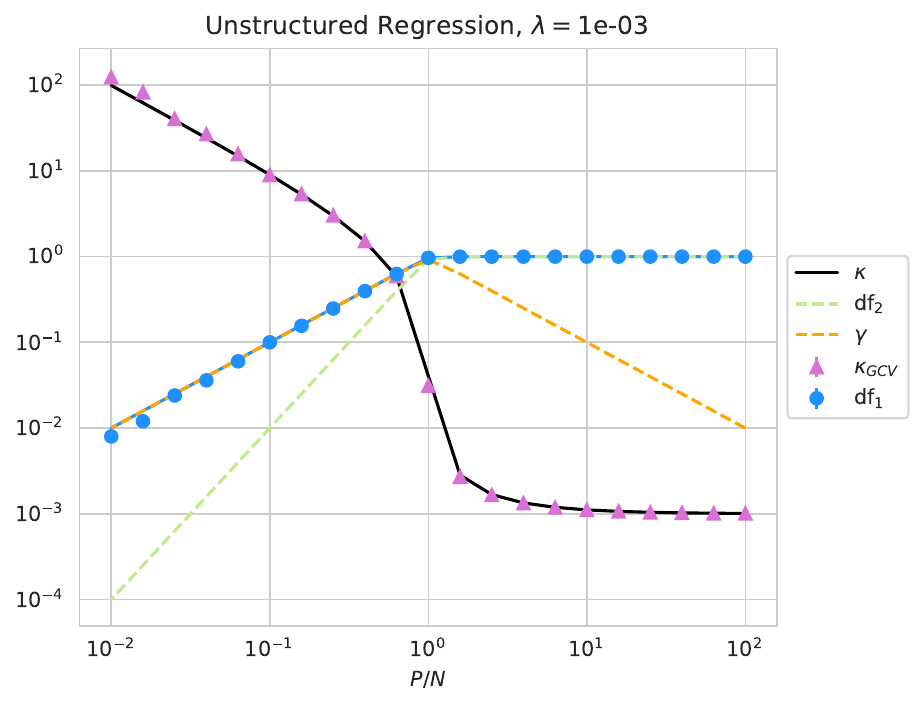}
    \includegraphics[width=4in]{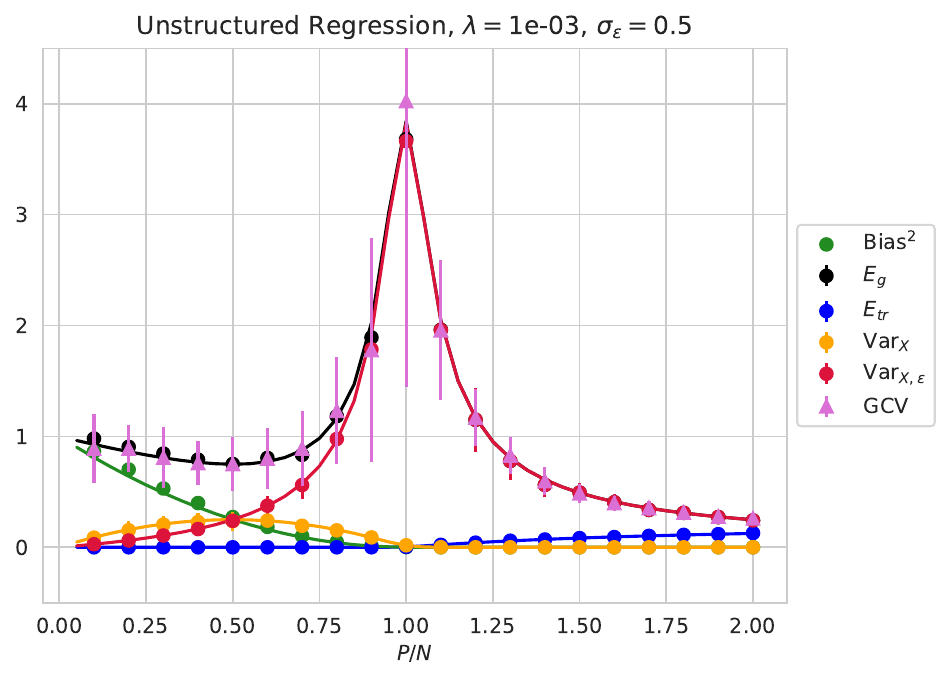}
    \caption{Linear regression on unstructured covariates, \textit{i.e.} $\S = \mathbf I$. Left: we plot theory (solid lines) for the various quantities of interest $\kappa, \gamma, \df_1, \df_2$. We also plot the empirical estimate of $\df_1$, namely $\df_{\Sh}(\lambda)$. Using this, we estimate of $\kappa_1$ using the training set and find excellent agreement. Right: We plot the training and generalization (blue, black respectively) as well as the bias (green) and variances (orange, red) due to the dataset and label noise. Dots and error bars indicate empirical simulations over 20 seeds over the training set. Solid curves show theory. We find excellent agreement for all relevant quantities. The GCV estimator is plotted as orchid triangles and again we find strong agreement with the generalization error. Here, $\lambda=10^{-3}$.}
    \label{fig:unstructured_LR_linspace}
\end{figure}

In the case where $\S = \mathbf I$, the formulas simplify. This setting has been studied in \citet{krogh1992generalization, advani2020high}. Here, $\df_{\S}^1(\kappa) = (1+\kappa)^{-1}$ and the self-consistent equation for the renormalized ridge $\kappa$ can be solved exactly:
\begin{equation}
    \kappa = \frac{\lambda}{1- \frac{N}{P} \frac{1}{1+\kappa}} \Rightarrow \kappa = \frac12 \left(\lambda+\tfrac{N}{P}-1 + \sqrt{(\lambda+\tfrac{N}{P}-1)^2 + 4 \tfrac{N}{P} \lambda} \right).
\end{equation}
The equations for the generalization of ridge regression can then be written down explicitly in terms of $\kappa$.
\begin{equation}
    E_g = \frac{1}{1-\gamma} \frac{\kappa^2}{(1+\kappa)^2} + \sigma_\epsilon^2 \frac{\gamma}{1-\gamma}, \quad \gamma = \frac{N}{P} \frac{1}{(1+\kappa)^2}.
\end{equation}
In the limit of $\lambda \to 0$ we get $\kappa = \max(0, \frac{N}{P}-1)$. Thus, in the underparameterized ridgeless limit where $P > N$, $\kappa =0$ and the ridge is not renormalized. However, in the overparameterized setting where $P < N$, even at zero ridge $\kappa$ has the finite value $\frac{N}{P} - 1$. Similarly we have $\gamma = \min(\frac{P}{N}, \frac{N}{P})$. Thus,
\begin{equation}
    E_g \simeq \begin{cases}
        \displaystyle\sigma_\epsilon^2 \frac{N/P}{1-N/P} & \text{underparameterized}\\
        \displaystyle\left(1 - \frac{P}{N} \right) + \sigma_\epsilon^2 \frac{P/N}{1-P/N} & \text{overparameterized}.
    \end{cases}
\end{equation}
We plot this in Figure \ref{fig:unstructured_LR_linspace}.

\subsection{Connection to Kernel Regression via Gaussian Universality}\label{sec:kernel}

So far, we have focused on linear regression directly from the space in which the covariates live. However, both in machine learning at large and in the specific setting of linearized neural networks as outlined in \sectionsymbol\ref{sec:nsl}, one is often interested in the case in which the covariates are transformed into some higher-dimensional feature space via a fixed mapping, \textit{i.e.}, in \textbf{kernel regression}. 

Concretely, consider a case in which we have $P$ datapoints $\x_{\mu} \in \mathbb{R}^{D}$ sampled i.i.d. from some probability measure $\rho(\x)$. Then, choose some kernel $K(\x,\x')$ with which to measure similarities. Then, under suitable conditions, the kernel has a Mercer decomposition
\begin{align}
    K(\bm x, \bm x') = \sum_{i=1}^{N} \eta_i \phi_i(\bm x) \phi_i(\bm x') 
\end{align}
with eigenvalues $\eta_{i} \geq 0$ and eigenfunctions $\phi_{i}$, which satisfy
\begin{equation}
    \begin{aligned}
    \int & \phi_i(\bm x) K(\bm x, \bm x') \phi_j(\bm x')\, d \rho(\bm x)\, d\rho(\bm x') = \S_{ij} = \delta_{ij} \eta_i,\\
    \mathbb E[\phi_i \phi_j] = \int  &\phi_i(\bm x) \phi_j(\bm x) \,d\rho( \bm x ) = \delta_{ij}, \quad \mathbb E[\phi_i] = \int \phi_i(\bm x) \,d\rho( \bm x) = 0. \label{eq:covar}
\end{aligned}
\end{equation}
We can write $K(\bm x, \bm x') = \sum_i \eta_i \phi_i(\bm x) \phi_i(\bm x') = \sum_i \psi_i(\bm x) \psi_i(\bm x')$ for features $\psi_i(\bm x) := \sqrt{\eta_i} \phi_i(\bm x)$. In this setting, we are performing linear regression from a feature space spanned by the functions $\psi_{i}$. We take $y$ to be generated from a linear combination of the features $\psi$ together with additive noise $\epsilon$:
\begin{align}
    y_\mu = \bar \w \cdot \psi(\x_\mu) + \epsilon_\mu .
\end{align}
Here, we have assumed that the dimension $N$ of the kernel's Hilbert space is finite. We will comment on how to relax this assumption and take $N \to \infty$ faster than $P$ at the end. We remark that very recent works show how one can work directly in an infinite-dimensional Hilbert space using ``dimension-free'' techniques \cite{cheng2022dimension, misiakiewicz2024non}.

Let $\bm \Psi \in \mathbb R^{P \times N}$ be the design matrix, with $\Psi_{\mu i} = \psi_i(\bm x^\mu)$. To apply our earlier results, we would like to claim that in the limit $P, N \to \infty$ with $N/P$ fixed we can replace the empirical covariance matrix $\Sh = \frac{1}{P} \bm \Psi^\top \bm \Psi$ with one where the features are drawn from a Gaussian distribution with matching population covariance. For certain combinations of data distribution and kernel---most simply for the case where $\rho(\x)$ is the uniform measure on the sphere and $K(\x,\x') = k(\x^{\top}\x')$ is a dot-product kernel and if the input dimension $D$ is taken to infinity proportionally with some power of the dataset size---this Gaussian equivalence can be rigorously justified \cite{mei2022hypercontractivity,xiao2022precise,misiakiewicz2022spectrum,hu2022sharp,dubova2023universality, misiakiewicz2024non}. 

Then, using \eqref{eq:lr_eg} and redefining $\kappa \to \kappa/P$, we recover the results of \citet{bordelon2020spectrum, canatar2021spectral}:
\begin{equation}
    E_g = \frac{1}{1-\gamma} \sum_{k=1}^N \frac{\eta_k \bar w_k^2 \kappa^2}{(\kappa + P \eta_k)^2}  + \sigma_\epsilon^2 \frac{\gamma}{1-\gamma}, \quad \gamma = \sum_{k=1}^N \frac{P \eta_k^2}{(\kappa + P \eta_k)^2}.
\end{equation}
Although this calculation was performed at finite $N$, assuming that the spectrum of $\S$ decays quickly enough (as $\eta \sim k^{-b}$ for $b > 1$), one can justify taking $N \to \infty$ at finite $\lambda$. This is because $\df_1$, $\df_2$, and $\tf_1$ will become independent of the cutoff $N$ at this spectral decay, as shown in \sectionsymbol\ref{sec:kernel_scaling}. However, when $\lambda \to 0$ it is not clear that one can interchange the ridgeless limit with the large $N$ limit. It is not obvious when Gaussian equivalence should hold for general kernel methods; some sufficient conditions are obtained in very recent work of \citet{misiakiewicz2024non}, who obtain dimension-free results with non-asymptotic error bounds in $P$. Indeed, one can consider low-dimensional settings in which this theory breaks; see \citet{tomasini2022failure} for examples.

\subsection{\texorpdfstring{The $S$-Transform as a Train-Test Gap}{The S-Transform as a Train-Test Gap}} \label{sec:KARE}

Returning to the general setting of linear regression with structured Gaussian covariates, we can use the same tools to efficiently calculate the training error:
\begin{equation}
\begin{aligned}
    E_{tr} &= \frac{1}{P} \| \y - \hat \y\|^2 \\
        &= \frac{\lambda^2}{P} \|  (\tfrac{1}{P} \X \X^\top + \lambda \mathbf I)^{-1} (\X \bar \w + \e) \|^2 \\
        &\simeq \lambda^2 \bar \w^\top \Sh (\Sh+\lambda \mathbf I)^{-2} \bar \w + \frac{\sigma_\epsilon^2 \lambda^2}{P}  \Tr \left[ (\tfrac{1}{P} \X \X^\top +  \lambda \mathbf I)^{-2} \right]\\
        &= - \lambda^2 \partial_\lambda \bar \w^\top \hat \S (\hat \S + \lambda \mathbf I)^{-1} \bar \w - \sigma_\epsilon^2 \lambda^2 \frac{N}{P} \partial_\lambda \left[ -g_{\frac1P \X \X^\top}(-\lambda) \right].
\end{aligned}
\end{equation}
Here, the passage from the second to the third line holds in expectation over the label noise at any finite size, and when $P$ is large the quadratic form concentrates about its mean over $\e$. Then, using \eqref{eq:tg_rel}, we can write the second term as a derivative on:
\begin{equation} \label{eq:kappa_approximator}
    - g_{\frac{1}{P} \X \X^\top} (-\lambda) = \frac{1 - \df_{\frac{1}{P} \X \X^\top}^1(\lambda)}{\lambda} = \frac{1 - \frac{N}{P} \df_{\hat \S}^1(\lambda)}{\lambda} \simeq \frac{1}{\kappa}.
\end{equation}
We now apply strong deterministic equivalence, giving:
\begin{equation}
\begin{aligned}
    E_{tr} &\simeq \frac{\lambda^2}{1-\gamma} \bar \w^\top \S (\S + \kappa \mathbf I)^{-2} \bar \w + \frac{\sigma_\epsilon^2 }{1-\gamma} \frac{\lambda^2}{\kappa^2}\\
    &= \frac{\lambda^2}{\kappa^2} \left[E_g + \sigma_\epsilon^2 \right].
\end{aligned}
\end{equation}
This relationship was studied in \citet{jacot2020kernel, wei2022more} and also derived in \citet{canatar2021spectral}. If we include noise at test time, the out-of-sample risk is $E_{out} = E_g + \sigma_\epsilon^2$. Recognizing $\lambda^2 / \kappa^2 = S_{\W}(t)^{-2}$ we get:
\begin{equation}
    E_{out} \simeq E_{tr} \, S_{\W}^2(t) = \frac{E_{tr}}{(1 - \frac NP \df_{\S}^1(\kappa))^2} \simeq \frac{E_{tr}}{(1 - \frac NP \df_{\hat \S}^1(\lambda))^2} \equiv E_{GCV}.
\end{equation}
Here, we have recognized the definition of the GCV risk estimator $E_{GCV}$ \cite{golub1979generalized,craven1978smoothing}, which can be estimated \textit{from the training data alone}. Estimating the $S$-transform in this way is also equivalent to the \textbf{kernel alignment risk estimator} (KARE) defined in \citet{jacot2020kernel}. By writing
\begin{equation}
    E_{tr} = \frac{\lambda^2}{P} \y^{\top} (\tfrac{1}{P} \X \X^\top + \lambda)^{-2} \y, \quad 1 - \frac NP \df_{\hat \S}^1(\lambda) = \lambda \tfrac{1}{P} \mathrm{Tr} [ (\tfrac{1}{P} \X \X^\top + \lambda)^{-1} ]
\end{equation}
we get the KARE: 
\begin{equation}
    E_{out} \simeq \frac{\frac{1}{P} \y^\top   (\tfrac{1}{P} \X \X^\top + \lambda \mathbf I)^{-2}  \y }{\left(\frac{1}{P} \mathrm{Tr}\left[ (\frac{1}{P} \X \X^\top + \lambda \mathbf I)^{-1} \right] \right)^2}.
\end{equation}
\citet{wei2022more} have found that this accurately predicts neural scaling laws for kernel regression with the (finite width) neural tangent kernel of a pretrained neural network. 

The $S$ transform also allows us to also estimate $\kappa$ directly from a given training set, without full knowledge of the data distribution of data generating process. This estimate comes from the relationship:
\begin{equation}\label{eq:empirical_S}
    \kappa \simeq \frac{\lambda}{1 - \frac NP \df_{\hat \S}^1(\lambda)}.
\end{equation}
This is equivalent to equation \eqref{eq:kappa_approximator}, namely
\begin{equation}
    \kappa \simeq \frac{1}{-g_{\frac{1}{P} \X \X^\top}(-\lambda)} = \frac{1}{\tfrac{1}{P} \mathrm{Tr}\left[\left( \tfrac{1}{P} \X \X^\top + \lambda \mathbf I\right)^{-1} \right]}.
\end{equation}
By virtue of $\df_{\hat \S}^1(\lambda) \geq 0$ we have that $S \geq 1$ implying that $\kappa \geq \lambda$ and $E_{out} \geq E_{tr}$.

In summary, given a finite size training set, one can come up with an estimate of $\hat S$ of the $S$ transform without full ``omniscient'' knowledge of the data distribution or data generating process. This is given by $\hat S = (1-q \df_{\hat \S}^1(\lambda))^{-1}$. This in turn gives estimates of the renormalized ridge and out of sample error via:
\begin{equation}
    \kappa \simeq \hat S \lambda, \quad E_{out} \simeq \hat S^2 E_{tr}.
\end{equation}

\subsection{Double Descent as a Renormalization Effect}\label{sec:double_desc}

In Equation \eqref{eq:lr_eg} and Figure \ref{fig:unstructured_LR_linspace}, we see that $E_g$ explodes when $\gamma \to 1$. This is the effect that drives the overfitting peak in classical statistical learning. 
In the underparameterized setting $P > N$, we have that $\lambda \to 0$ will imply that the renormalized ridge will also go to zero. Since $\gamma = \frac{N}{P} \df_2 \leq \frac{N}{P}$ we get that the variance explodes only when $N \to P$ and $\lambda \to 0$. In Section \ref{sec:bias_var_kern} we will do a fine-grained analysis of the sources of this variance explosion.

Because one can write $\gamma  =  \df^2_{\frac{1}{P} \X \X^\top}(\kappa) \leq 1$, if $\kappa$ stays at zero in the overparameterized limit, then $\gamma = 1$ and the model will continue to overfit. One will then get infinite generalization error in this setting. 

However, because $\kappa$ becomes renormalized in equation \eqref{eq:kernel_Eg} to be nonzero even when $\lambda = 0$ when $N > P$, one gets that $\df_2 < 1$ in the overparameterized setting. Indeed, in that setting we have $\df_1 = P/N$ so that $S_{W}$ has a pole at $\lambda = 0$. By Equation \eqref{eq:df2_bound} we have $\gamma \leq \frac{1}{1+\kappa/\eta_1}$ where $\eta_1$ is the maximal eigenvalue. Moreover, because, $\kappa$ grows with $N$ in the overparameterized setting, we have that $\gamma$ shrinks away from $1$. The $(1-\gamma)^{-1}$ divergence is then reduced. In this way, the renormalized ridge captures the \textbf{inductive bias} of overparameterization towards simple interpolating solutions that can still generalize well.

\subsection{Multiple Descent without Label Noise}\label{sec:multiple_descent}

\begin{figure}[t]
    \centering
    \includegraphics[width=4in]{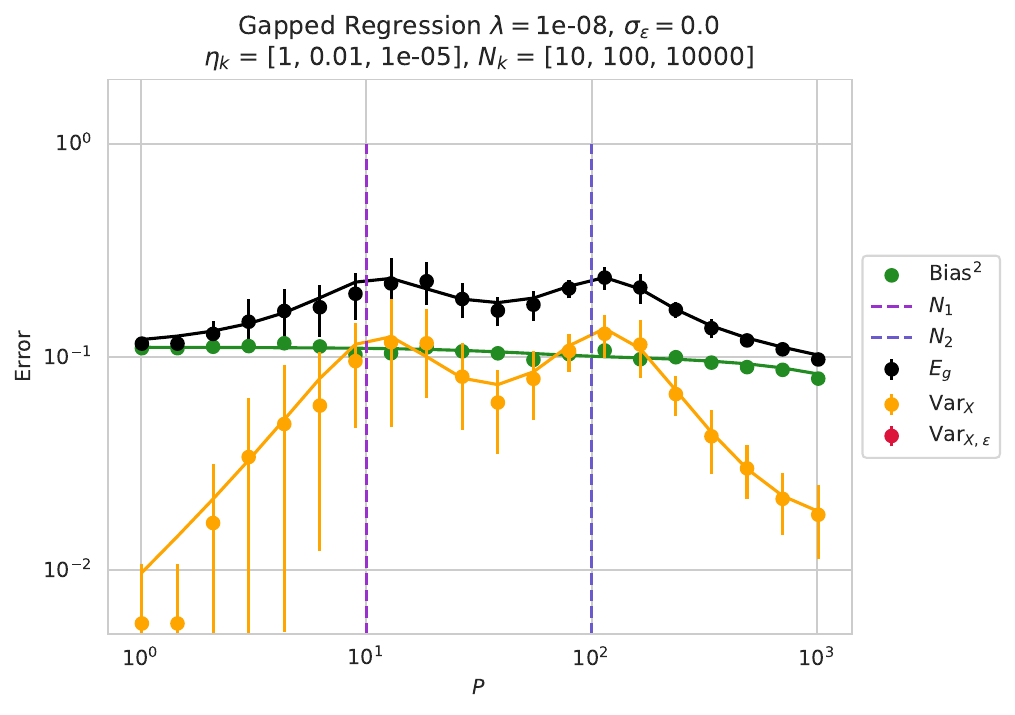}
    \caption{Double descent without label noise in a linear regression task. Here, $\S$ has an eigenspectrum with eigenvalues $\eta_1, \eta_2, \eta_3$ that have values $1, 10^{-2}, 10^{-5}$ and multiplicities $10, 10^2, 10^4$ respectively. The dashed line indicates when $P \approx N_k$. The teacher $\bar \w$ has increasing power in higher modes, given by $1, 10, 10^2$ respectively. The fact that the higher modes are not learnable leads to an effective label-noise like effect that causes this multiple descent phenomenon. We stress that the variance $\mathrm{Var}_{\X, \e} = 0$ since there is no label noise. }
    \label{fig:multiple_descent}
\end{figure}

If one assumes that the spectrum is a series of plateaus at value $\eta_k$ with degeneracy $N_k$ with a large separation of scales between $\eta_k \gg \eta_{k+1}$ and $N_{k+1} \gg N_k$, one can obtain multiple descents, even in the absence of label noise. This phenomenon was studied in the kernel regression setting by \citet{canatar2021spectral,misiakiewicz2022spectrum,dubova2023universality,xiao2022precise,hu2022sharp} and the linear regression setting by \citet{mel2021theory}. In the vicinity of each plateau, one can approximately solve the equation for $\kappa$ by recognizing:
\begin{equation}\label{eq:df1_degen}
    \frac{N}{P} \df_1(\kappa) \approx \frac{1}{P} \left[ \sum_{k<\ell} N_k  + \frac{ \eta_\ell N_\ell}{\kappa + \eta_\ell} + \sum_{k > \ell} \frac{N_k \eta_k}{\kappa} \right].
\end{equation}
    The first term represents all the modes that have been learned. This requires $N_k \ll P$ for each $k$. Since there are only a finite number of $k < \ell$, taking $P, N_\ell$ to scale together linearly and assuming $N_k, k<\ell$ scales sub-linearly compared to $P$, we can neglect the first term. Then, defining $\tilde \sigma_\ell^2 \equiv \frac1P \sum_{k > \ell} N_k \eta_k$ and $q_\ell \equiv N_\ell / P$ we get:
\begin{equation}\label{eq:multiple_descent_kappa}
    \kappa \left( 1- q_\ell \frac{\eta_\ell }{\eta_\ell + \kappa} - \frac{\tilde \sigma_\ell}{\kappa} \right) = \lambda.
\end{equation}
We recognize this as equivalent to the self-consistent equation for $\kappa$ given a spectrum of $N_\ell$ eigenvalues all equal to $\eta_\ell$ and ridge equal to $\tilde \lambda_\ell = \lambda + \tilde \sigma_\ell^2$. This is given by the solution to isotropic linear regression. Explicitly:
\begin{equation}
    \kappa = \frac12 \left(\eta_\ell (q_\ell - 1) +  \tilde \lambda_\ell + \sqrt{(\eta_\ell (q_\ell - 1) +  \tilde \lambda_\ell)^2 + 4 \eta_\ell \tilde \lambda_\ell} \right).
\end{equation}
Similarly, by evaluating $\df_2 = \partial_\kappa (\kappa \df_1)$ from Equation \eqref{eq:df1_degen} one gets:
\begin{equation}
    \gamma \approx q_\ell \frac{\eta_\ell^2}{(\kappa + \eta_\ell)^2}.
\end{equation}
We can then write the generalization error as:
\begin{equation}
    E_g = \frac{\kappa^2}{1-\gamma} \frac{N_d \eta_\ell \bar w_\ell^2}{(\eta_\ell + \kappa)^2} + \frac{1}{1-\gamma}  \sum_{k > \ell} N_k \eta_k \bar w_k^2 + \sigma_\epsilon^2 \frac{\gamma}{1-\gamma}.
\end{equation}
We see that even when $\sigma_\epsilon = 0$, the second term (coming from the non-learnable higher modes) acts as an effective source of noise. We can thus get nonmonotonicity in the generalization error when $\gamma$ increases. We can get the maximum value of $\gamma$ as a function of $q_\ell$ and find that it happens when $q_{\ell} = \frac{\eta_\ell + \tilde \lambda_\ell}{\eta_\ell}$. This gives a double descent peak without label noise, due solely to the variance over the choice of dataset $\X$. We give an example plot of this in Figure \ref{fig:multiple_descent}. We define $\mathrm{Var}_{\X}$ in the subsequent section, Section \ref{sec:bias_var_kern}.

\subsection{Bias-Variance Decomposition} \label{sec:bias_var_kern}

\begin{figure}[t]
    \centering
    \includegraphics[width=4in]{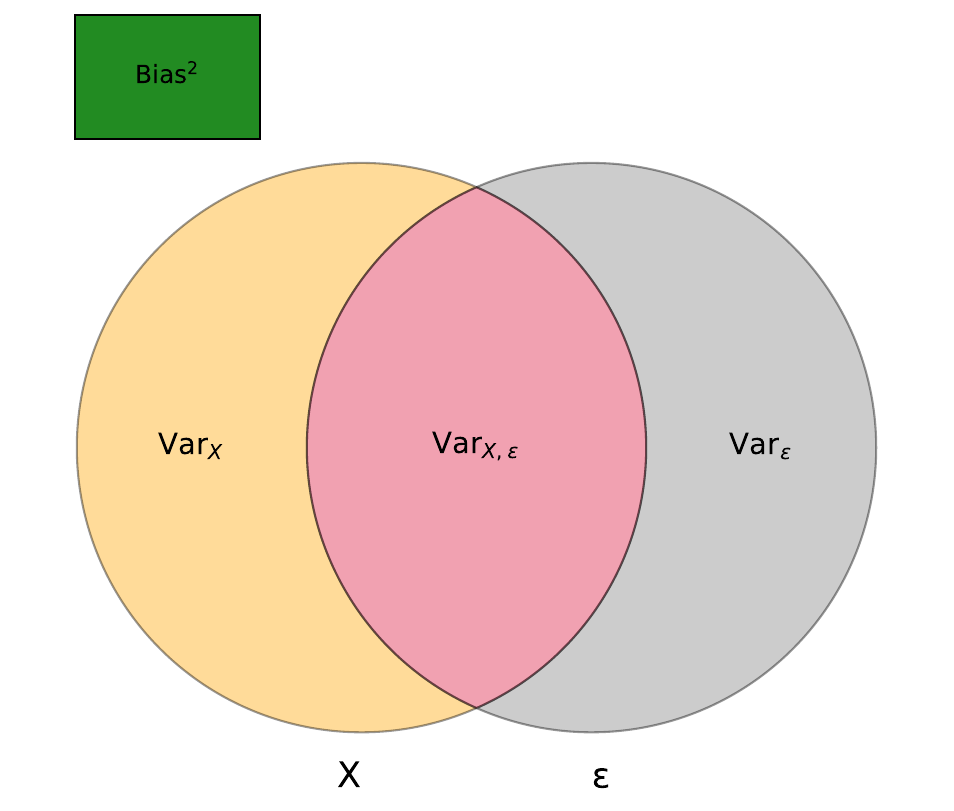}
    \caption{Schematic of the bias-variance decomposition for linear regression. The color scheme matches the plots in Figures \ref{fig:unstructured_LR_linspace}, \ref{fig:multiple_descent} and \ref{fig:structured_LR_ab}. Grey regions do not contribute to variance. }
    \label{fig:LR_BV}
\end{figure}

Although one may be tempted to call the two terms in $E_g$ the \textit{bias} and \textit{variance}, the technical definition in of these two terms in statistical learning is different. The \textit{bias} of an estimator $\hat{\w}$ is defined as: 
\begin{equation}
    \mathrm{Bias}^2 = (\mathbb E_{\mathcal D}[ \hat{\w} ] - \bar \w)^{\top} \S (\mathbb E_{\mathcal D}[ \hat{\w} ] - \bar \w).
\end{equation}
Similarly, the variance is given by:
\begin{equation}
    \mathrm{Variance} = \mathbb E_{\mathcal D} \left[ (\hat{\w} - \mathbb E_{\mathcal D} \hat{\w})^\top \S (\hat{\w} - \mathbb E_{\mathcal D} \hat{\w})^\top  \right].
\end{equation}
The mean squared generalization error can then be written as
\begin{equation}
\begin{aligned}
    E_g &= \mathbb E_{\mathcal D} \left[ (\hat{\w} - \bar \w )^\top \S (\hat{\w} - \bar \w)  \right]\\
    &=  \underbrace{(\mathbb E_{\mathcal D}[ \hat{\w} ] - \bar \w)^{\top} \S (\mathbb E_{\mathcal D}[ \hat{\w} ] - \bar \w)}_{\text{Bias}^2} + \underbrace{\mathbb E_{\mathcal D} \left[ (\hat{\w} - \mathbb E_{\mathcal D}[ \hat{\w} ]) )^\top \S (\hat{\w} - \mathbb E_{\mathcal D}[ \hat{\w} ])  \right]}_{\text{Variance}},
\end{aligned}
\end{equation}
as the cross term vanishes upon expanding the square. Using RMT, one can easily calculate the averaged weights by applying deterministic equivalence:
\begin{equation}
    \mathbb E_{\mathcal D} \hat{\w} = \mathbb E_{\X, \bm \epsilon} \left[ (\hat \S + \lambda)^{-1} (\hat \S \bar \w + \tfrac1P \X^\top \bm \epsilon) \right] \simeq \S (\S + \kappa)^{-1} \bar \w.
\end{equation}
This implies that the Bias$^2$ term is:
\begin{equation}
    \bar \w^\top (\mathbf I - \S (\S + \kappa)^{-1} ) \S (\mathbf I - \S (\S + \kappa)^{-1} ) \bar \w \simeq \kappa^2  \bar \w^\top \S (\bm \S + \kappa)^{-2} \bar \w.
\end{equation}
Given that we know the total generalization error, the correct bias-variance decomposition over $\mathcal D$ is:
\begin{equation}
    E_g \simeq \underbrace{\kappa^2  \bar \w^\top \S (\bm \S + \kappa)^{-2} \bar \w}_{\text{Bias}^2} + \underbrace{\frac{\gamma}{1-\gamma} \left[ \kappa^2  \bar \w^\top \S (\bm \S + \kappa)^{-2} \bar \w  + \sigma_\epsilon^2 \right]}_{\text{Variance}}.
\end{equation}

Assume we have $B$ different datasets all of size $P$ with estimators given by $\hat{\w}_b$. We can \textbf{bag} by taking our final learned weights to be 
\begin{equation}
    \hat{\w}_B = \frac{1}{B} \sum_{b =1}^B \hat{\w}_b.
\end{equation}
We note that by linearity of expectation $\mathbb E[\hat{\w}_B] = \mathbb E[\hat{\w}_b]$ for each $b$. Thus the bias term remains the same, while the variance is reduced by $1/B$. This means that bagging corresponds to keeping $\kappa$ fixed but performing an effective rescaling 
\begin{equation}
    \frac{\gamma}{1-\gamma} \to \frac{1}{B} \frac{\gamma}{1-\gamma}.
\end{equation}

The variance term can in fact be further decomposed, as in \citet{adlam2020understanding}, into the variance due to the choice of training set $\mathrm{Var}_{\bm X}$, the variance due to the label noise $\mathrm{Var}_{\bm \epsilon}$, and the joint variance due to their interaction $\mathrm{Var}_{\X, \e}$. We can remove the latter two without affecting the former by averaging over label noise holding training set fixed. We get that:
\begin{equation}
    \mathbb E_{\bm \epsilon} \hat{\w} = (\hat \S + \lambda)^{-1} \hat \S \bar \w.
\end{equation}
For this estimator, we see that the respective generalization error and variance (over $\X$) are
\begin{align}
    E_g(\mathbb E_{\bm \epsilon} \hat{\w}) = \frac{\kappa^2}{1-\gamma}  \bar \w^\top \S (\bm \S + \kappa)^{-2} \bar{\w},\\ 
    \mathrm{Var}_{\X} = \mathrm{Var} [\mathbb E_{\bm \epsilon} \hat{\w}] = \frac{\kappa^2 \gamma}{1-\gamma}  \bar \w^\top \S (\bm \S + \kappa)^{-2} \bar{\w}.
\end{align}
This gives an interpretation of $\gamma$ as the fraction of the test error due to the variance induced by the choice of training set $\X$ (after removing the effect of noise): 
\begin{equation}
    \gamma = \frac{\mathrm{Var} [\mathbb E_{\bm \epsilon} \hat{\w}]}{E_g(\mathbb E_{\bm \epsilon} \hat{\w})} = \frac{\mathrm{Var}_{\X}}{\mathrm{Bias}^2 + \mathrm{Var}_{\X}}.
\end{equation}
Because averaging over $\X$ at a fixed noise level $\sigma_\epsilon$ also removes the label noise term, we get that $\mathrm{Var}_{\e} = 0$ and 
\begin{equation}
    \mathrm{Var}_{\X, \e} = \frac{\sigma_\epsilon^2 \gamma}{1-\gamma}.
\end{equation}
That is, the variance due to noise always enters through its interaction with the variance due to the finite choice of training set.
Inspired by the work of \citet{adlam2020understanding}, we visualize this decomposition as a Venn diagram in Figure \ref{fig:LR_BV}. We will do the same in the next section as well, in Figure \ref{fig:LRF_BV}.

\subsection{\texorpdfstring{Scaling Laws in $P$}{Scaling Laws in P}} \label{sec:kernel_scaling}

\subsubsection{Normalizable Spectra}

We consider here the derivation of the scaling properties of the loss when both the singular values for the covariance and the target weights decay as power laws. The scalings of the loss under these assumptions were obtained in \citet{bordelon2020spectrum, spigler2020asymptotic, caponnetto2005fast, caponnetto2007optimal}. One motivation studying such power law structure datasets comes from the observation of its presence across a wide variety of modern machine learning datasets \cite{levi2023underlying, maloney2022solvable}. In vision datasets, the presence of power law structure in their covariances has been observed in \citet{ruderman1997origins, hyvarinen2009natural}.

We take the spectrum of the kernel to scale as
\begin{equation}
    \eta_k \sim k^{-\alpha}.
\end{equation}
Here $\alpha$ is known as the \textbf{capacity} exponent as in \citet{caponnetto2005fast, caponnetto2007optimal, pillaud2018statistical, cui2023error, steinwart2009optimal, cui2021generalization}. The task decomposes into the eigenspaces also as a power law with
\begin{equation}
    \bar w_k^2 \eta_k \sim k^{-(1+2 \alpha r)}.
\end{equation}
Here $r$ is the \textbf{source} exponent. The exponent $2 \alpha r$ determines how much of $\w$ remains above eigenmode $k$ as measured by $\w^\top \S \w$. That is, $\sum_{k' > k} w_{k'}^2 \eta_{k'} \sim k^{-2 \alpha r}$. The source exponent also plays a fundamental importance for the scaling SGD after $t$ steps of population gradient flow on this dataset, where one can show that the online loss $\mathcal L$ scales as $t^{-2r}$ \cite{bordelon2021learning}. 

Interpreting the input space as the reproducing kernel Hilbert space of a kernel with eigenspectrum given by $\eta_k$, then $\alpha$ controls the spectral decay of the kernel. Smaller $\alpha$ lead to more expressive but jagged functions while larger $\alpha$ lead to a stronger prior towards smoothness.

The self-consistent equation for $\kappa$ is approximated by:
\begin{equation}
    \kappa \approx \frac{\lambda}{1-\frac{1}{P} \int_1^\infty \frac{k^{-\alpha}}{k^{-\alpha}+\kappa} dk}.
\end{equation}
Making the change of variables $u = k \kappa^{1/\alpha}$ then gives
\begin{equation}\label{eq:sce_power}
    \kappa \approx \frac{\lambda}{1 - \frac{\kappa^{-1/\alpha}}{P} \int_{\kappa^{1/\alpha}}^\infty \frac{1}{1+u^\alpha}  du} = \frac{\lambda}{1 - \frac{\kappa^{-1/\alpha}}{P} F(\alpha, \kappa)}
\end{equation}
for a function $F$ that depends on $\alpha, \kappa$. Let's consider first the ridgeless limit $\lambda \to 0$. Then in order to get a nonzero value of $\kappa$, we need
\begin{equation}
    \kappa^{-1/\alpha} F(\alpha, \kappa) \sim P.
\end{equation}
Note as $\kappa \to 0$, $F$ tends to a constant and so we get the scaling $\kappa \sim P^{-\alpha}$. In the other case, when $\lambda$ is large, namely $\lambda \gg P^{-\alpha}$ we get that $\kappa \sim \lambda$.

\begin{figure}[t]
    \centering
    \includegraphics[width=4in]{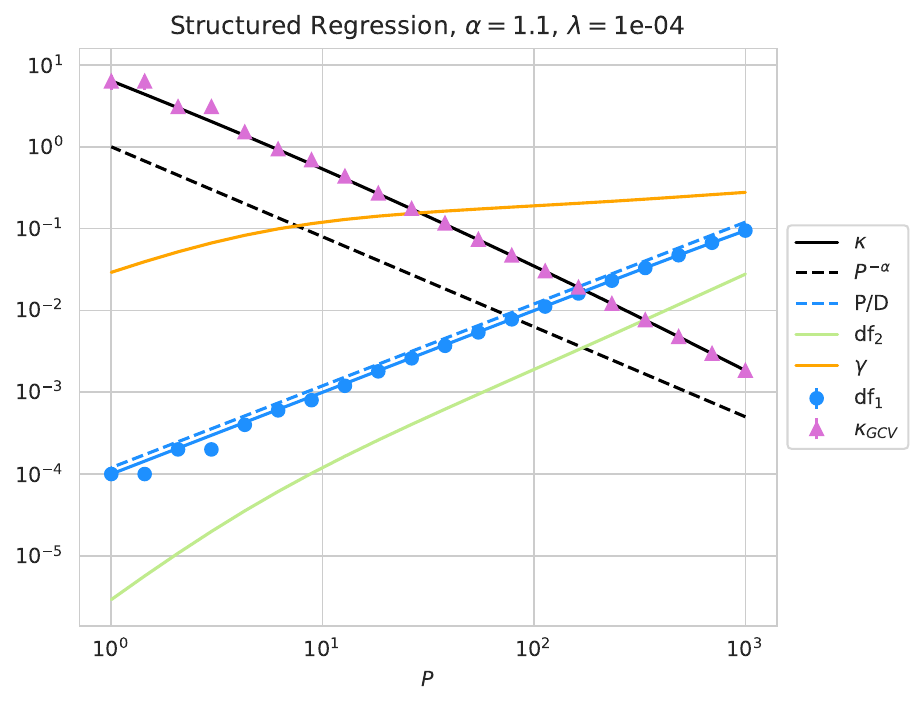}
    \includegraphics[width=4in]{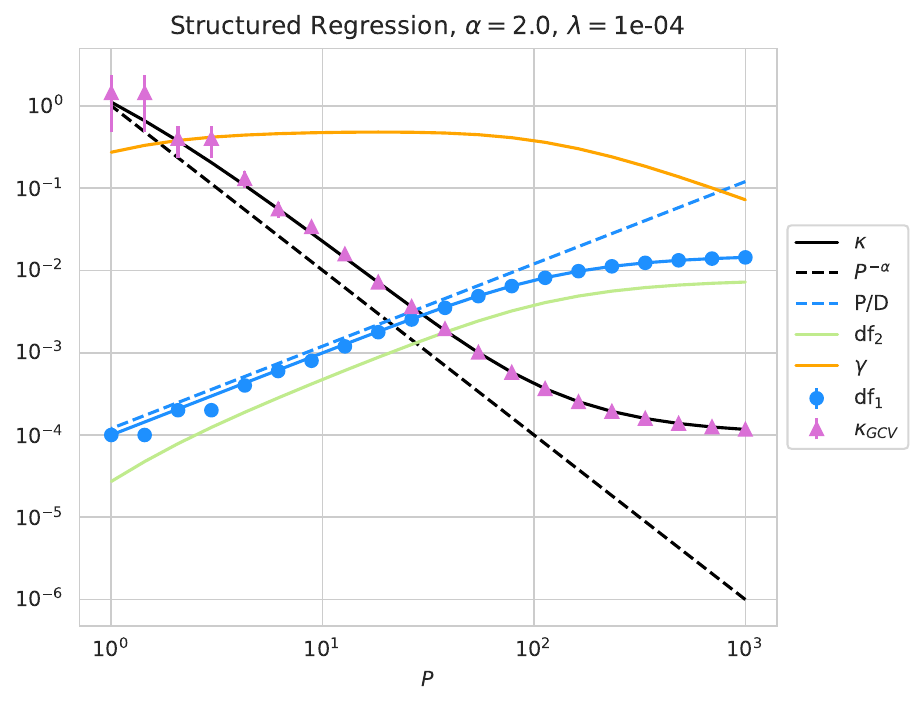}
    \caption{Left: Scaling of various relevant parameters for power-law structured data. The analytic solution for $\kappa$ is plotted (solid black), as well as its GCV estimate from the data given by $S(-\df_{\Sh}(\lambda)) \lambda$ (orchid triangles). The scaling law $P^{-\alpha}$ is also plotted (dashed black), showing excellent agreement. We also plot $\df^1_{\S}(\kappa)$ (solid blue) and its empirical estimate $\df^1_{\Sh}(\lambda)$ (blue circles), finding excellent agreement. We also plot the scaling law $P/N$ (dashed blue). Finally, we plot $\df_2$ and $\gamma = \frac{P}{N} \df_2$ (dashed green and yellow respectively). We see that $\gamma$ is relatively constant across $P$. For faster decays it would be more constant still. Right: The same, with faster spectral decay. We find agreement until $\kappa \sim \lambda$, where we enter the ridge-dominated scaling regime highlighted in Equation \eqref{eq:LR_Eg_scaling}. }
    \label{fig:structured_LR_params}
\end{figure}

Similarly for $\gamma$ one gets the approximation:
\begin{equation}
    \gamma \approx \frac{1}{P} \int_1^\infty \left(\frac{k^{-\alpha}}{k^{-\alpha}+\kappa}\right)^2 dk = \frac{\kappa^{-1/\alpha}}{P} \int_{\kappa^{1/\alpha}}^\infty \frac{1}{(1 + u^\alpha)^2} du.
\end{equation}
Taking $\kappa \sim P^{-\alpha}$ we see that $\gamma$ remains constant as $P$ increases. If $\kappa \sim \lambda$ one gets that $\gamma \sim \lambda^{-1/\alpha}/P \to 0$ as $P$ increases. In all cases, $1/(1-\gamma)$ tends to a constant, so we can therefore write the generalization error scaling as:
\begin{equation}
    E_g \sim \int_1^\infty \frac{k^{-(1+2\alpha r)}}{(1+k^{-\alpha}/\kappa)^2} dk \sim P^{-2\alpha r} \int_{1/P}^\infty \frac{u^{-(1+2\alpha r)}}{(1+u^{-\alpha})^2} du, \quad u= k/P.
\end{equation}
We can split this integral into a part near $u \sim 1/P$ and a part away from that. The part near $1/P$ will scale as $(1/P)^{-2\alpha r +2\alpha}$ and thus give a contribution scaling as $P^{-2\alpha}$. The part away from that is $P$-independent and thus its contribution scales as $P^{-2\alpha r}$. 

\begin{figure}[t]
    \centering
    \includegraphics[width=4in]{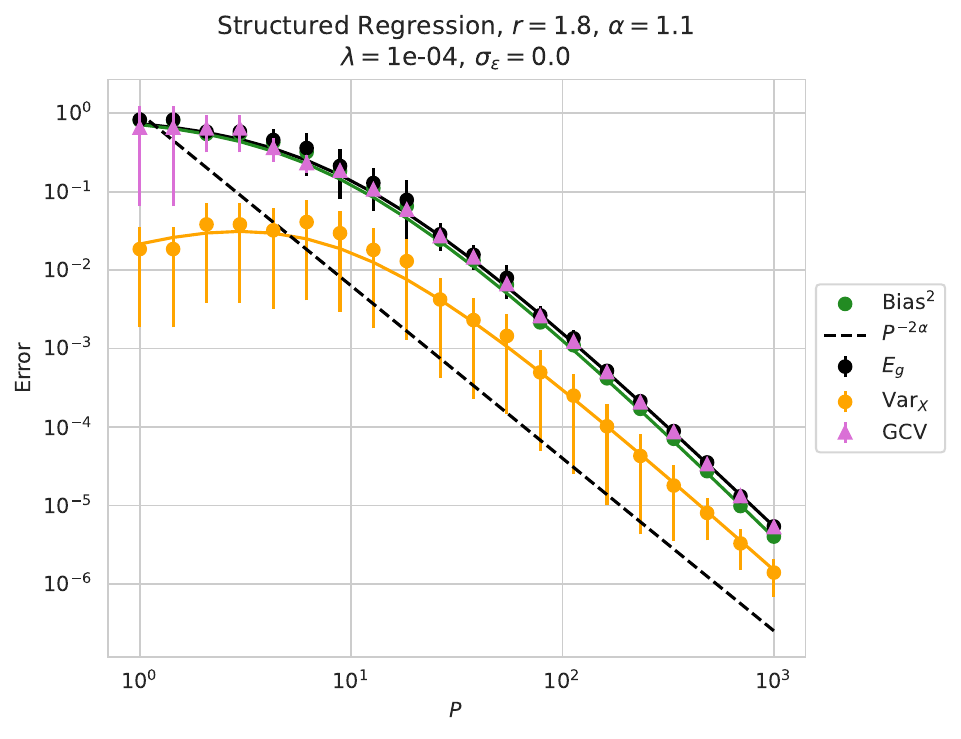}
    \includegraphics[width=4in]{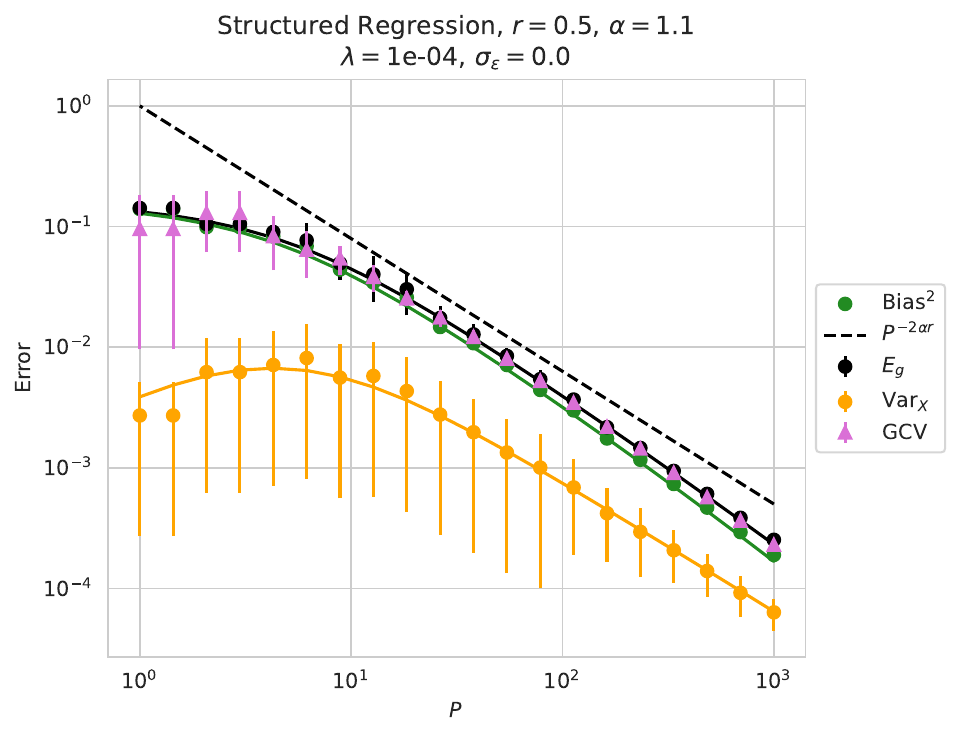}
    \caption{Generalization error (solid black) for two different teacher decay constants. We see that $\min(1, r)$ determines the whether the scaling law is due solely to the capacity or if the source also plays a role. The bias (solid green) variance over the dataset (solid orange) follow identical scaling laws. The results of empirical simulations are plotted in solid dots, showing excellent agreement. The GCV estimate from the training error is given by orchid triangles. Here, $N=10000$, and the spectral decay makes the final result insensitive to $N$.}
    \label{fig:structured_LR_ab}
\end{figure}

When $\kappa \sim \lambda$ we can similarly change variables taking $u = k \lambda^{1/\alpha}$ and track the $\lambda$ dependence:
\begin{equation}
    E_g \approx \lambda^{2 r} \int_{\lambda^{1/\alpha}}^\infty \frac{u^{-(1+2 \alpha r)}}{(1+u^{-\alpha})^2} du.
\end{equation}
The contributions of this integral can again be broken up into the part near $\lambda^{1/\alpha}$ and the part away from it, which is $\lambda$ independent. The two contributions then scale as $\lambda^2$ and $\lambda^{2r}$ respectively.

This altogether gives the scaling laws:
\begin{equation}\label{eq:LR_Eg_scaling}
    E_g \sim \begin{cases}
        P^{-2 \alpha \min(r, 1)}, & P \ll \lambda^{-1/\alpha}\\
        \lambda^{2 \min(r, 1)}, & P \gg \lambda^{-1/\alpha},\\
    \end{cases}
\end{equation}
where we remind the reader that $\lambda$ is assumed to be small. 
After redefining $\lambda \to \lambda / P$, one obtains the scaling laws of \cite{bordelon2020spectrum}. Given that $\alpha > 1$ for the spectrum to be normalizable, we get that in the noiseless setting, adding explicit regularization will hurt generalization. Further, we see that faster spectral decays will improve performance, as will having more of the task's power in the top eigenmodes. Either effect can bottleneck the other, hence the $\min$ in the exponents.

One can also average over teachers. This corresponds to taking $\bar \w_k$ to be constant, or equivalently $1+2 \alpha r = \alpha$.  This sets $r = \frac12 \frac{\alpha-1}{\alpha}$. In the ridgeless limit this gives the scaling $E_g \sim P^{- (\alpha - 1)}$

In the case where $\lambda$ itself scales as $P^{-l}$ for some value $l$ as in \citet{cui2021generalization}, one gets:
\begin{equation}\label{eq:LR_Eg_scaling_l}
    E_g \sim P^{-2\, \min(\alpha, l)\, \min(r, 1)}.
\end{equation}
If we incorporate label noise, using the fact that $\gamma$ is a constant in the first case of Equation \eqref{eq:LR_Eg_scaling} and $\gamma \sim \lambda^{-1/\alpha}/P$ in the second, we find four scaling regimes:
\begin{equation}\label{eq:LR_Eg_scaling_full}
\begin{aligned}
     E_g &\sim \begin{cases}
        P^{-2 \alpha \min(r, 1)} + \sigma_\epsilon^2, & \alpha \ll l\\
        P^{-2 \ell \min(r, 1)} + \sigma_\epsilon^2 P^{-(\alpha-l)/\alpha}  , & l \ll \alpha
    \end{cases} \\
    &\sim \begin{cases}
        P^{-2 \alpha \min(r, 1)}, &  l > \alpha, \; \sigma_\epsilon \ll P^{-\alpha \min(r,1)}, \quad \text{Signal dominated} \\
        \sigma_\epsilon^2 P^0 & l > \alpha, \; \sigma_\epsilon \gg P^{-\alpha \min(r,1)}, \quad \text{Noise dominated} \\
        P^{-2 l \min(r, 1)} &  l < \alpha, \; l < \frac{\alpha}{1 + 2 \alpha \min(r,1)}, \quad \;\; \text{Ridge dominated} \\
        \sigma_\epsilon^2 P^{-(\alpha-l)/\alpha}, & l < \alpha, \; l \geq \frac{\alpha}{1 + 2 \alpha \min(r,1)}, \quad \;\; \text{Noise mitigated}
    \end{cases}
\end{aligned}
\end{equation}
This recovers the four scaling regimes studied in \citet{cui2021generalization}. These four regimes yield the different possible resolution-limited scalings of wide neural networks in the kernel setting trained on power-law data. The first two are effectively ridgeless, whereas one requires a explicit ridge to achieve the second two scaling laws.

\subsubsection{Non-Normalizable Spectra} \label{sec:non_normalizable}
If the spectrum has $\alpha \leq 1$, then the final scaling laws will depend on the value of $N$. We study the regime where $P \ll N$.  In this case, in the ridgeless limit the following term must be order 1. The integral is dominated by the large $N$ limit:
\begin{equation}
    \frac{N}{P} \df_1 = \frac{1}{P} \int_1^N \frac{dk}{1 + \kappa k^{\alpha}}  \sim \frac{N^{1-\alpha}}{P\kappa} \Rightarrow \kappa \sim \frac{N^{1-\alpha}}{P}.
\end{equation}
When $\alpha = 0$ this reproduces the leading order in $N$ scaling of isotropic linear regression, where $\kappa = q-1$. Further, we get that $\gamma$ has a nontrivial $P$ scaling:
\begin{equation}\label{eq:gamma_scaling}
    \gamma = \frac{1}{P} \int_1^N \frac{dk}{(1 + \kappa k^{\alpha})^2} \sim \left(\frac{P}{N}\right)^{\min(1, \frac{1-\alpha}{\alpha})}.
\end{equation}
The former scaling occurs when $\alpha < 1/2$, leading to the upper limit dominating, while the latter happens when $\alpha > 1/2$.  In this setting,  when $P \to N$ we get that $\gamma \to 1$ and the generalization error explodes. We thus see how a slowly decaying spectrum can lead to non-monotonicity in the generalization error.

\clearpage

\section{Linear Random Features}
\label{sec:lrf}

In this section, we will make extensive use of the \textbf{push-through identity} \cite{horn2012matrix}:
\begin{equation}\label{eq:pushthrough}
    \A (\B \A + \lambda \mathbf I)^{-1} = (\A \B + \lambda \mathbf I)^{-1} \A.
\end{equation}

\subsection{Setup and Motivation}\label{sec:LRF_motivation}

We consider a general class of linear random feature models of the form
\begin{equation}\label{eq:RF_dfn}
    f(\bm x) = \bm{x}^{\top} \F \bm v,
\end{equation}
where $\F \in \mathbb{R}^{D \times N}$ is not trainable and maps the data from $\mathbb R^D$ to an $N$-dimensional \textbf{feature space}. Here, $\bm v \in \mathbb R^N$ is a vector of trainable parameters. Our statistical assumptions on the training data are the same as in \sectionsymbol\ref{sec:LR}: we take $\X \in \mathbb R^{P \times N}$ with rows distributed as $\x_\mu \sim \mathcal N(\bm 0, \S)$, and generate labels as $y_\mu = \bar \w \cdot \x_\mu + \epsilon_\mu$ with each $\epsilon_\mu \sim \mathcal{N}(0, \sigma_{\epsilon}^2)$.

This is the simplest solvable model where the notion of parameters $N$ can enter  on a different footing from the input dimension. This model is very limited when viewed literally as a neural network learning functions from a $D$-dimensional input space, since it can only learn linear functions. However, an alternative perspective put forth in \citet{maloney2022solvable} considers that with $D \gg N, P$, one can instead view the $D$-dimensional space as an abstract feature space. This space can be viewed e.g. as the Hilbert space of functions that are square-integrable with respect to the Gaussian data distribution, or the Hilbert space induced by the NTK of some infinitely wide network. From this space, we are taking an $N$-dimensional random feature projection corresponding to the $N$ parameters of some model. Similar motivation is given in \citet{atanasov2022onset, bordelon2024dynamical} where the input space is viewed as an analogue of the infinite-width NTK's kernel Hilbert space.

We minimize the same MSE objective as in Equation \eqref{eq:ridge_loss}. This gives the following learned weights $\hat{\v}$:
\begin{equation}
\begin{aligned}
    \hat{\v} &= (\F^\top \X^\top \X \F + P \lambda \mathbf I)^{-1} \F^\top \X^\top \y.
\end{aligned}
\end{equation}
The corresponding learned weights in $\mathbb R^D$ are $\hat{\w} = \F \hat{\v} \in \mathbb R^D $. Then, taking $\Sh = \frac{1}{P} \X^\top \X$ and applying the pushthrough identity \eqref{eq:pushthrough}:
\begin{equation}
    \bar \w - \hat{\w} = \lambda (\F \F^\top \Sh + \lambda \mathbf I)^{-1} \bar \w - (\F \F^\top  \Sh  +  \lambda \mathbf I)^{-1} \F \F^\top \frac{\X^\top \bm \epsilon}{P}.
\end{equation}
The generalization error is $E_{g} = (\bar \w - \hat{\w})^\top \S (\bar \w - \hat{\w})$ and just as in Equation \eqref{eq:LR_Eg_signal_noise} in the linear regression setting, it can be  decomposed into signal and noise components. After expanding and applying \eqref{eq:pushthrough} again, the noise component can be written as:
\begin{equation}\label{eq:LRF_noise}
\begin{aligned}
    \text{Noise} &= \frac{\sigma_\epsilon^2}{P} \Tr[\Sh  \F \F^\top (\Sh \F \F^\top + \lambda  \mathbf I)^{-1} \S \F \F (\Sh \F \F^\top + \lambda  \mathbf I)^{-1}  ] \\
    &= -\frac{\sigma_\epsilon^2}{P} \partial_\lambda  \left[\lambda \Tr[\S \F \F^\top (\Sh \F \F^\top +\lambda \mathbf I)^{-1} ] \right].
\end{aligned}
\end{equation}
For now, we will assume that $\F \F^\top$ is invertible. Then, the signal component is:
\begin{equation}\label{eq:LRF_signal}
\begin{aligned}
    \text{Signal}&=\lambda^2 \bar \w^\top (\hat \S \F \F^\top + \lambda \mathbf I)^{-1} \S (\F \F^\top \hat \S + \lambda \mathbf I)^{-1}  \bar \w\\
    &=\lambda^2 \bar \w^\top (\hat \S \F \F^\top + \lambda \mathbf I)^{-1} \S \F \F^\top (\hat \S \F \F^\top + \lambda \mathbf I)^{-1} (\F \F^\top)^{-1}  \bar \w\\
    &= -\lambda^2 \partial_{J}|_{J=0} \left[ \bar \w^\top \left[ (\hat \S + J \S) \F \F^\top + \lambda\right]^{-1} (\F \F^\top)^{-1}  \bar \w \right].
\end{aligned}
\end{equation}
Here, we have applied the push-through identity \eqref{eq:pushthrough} and used the same differentiation trick as in Section \ref{sec:LR_derivation}. 

\subsection{Averaging Over Data}

We will now perform an $\X$ average, viewing $\F$ as fixed. 
Then, applying the subordination relation Equation \eqref{eq:subordination2}, we have the following deterministic equivalence:
\begin{align}
    \Sh \F \F^\top (\Sh \F \F^\top + \lambda \mathbf I)^{-1} &\simeq  \F \F^\top (\F \F^\top + \lambda S_{\S} S_{\W} \mathbf I)^{-1} \nonumber \\
    &\simeq \S \F \F^\top (\S \F \F^\top + \kappa_1 \mathbf I)^{-1}, \\
    \Rightarrow \lambda (\Sh \F \F^\top + \lambda \mathbf I)^{-1} &\simeq   \kappa_1 (\S \F \F^\top + \kappa_1 \mathbf I)^{-1}, \label{eq:kappa1_det_equiv}\\
    \kappa_1 \equiv \lambda S_{\W} &= \frac{\lambda}{1 - \frac{D}{P} \df^1_{\S \F \F^\top}(\kappa_1)}. \label{eq:kappa1_defn}
\end{align}
Here $\W$ is a white Wishart with $q=D/P$. Defining $\S_{\F} \equiv \S^{1/2} \F \F^\top \S^{1/2}$, we see that because this shares the same nonzero eigenvalues as $\S \F \F^\top$ that $\df^1_{\S \F \F^\top}(\kappa_1) = \df^1_{\S_{\F}}(\kappa_1)$. Then,
\begin{equation}
    \frac{d \kappa_1}{d \lambda} = \frac{1}{1-\gamma_1}, \quad \gamma_1 \equiv \frac{D}{P} \df^2_{\S_{\F}}(\kappa_1).
\end{equation}
Applying \eqref{eq:kappa1_det_equiv} to \eqref{eq:LRF_noise}, the $\X$-averaged noise term  becomes:
\begin{equation}
\begin{aligned}
    \text{Noise} &\simeq - \sigma_\epsilon^2\frac{D}{P} \partial_\lambda [\kappa_1 \df^1_{\S_{\F}}(\kappa_1) ] \\
    &= \sigma_\epsilon^2 \frac{d \kappa_1}{d \lambda} \frac{D}{P} \df^2_{\S_{\F}}(\kappa_1) 
    = \sigma_\epsilon^2 \frac{\gamma_1}{1-\gamma_1}.
\end{aligned}
\end{equation}
Here, we have used Equation \eqref{eq:df1df2_rel1}. Here we have used the fact that all quantities concentrate over $\F$ to drop the expectation.

The signal term \eqref{eq:LRF_signal} can be obtained using the exact same argument as in Equations \eqref{eq:eval_signal1} and \eqref{eq:eval_signal2}. This gives
\begin{equation}
\begin{aligned}\label{eq:RF_bias}
    \text{Signal} &\simeq
     \frac{\kappa_1^2}{1-\gamma_1} \bar \w^\top  \S \F \F^\top (\S \F \F^\top   +  \kappa_1 \mathbf I)^{-2} (\F \F^\top)^{-1} \bar \w\\
     &= \frac{\kappa_1^2}{1-\gamma_1} \bar \w^\top  \S^{1/2}  (\S_{\F}   +  \kappa_1 \mathbf I)^{-2} \S^{1/2}  \bar \w.
\end{aligned}
\end{equation}
We can thus write the full generalization compactly as:
\begin{equation}\label{eq:LRF_gen}
    E_g^{\F} \simeq - \frac{\kappa_1^2}{1-\gamma_1} \partial_{\kappa_1} \widetilde{\tf}_1(\kappa_1) +\sigma_\epsilon^2 \frac{\gamma_1}{1 - \gamma_1}.
\end{equation}
Here, we have defined the function
\begin{align}
    \widetilde{\tf}_1(\kappa_1) \equiv  \bar \w^\top  \S^{1/2}  (\S_{\F}   +  \kappa_1 \mathbf I)^{-1} \S^{1/2} \bar{\w}.
\end{align}
We add a tilde to highlight that $\widetilde{\tf}_1(\kappa_1)$ depends on both $\S$ and $\F$.

Importantly, observe that these asymptotic results are continuous in $\F$, even when $\F \F^\top$ is not invertible. To extend this argument to the regime in which $\F\F^{\top}$ is singular, we infinitesimally regularize $\F \F^\top$ as $\F \F^\top + \tau \mathbf I_{D}$, and then let $\tau$ tend to zero after averaging over $\X$. The validity of this interchange of limits can be justified using dominated convergence. An alternative proof of this fact would follow from high-probability bounds on the deviation of the non-averaged generalization error from the deterministic limit, in a similar spirit to the bounds given in \citet{hastie2022surprises}.

\subsection{Averaging Over Features}

We can now perform the $\F$ average in the above equations. Again applying Equation \eqref{eq:subordination2}, we have the deterministic equivalence:
\begin{equation}\label{eq:kappa2_det_equiv}
    \S_{\F}  (\S_{\F} + \kappa_1 \mathbf I)^{-1} = \S  (\S + \kappa_1 S_{\F \F^\top} \mathbf I)^{-1}.
\end{equation}
We thus have that $\kappa_1$ will be further renormalized to
\begin{equation}\label{eq:S_RF}
\boxed{
\begin{aligned}
    \kappa_2 &\equiv \kappa_1 S_{\F \F^\top}(-\df_1) = \lambda S_{\W}(-\df_1)  S_{\F \F^\top}(-\df_1).\\
\end{aligned}
}
\end{equation}
We adopt the shorthand $\df_1 \equiv \df^{1}_{\S}(\kappa_2) \simeq \df^1_{\S_{\F}}(\kappa_1) \simeq \df^1_{\Sh \F \F^\top}(\lambda)$, $\df_2 \equiv \df^2_{\S}(\kappa_2)$.
This is a different renormalization effect, due to the fluctuations not in the data, but in the features. It is equivalent to the effect studied in \citet{jacot2020implicit,patil2024asymptotically}. Then, we have
\begin{equation}
\begin{aligned}
    \gamma_1 &= \frac{D}{P} \df^1_{\S_{\F}}(\kappa_1) \left(1 + \frac{\kappa_1}{\df^1_{\S_{\F}}(\kappa_1)} \partial_{\kappa_1} \df^1_{\S_{\F}}(\kappa_1) \right)\\
    &\simeq \frac{D}{P} \df_1 \, \left(1 + \frac{\kappa_1}{\df_1} \frac{d \kappa_2}{d \kappa_1} \partial_{\kappa_2} \df_1 \right).
\end{aligned}
\end{equation}
Applying Equation \eqref{eq:df1df2_rel3} gives:
\begin{equation}\label{eq:gamma2}
\boxed{
    \gamma_1 = \frac{D}{P} \df_1 \, \left(1 - \frac{\df_1-\df_2}{\df_1} \frac{d \log \kappa_2}{d \log \kappa_1}  \right).
    }
\end{equation}
Next, we can apply Equation \eqref{eq:kappa2_det_equiv} to the signal term in Equation \eqref{eq:LRF_gen} and get:
\begin{equation}\label{eq:F_avg_signal}
\begin{aligned}
    \text{Signal} = - \frac{\kappa_1^2}{1-\gamma_1} \partial_{\kappa_1} \left[ \frac{\kappa_2}{\kappa_1} \tf_{\S}^1 (\kappa_2) \right]
    &= -\frac{\kappa_2^2 \tf_1'}{1-\gamma_1} \frac{d \log \kappa_2}{d \log \kappa_1} + \frac{\kappa_2 \tf_1}{1-\gamma_1} \left[ 1 -  \frac{d \log \kappa_2}{d \log \kappa_1}\right],
\end{aligned}
\end{equation}
where again we have used shorthand $\tf_1 = \tf_{\S, \bar \w}(\kappa_2)$. Together with Equations \eqref{eq:S_RF} and \eqref{eq:gamma2}, this gives the final result:
\begin{equation}\label{eq:LRF_finite_ridge}
\boxed{
    E_g = -\frac{\kappa_2^2 \tf_1'}{1-\gamma_1} \frac{d \log \kappa_2}{d \log \kappa_1} + \frac{\kappa_2 \tf_1}{1-\gamma_1} \left[ 1 -  \frac{d \log \kappa_2}{d \log \kappa_1}\right] + \sigma_\epsilon^2 \frac{\gamma_1}{1-\gamma_1}. 
    }   
\end{equation}
This equation  recovers and extends the generalization error formulas of all linear random feature models in the literature. We will give explicit examples in Section \ref{sec:lrf_examples}.

\subsection{Ridgeless Limits}\label{sec:LRF_ridgeless}

In the limit of $\lambda \to 0$, we see that nonzero values of $\kappa_2$ will correspond to $\df$ taking a value so that it lands on of the poles of one of the $S$-transforms. In this way, we see that poles in the $S$-transform determine the different regimes of the ridgeless limit. In what follows, let $N_\ell$ be the rank of $\F \F^\top$. $N_\ell$ will correspond to the narrowest width in the random feature model in the subsequent examples.  There are three possible behaviors as $\lambda \to 0$:
\begin{enumerate}
    \item $\kappa_2$ stays zero. This happens when $\mathrm{rank}(\Sh \F \F^\top) = \mathrm{rank}(\S \F \F^\top) = \mathrm{rank}(\S)$.  All matrices are full rank, which constrains $D \leq N_\ell, P$. This is the \textbf{underparameterized} setting.
    
    Here, because $\tf_1(\kappa_2)$ is analytic as $\kappa_2 \to 0$, we get that the signal term vanishes completely. Further, because $\df_1 = \df_2 = 1$ at $\kappa_2 = 0$, we have that $\gamma = D/P$. Altogether this gives a generalization error of
    \begin{equation}\label{eq:underparam_RF}
        E_g = \frac{D/P}{1 - D/P} \sigma_\epsilon^2.
    \end{equation}
    This is independent of any details of the structure of the features $\F$.

    \item $\kappa_1$ stays zero but $\kappa_2$ is nonzero. This happens when $\mathrm{rank}(\Sh \F \F^\top) = \mathrm{rank}(\S \F \F^\top) < \mathrm{rank}(\S)$. This means that $\F \F^\top$ is no longer full rank. We have $N_\ell < D, P$.  This is the \textbf{bottlenecked} setting.

    Here, we get that $\df_1 = \df^{(1)}_{\S}(\kappa_2) \to \df_{\Sh}^{(1)}(0) = \frac{N_{\ell}}{D}$ since $\Sh$ has rank $N_\ell$ instead of $D$. We also get  $\frac{d \log \kappa_2}{d \log \kappa_1} = \frac{\kappa_1}{\kappa_2} \frac{d \kappa_2}{d \kappa_1} \to 0$ as $\kappa_1 \to 0$. Consequently $\gamma_1 = N_{\ell}/P$. This gives:
    \begin{equation}\label{eq:bottlenecked_RF}
        E_g = \frac{\kappa_2 \tf_1}{1- N_{\ell}/P} + \frac{N_\ell /P}{1-N_\ell/P} \sigma_\epsilon^2.
    \end{equation}
    The structure of the features $\F$ only effects the signal term. The noise term is universal and depends only on the narrowest width $N_\ell$. 
    
    \item Both $\kappa_1$ and $\kappa_2$ are nonzero. This happens when $\mathrm{rank}(\Sh \F \F^\top) < \mathrm{rank}(\Sh \F \F^\top) \leq \mathrm{rank}(\S)$. This means that $\Sh$ has rank less than $\F \F^\top$. We have $P < D, N_\ell$. This is the \textbf{overparameterized} setting. 

    In order for $\kappa_1$ to be nonzero we must have a pole in $S_{\W}(t)$, so  $\df_1 = P/D$. This implies
    \begin{equation}
        \frac{\gamma_1}{1-\gamma_1} = \frac{\df_1}{\df_1 - \df_2} \frac{d \log \kappa_1}{d \log \kappa_2} - 1 = \frac{\df_2}{\df_1 - \df_2} + \frac{\df_1}{\df_1 - \df_2} \left(\frac{d \log \kappa_1}{d \log \kappa_2} - 1 \right).
    \end{equation}
    Using equation \eqref{eq:S_RF} and \eqref{eq:df1df2_rel3} we can write:
    \begin{equation}\label{eq:dlogs_to_S}
        \frac{d \log \kappa_1}{d \log \kappa_2} = 1 - \frac{d \log S_{\F \F^\top}(-\df_1(\kappa_2))}{d \log \kappa_2} = 1 +  \frac{\df_1 - \df_2}{\df_1} \frac{d \log S_{\F \F^\top}(-\df_1)}{d \log \df_1}.
    \end{equation}
    Defining $\gamma_2 \equiv \frac{D}{P} \df_2(\kappa_2)$ and using shorthand $S = S_{\F \F^\top}(-\df_1)$ yields:
    \begin{equation}\label{eq:overparam_RF}
    \begin{aligned}
        E_g =   -\frac{\kappa_2^2 \tf_1'(\kappa_2)}{1-\gamma_2}  + \kappa_2 \tf_1 \frac{d \log S}{d \log \df_1} 
         + \sigma_\epsilon^2 \left[\frac{\gamma_2}{1-\gamma_2} + \frac{d \log S}{d \log \df_1}  \right].
    \end{aligned}
    \end{equation}
    
\end{enumerate}

\subsection{Examples}\label{sec:lrf_examples}

In this subsection we will apply the formulas \eqref{eq:underparam_RF}, \eqref{eq:bottlenecked_RF}, and \eqref{eq:overparam_RF} to obtain the generalization error of many of the linear random feature models studied in the literature. We will consider both shallow and deep random feature models with varying amounts of structure in the data and features. 

\subsubsection{1-Layer White Random Feature Model}

We consider the simple case of $\S = \mathbf I$ and unstructured features $\F$. That is, $\F^\top \F$ is distributed as a white Wishart. We then have that $\S_{\F} = \F \F^\top$ is distributed as a White Wishart Gram matrix. The $S$-transform was computed in Equation \eqref{eq:S_gram} and is given by
\begin{equation}
    S_{\F \F^\top} = \frac{1}{\frac{N}{D} - \df_1}.
\end{equation}
As a consequence we get:
\begin{equation}
    \df^1_{\S_{\F}}(\kappa_1) = \df^1_{\S}(\kappa_2) = \frac{1}{1 + \kappa_2},
\end{equation}
\begin{equation}\label{eq:renorm_1RF}
    \kappa_2 = \frac{\kappa_1}{\frac{N}{D} - \frac{1}{1+\kappa_2}} = \frac{\lambda}{(\frac{N}{D} - \frac{1}{1+\kappa_2}) ( 1 - \frac{P}{N} \frac{1}{1+\kappa_2})},
\end{equation}
\begin{equation}
    \frac{d \log S}{d \log \df_1} = \frac{\df_1}{N/D - \df_1}.
\end{equation}
We see that at finite ridge, solving for $\kappa_2$ in terms of $\lambda$ will involve solving a cubic equation, as noted by \citet{rocks2022bias}.

We now consider the generalization performance in the ridgeless limit $\lambda \to 0$. When we take this limit, we see that either $\kappa_2 \to 0$ or $\kappa_2$ lands on one of the poles of equation \eqref{eq:renorm_1RF}.  The possible values of $\kappa_1, \kappa_2$ as $\lambda \to 0$ are:
\begin{enumerate}
    \item Underparameterized regime: $\lambda = \kappa_1 = \kappa_2 = 0$, $\df_1 = 1$.
    \item Bottlenecked regime: $\lambda = \kappa_1 = 0, \kappa_2 = \frac{D}{N} - 1$, $\df_1 = N/D$.
    \item Overparameterized regime: $\lambda = 0, \kappa_1 \neq 0, \kappa_2 = \frac{D}{P} - 1$,  $\df_1 = P/D$, $\df_2 = (P/D)^2$.
\end{enumerate}
Further, because of the isotropy of the problem, we see that  $\tf_1 = \df_1$ for any value of $\bar \w$. 
 This gives a generalization error of
\begin{equation}
    E_g = \begin{cases}
        \displaystyle\frac{D/P}{1 - D/P} \sigma_\epsilon^2, &\quad P > D, N\\
        \displaystyle\frac{1 - N/D}{1-N/P} + \frac{N/P}{1 - N/P} \sigma_\epsilon^2, &\quad N < \min(P, D)\\
        \displaystyle\left(1 - \frac PD \right) \left(1  + \frac{P/N}{1-P/N} \right) + \left( \frac{P/D}{1-P/D} + \frac{P/N}{1-P/N} \right) \sigma_\epsilon^2, &\quad P < D, N.
    \end{cases}
\end{equation}

\begin{figure}[t]
    \centering
    \includegraphics[width=4in]{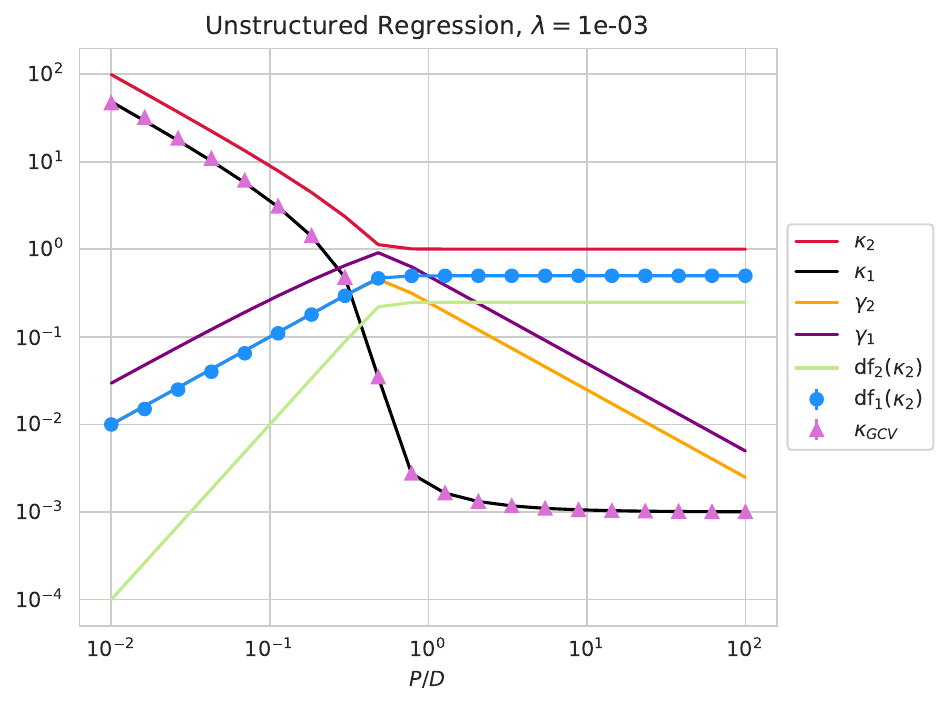}
    \includegraphics[width=4in]{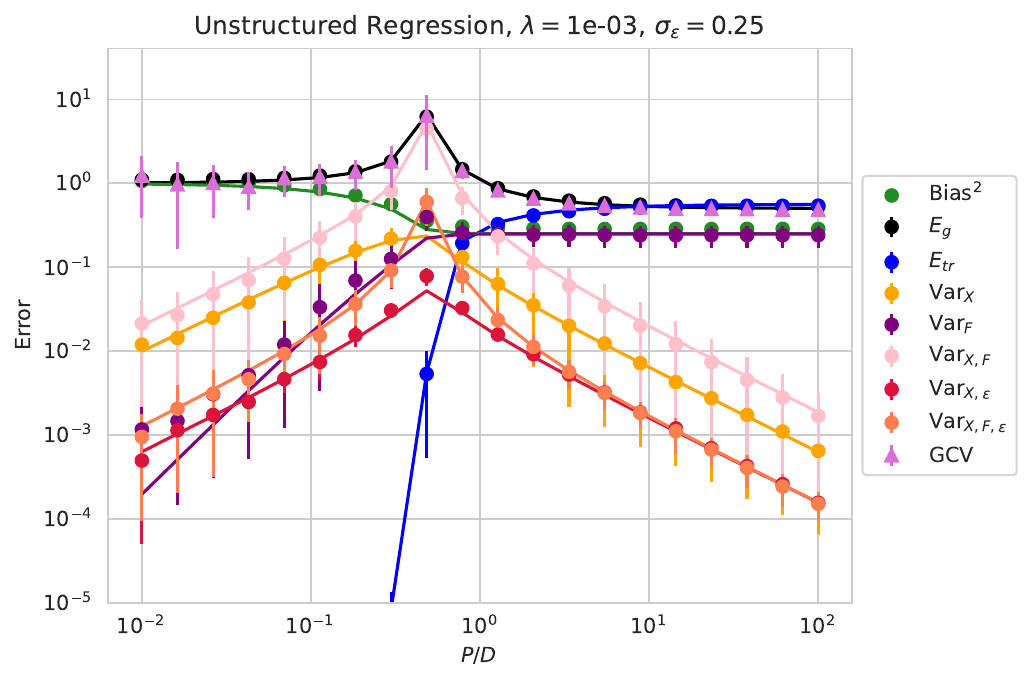}
    \caption{1-layer linear random features with unstructured covariates, \textit{i.e.} $\S = \mathbf I$. Left: We plot theory (solid lines) for the various quantities of interest: $\kappa_1, \kappa_2, \gamma_1, \gamma_2$ as well as $\df_{\S}^1(\kappa_2), \df_{\S}^2(\kappa_2)$. We also plot the estimate of $\kappa_1$ using the training set and find excellent agreement. Right: We plot the training and generalization (blue, black respectively) as well as the bias (green) and variances (orange, purple, pink, red, coral) due to all relevant quantities in the regression. Dots and error bars indicate empirical simulations over 25 seeds over training set and 25 seeds over random feature initializations. Solid curves show theory. We see strong agreement for all relevant quantities. The GCV estimator is plotted as orchid triangles and again we find excellent agreement with the generalization error.   }
    \label{fig:unstructured_LRF_linspace}
\end{figure}

\subsubsection{Deep White Random Feature Model}

We now consider the setting where the random features $\F$ consist of a composition of $L$ layers of random unstructured linear transformations. Regression with this model is analogous to regression with the final layer weights of a deep linear network at initialization. Writing $D = N_0$, we will take things to be normalized so that 
\begin{equation}
    \F^\top \F = \F_1^\top \cdots \F_L^\top \F_L \cdots \F_1 , \quad \mathbb E \left[ \F_\ell^\top \F_\ell \right] = \mathbf I.
\end{equation}
This as exactly an element of the deep product Wishart ensemble. We calculated the $S$-transform of $ \F^\top \F$ and $\F \F^\top$ in Equations \eqref{eq:deep_white_wish} and \eqref{eq:deep_white_gram} of Section \ref{sec:deep_white_wish}. The latter will be more useful to us:
\begin{equation}
    S_{\F \F^\top} = \prod_{\ell=1}^L \frac{1}{\frac{N_\ell}{D} - \df_1}.
\end{equation}
This directly yields the self-consistent equation for $\kappa_2$:
\begin{equation}
    \df^1_{\S_{\F}}(\kappa_1) = \df^1_{\S}(\kappa_2) = \frac{1}{1 + \kappa_2},
\end{equation}
\begin{equation}\label{eq:renorm_DRF}
    \kappa_2 \prod_{\ell=1}^L \left( \frac{N_\ell}{D} - \frac{1}{1 + \kappa_2} \right) = \kappa_1,
\end{equation}
\begin{equation}
    \frac{d \log S}{d \log \df_1} = \sum_{\ell =1}^L\frac{\df_1}{N_\ell/D - \df_1}.    
\end{equation}
Now as $\lambda = 0$ we see that the number of poles expands to one at every layer of the random features. Writing $N_0 = D$, the final generalization error is then:
\begin{equation}
    E_g = \begin{cases}
        \displaystyle\frac{D/P}{1 - D/P} \sigma_\epsilon^2, &\quad P > D, N_\ell \; \forall \ell\\
        \displaystyle\frac{1 - N_\ell/D}{1-N_\ell/P} + \frac{N_\ell/P}{1 - N_\ell/P} \sigma_\epsilon^2, &\quad N_\ell < \min(P, D)\\
        \displaystyle\left(1 - \frac PD \right) \left(1  + \sum_{\ell=1}^L \frac{P/N_\ell}{1-P/N_\ell} \right) + \sum_{\ell=0}^L \frac{P/N_\ell}{1-P/N_\ell} \sigma_\epsilon^2, &\quad P < D, N_\ell\; \forall \ell.
    \end{cases}
\end{equation}
This recovers the results obtained in prior works by the second and third authors using the replica trick \cite{zavatone2022contrasting, zavatone2023learning}.

\subsubsection{1-Layer Structured Random Feature Model}

We now consider the setting where $\F$ are still unstructured and shallow so that $\F^\top \F$ is distributed as a white Wishart with parameter $N/D$, and conseqently $\F \F^\top$ is distributed as a white Wishart Gram matrix. Here, the inputs $\x_\mu$ are now drawn from a structured distribution with covariance $\S$. 

We return to the shorthand $\df_1 = \df^1_{\S}(\kappa_2) \simeq \df^1_{\S_{\F}}(\kappa_1) , \df_2 = \df^2_{\S}(\kappa_2)$. Then:
\begin{equation}\label{eq:renorm_1CRF}
    \kappa_2 = \frac{\kappa_1}{\frac{N}{D} - \df_1} = \frac{\lambda}{(\frac{N}{D} - \df_1) ( 1 - \frac{D}{P} \df_1)},
\end{equation}
\begin{equation}
    \frac{d \log S_{\F \F^\top}}{d \log \df_1} =\frac{\df_1}{N/D - \df_1}.
\end{equation}
Applying Equations \eqref{eq:underparam_RF}, \eqref{eq:bottlenecked_RF}, and \eqref{eq:overparam_RF} gives the generalization error in terms of the degrees of freedom of $\S$. 

\begin{equation}
    E_g = \begin{cases}
        \displaystyle \frac{D/P}{1 - D/P} \sigma_\epsilon^2, & D < P, N\\
        \displaystyle\frac{\kappa_2 \tf_1}{1-N/P} + \frac{N/P}{1 - N/P} \sigma_\epsilon^2, & N < P,D\\
        \displaystyle-\frac{\kappa_2^2 \tf_1'}{1 - \frac DP \df_2}  + \frac{\kappa_2 \tf_1 \, P/N}{1-P/N}  + \sigma_\epsilon^2 \left[\frac{\frac{D}{P} \df_2}{1 - \frac{D}{P} \df_2} + \frac{P/N}{1-P/N} \right], & P < D, N.
    \end{cases}
\end{equation}
These are the same equations as obtained by \citet{bach2024high}. When averaging over $\bar \w$ we recover the equations of \citet{maloney2022solvable}.

\subsubsection{Orthogonal Projections of Structured Covariates}
We now let $\x_\mu$ be taken from a structured distribution with covariance $\S$. We take $\F$ to be a projection to an $N$ dimensional space with $N < D$ so that $\F \F^\top = \P \in \mathbb R^{D \times D}$ is a square projection. Then, using the $S$-transform for square projections calculated in Equation \eqref{eq:S_square_proj}, we have
\begin{equation}
    \kappa_2 = \kappa_1 \frac{1-\df_1}{\frac{N}{D} - \df_1},
\end{equation}
\begin{equation}
    \frac{d \log S}{d \log \df_1} = \frac{\df_1}{\frac{N}{D} - \df_1} - \frac{\df_1}{1 - \df_1}.
\end{equation}
Note that $\kappa_2$ is not renormalized as strongly as in the case of a Wishart. Intuitively, a matrix with random Gaussian entries projecting down to $N < D$ dimensions not only projects, but also adds noise. This leads to a larger renormalization relative to the case of a simple projection. 

We now evaluate the ridgeless limit. There is no underparameterized case. Applying Equations \eqref{eq:bottlenecked_RF}, and \eqref{eq:overparam_RF} gives:
\begin{equation}
    E_g = \begin{cases}
        \displaystyle\frac{\kappa_2 \tf_1}{1-N/P} + \frac{N/P}{1 - N/P} \sigma_\epsilon^2, & N < P, D\\
        \displaystyle\begin{aligned}
                    &-\frac{\kappa_2^2 \tf'}{1 - \frac DP \df_2}  + \kappa_2 \tf_1 \frac{1-N/D}{1-P/D} \frac{P/N}{1-P/N}  \\
                    & \quad + \sigma_\epsilon^2 \left[\frac{\frac{D}{P} \df_2}{1 - \frac{D}{P} \df_2} + \frac{1-N/D}{1-P/D} \frac{P/N}{1-P/N} \right]
        \end{aligned}
        & P <  N, D.
    \end{cases}
\end{equation}
When $N = D$ this recovers the results for linear regression. To our knowledge, this result has not been explicitly obtained in past works. 

\subsubsection{Deep Structured Random Feature Model}\label{sec:deep_structured_lrf}

We now generalize the previous example to the case of several layers of random features, each of which has nontrivial structure in its covariance. That is, we take
\begin{equation}
    \F^\top \F = \F_1^\top \cdots \F_L^\top \F_L \cdots \F_1 , \quad \mathbb E \left[\F_\ell^\top \F_\ell \right] = \S_\ell.
\end{equation}
The $S$-transform we will need is that evaluated for the Gram matrix of a deep structured Wishart product. This has been computed in Equation \eqref{eq:deep_structured_gram} of Section \ref{sec:deep_structured_wish}. Taking the shorthand $\df_1 = \df^1_{\S}, \df_2 = \df^2_{\S}$, we again have:
\begin{equation}
    \df^1_{\S^D_{\F}}(\kappa_1) = \df_1(\kappa_2),
\end{equation}
\begin{equation}\label{eq:renorm_DCRF}
    \kappa_2 \prod_{\ell=1}^L (-\df_1) \zeta_{\S_\ell}\left(-\frac{D}{N_\ell} \df_1\right) = \kappa_1, %
\end{equation}
\begin{equation}\label{eq:S_LRF_deep_structured}
\begin{aligned}
  \frac{d \log S}{d \log \df_1} &=  \sum_\ell \left(- 1 + \frac{D}{N_\ell} \df_1 \frac{\zeta_{\S_\ell}'(-\frac{D}{N_\ell} \df_1)}{\zeta_{\S_\ell}(-\frac{D}{N_\ell} \df_1)} \right) \\
  &= \sum_\ell \left(-1 - \frac{\frac{D}{N_\ell}  \df_1}{\kappa_\ell {\df_{\S_\ell}^1}' (\kappa_\ell)} \right), \quad \kappa_\ell \equiv -\zeta_{\S_\ell}(-\tfrac{D}{N_\ell} \df_1)\\
  &=  \sum_{\ell=1}^L \frac{\df_{\S_\ell}^2(\kappa_\ell)}{\df_{\S_\ell}^1(\kappa_\ell)- \df_{\S_\ell}^2(\kappa_\ell)}.
\end{aligned}
\end{equation}
In the last line we have used the fact that $\df_{\S_\ell}^1(\kappa_\ell) = \frac{D}{N_\ell} \df_1$ and applied Equation \eqref{eq:df1df2_rel2}. In the ridgeless limit $\df_{\S_\ell}^1 = P/N_\ell$. Adopting the notation $\gamma^{(\ell)} \equiv \frac{N_\ell}{P} \df_{\S_\ell}^2$, $\gamma^{(0)} \equiv \frac{D}{P} \df_2 = \gamma_2$ gives the formula for the generalization error:

\begin{equation}\label{eq:Eg_LRF_deep_structured}
    E_g = \begin{cases}
        \displaystyle\frac{D/P}{1-D/P} \sigma_\epsilon^2, & P > D, \{N_\ell\}_{\ell=1}^L\\
        \displaystyle\frac{\kappa_2 \tf_1}{1- N_\ell/P} + \sigma_\epsilon^2 \frac{N_\ell/P}{1-N_\ell/P} & N_\ell < D, P, \{N_\ell'\}_{\ell' \neq \ell}\\
        \displaystyle - \frac{\kappa_2^2 \tf_1' }{1 - \gamma^{(0)}} + \kappa_2 \tf_1 \sum_{\ell=1}^L \frac{\gamma^{(\ell)}}{1 - \gamma^{(\ell)}}    + \sigma_\epsilon^2 \sum_{\ell=0}^L \frac{\gamma^{(\ell)}}{1-\gamma^{(\ell)}} & P < D, \{N_\ell\}_{\ell=1}^L.
    \end{cases}
\end{equation}
This is the same result as  obtained in \citet{zavatone2023learning} using replica theory.
Lastly, taking Equation \eqref{eq:renorm_DCRF} and \eqref{eq:S_LRF_deep_structured} plugging in to Equation \eqref{eq:LRF_finite_ridge} gives the finite ridge result quoted in \citet{zavatone2023learning}.

\subsection{Training Error}\label{sec:RF_KARE}

One can also compute the training error as in Section \ref{sec:KARE}, yielding
\begin{equation}
\begin{aligned}
    E_{tr}
    &= \frac{\lambda^2}{P} \bar \w^\top \X^\top (\tfrac1{P} \X \F \F^\top \X^\top + \lambda \mathbf I)^{-2} \X \bar \w + \sigma_\epsilon^2 \lambda^2  \tr\left[ (\tfrac1{P} \X \F \F^\top \X^\top + \lambda\mathbf I)^{-2} \right] \\
    &= -\lambda^2 \partial_\lambda \bar \w^\top \hat \S (\F \F^\top \hat \S + \lambda \mathbf I)^{-1} \bar \w - \sigma_\epsilon^2 \lambda^2 \partial_{\lambda} \underbrace{\left[ \frac{1 - \frac{D}{P} \df_{\S^D_{\F}}^1(\kappa_1)}{\lambda} \right]}_{1/\kappa_1} \\
    &\simeq -\frac{\lambda^2}{1-\gamma_1} \tf_1'(\kappa_1)  + \frac{\sigma_\epsilon^2 \lambda^2/\kappa_1^2}{1-\gamma_1}   = \frac{\lambda^2}{\kappa_1^2} (E_g + \sigma_\epsilon^2).
\end{aligned}
\end{equation}
Thus we see that the analogue of the KARE (\textit{i.e.}, the GCV estimator) is given by multiplying the training error by $S_{\W}(-\df_1)^2$. This is also asymptotically equal to $S_{\W} ( -\df_{\hat \S_{\F}}(\lambda))^2$, which can be calculated from the training data alone.

\subsection{Implicit Regularization of Ensembles}\label{sec:lrf_ensembles}

Consider the taking $E$ different sets of random features $\F_{e}$ all drawn from the same distribution. On each independent ensemble, one runs regression with ridge $\lambda$ and obtains $\hat \w_e$. Taking the average of all of these gives the \textbf{ensembled} predictor:
\begin{equation}
    \hat \w_{E} = \frac{1}{E} \sum_{e} \hat \w_{e}.
\end{equation}
Similar to how in Section \ref{sec:bias_var_kern} we saw that bagging reduces the variance over $\X, \e$ by a factor of $1/B$, ensembling reduces the variance over the features $\F$ by $1/E$. For a large ensemble of models, we can ask what $\lim_{E \to \infty} \hat \w_{E}$ converges to. Applying the deterministic equivalent \eqref{eqn:t_tsfm_sub_no_a} to the features, this becomes:
\begin{equation}\label{eq:LRF_ensemble}
\begin{aligned}
    \mathbb E_{\F} \hat{\w} &= \mathbb E_{\F} (\F \F^\top \X^\top \X  + P \lambda \mathbf I)^{-1} \F \F^\top \X^\top  \y\\
    &\simeq   \left(\X^\top \X + P  \lambda S_{\F \F^\top} \mathbf I \right)^{-1} \X^\top \y.
\end{aligned}
\end{equation}
This is just ridge regression in the original input space $\mathbb R^{D}$ but with the ridge $\lambda$ renormalized to $ \lambda S_{\F \F^\top} = \lambda \kappa_2/\kappa_1 $. In the case where $\F \F^\top$ is a projection, this was obtained in \citet{lejeune2020implicit,patil2024asymptotically,yao2021minipatch}. Our results hold for any features $\F$ such that $\F \F^\top$ is free of $\Sh$, as in \citet{patil2024asymptotically}. 

\subsection{Fine-Grained Bias-Variance Decomposition}\label{sec:bias_var_LRF}

\begin{figure}[t]
    \centering
    \includegraphics[width=4in]{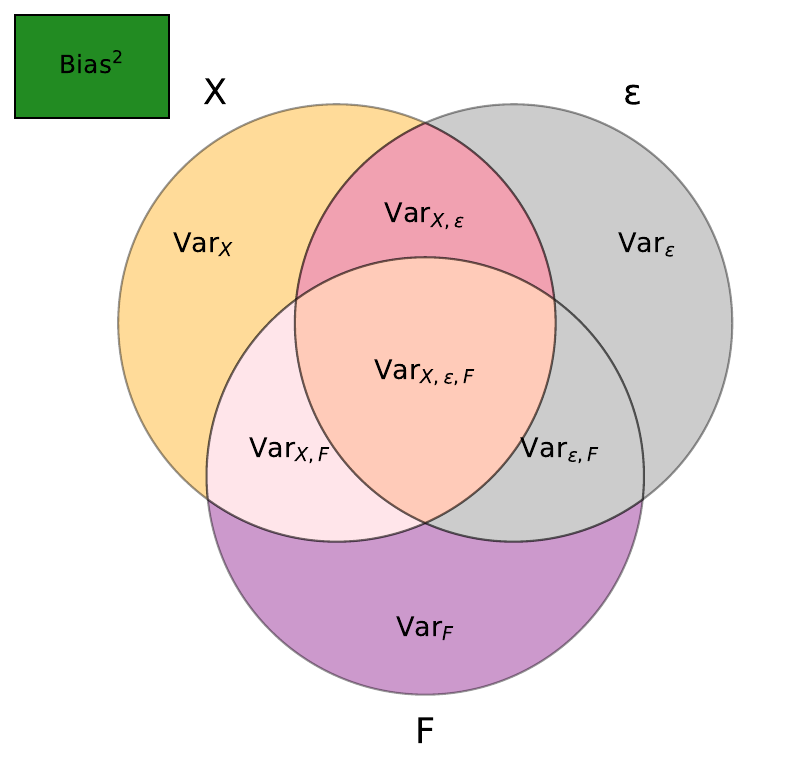}
    \caption{Schematic of the bias-variance decomposition for linear random features, as in \citet{adlam2020understanding}. The color scheme matches the plots in Figures \ref{fig:unstructured_LRF_linspace} and \ref{fig:deep_LRF_OV}. Grey regions do not contribute to variance. }
    \label{fig:LRF_BV}
\end{figure}

Extending the results of Sections \ref{sec:bias_var_kern} and \ref{sec:lrf_ensembles}, we consider averaging the learned weights $\bar \w$ over three sources of variance for a general feature map $\F$. The three sources are the choice of training set $\X$, the label noise $\bm \epsilon$, and the random features $\F$. We have 
\begin{equation}
\begin{aligned}
    \mathbb E_{\X, \bm \epsilon, \F} \hat{\w} &= \mathbb E_{\X, \F} \F (\F^\top \X^\top \X \F + P \lambda \mathbf I)^{-1} \F^\top \X^\top \X \bar \w \\
    &= \mathbb E_{\F}  \F \F^\top  (\S  \F \F^\top + \kappa_1 \mathbf I )^{-1} \S  \bar \w \\
    &= \S  (\S + \kappa_2 \mathbf I )^{-1} \bar \w .
\end{aligned}
\end{equation}
This yields that:
\begin{equation}
    \mathrm{Bias}^2 = E_g(\mathbb E_{\X, \bm \epsilon, \F} \hat{\w}) = \kappa_2^2\, \bar \w^\top \S  (\S + \kappa_2 \mathbf I )^{-2} \bar \w =  -\kappa_2^2 \tf_1'.
\end{equation}
The variance term is similarly decomposable into contributions from the various combinations of $\X, \bm \epsilon$, and  $\F$ as in the works of \citet{adlam2020understanding} and \citet{lin2021anova}. We sketch this in Figure \ref{fig:LRF_BV}.  
We can explicitly get $\mathrm{Var}_{\X}, \mathrm{Var}_{\X, \e}, \mathrm{Var}_{\e}$ by considering. $\mathbb E_{\F} \hat{\w}$. This was seen to be equivalent to ridge regression with a rescaled ridge $\lambda S_{\F \F^\top}$ in Equation \eqref{eq:LRF_ensemble}. This ridge will be further renormalized to $\kappa_2$ in the final deterministic expression for the generalization error. Thus, the bias-variance results of Section \ref{sec:bias_var_kern} apply with $\kappa=\kappa_2, \gamma=\gamma_2$ and we get:
\begin{equation}
   \mathrm{Var}_{\X} = -\frac{\gamma_2}{1-\gamma_2} \kappa_2^2 \tf_1', \quad \mathrm{Var}_{\X, \e} = \frac{\gamma_2}{1-\gamma_2} \sigma_\epsilon^2, \quad \mathrm{Var}_{\e} = 0.
\end{equation}
One can similarly compute $\mathrm{Var}_{\F}, \mathrm{Var}_{\F, \e}$ by instead averaging the estimator $\hat{\w}$ over $\X$:
\begin{equation}
\begin{aligned}
    \mathbb E_{\X} \hat{\w} &= \mathbb E_{\X} \F (\F^\top \X^\top \X \F + P \lambda \mathbf I)^{-1} \F^\top \X^\top (\X \bar \w  + \e)\\
    &=   \F \F^\top \S  (\F \F^\top \S + \kappa_1 \mathbf I )^{-1} \bar \w. \\
\end{aligned}
\end{equation}
This gives that $\mathrm{Var}_{\F, \e} = 0$. The generalization error is then averaged over $\F$ to yield:
\begin{equation}
\begin{aligned}
      E_{g}(\mathbb E_{\X} \hat{\w}) = (\bar \w - \mathbb E_{\X} \hat{\w})^\top \S (\bar \w - \mathbb E_{\X} \hat{\w}) &= \kappa_1^2 \bar \w^\top \S (\F \F^\top \S + \kappa_1 \mathbf I )^{-2} \bar \w.
\end{aligned}
\end{equation}
This gives $\mathrm{Var}_{\F}$ via:
\begin{equation}
    \mathrm{Bias}^2 + \mathrm{Var}_{\F} =  E_g(\mathbb E_{\X} \hat{\w}) = -\kappa_1^2 \partial_{\kappa_1} \widetilde{\tf}_1(\kappa_1) \Rightarrow \mathrm{Var}_{\F} = \left(1 - \tfrac{d \log\kappa_2}{d \log \kappa_1} \right) \kappa_2 \tf_2(\kappa_2).
t\end{equation}
The joint variance $\mathrm{Var}_{\X, \F}$ is then given by the subtraction.
\begin{equation}
    \mathrm{Var}_{\X, \F} = \frac{\gamma_1}{1-\gamma_1} [- \kappa_1^2  \widetilde{\tf}_1'(\kappa_1)] - \frac{\gamma_2}{1-\gamma_2}[-\kappa_2^2 \tf_1'(\kappa_2)]
\end{equation}
Finally, we get that all the variance due to $\e$ is in $\mathrm{Var}_{\X, \e}, \mathrm{Var}_{\X, \F, \e}$, with:
\begin{equation}
    \mathrm{Var}_{\X, \F, \e} = \sigma_\epsilon^2 \left[ \frac{\gamma_1}{1 - \gamma_1} - \frac{\gamma_2}{1-\gamma_2} \right].
\end{equation}
All these terms are graphically presented in Figure \ref{fig:LRF_BV}.  The expressions are consistent with what \citet{adlam2020understanding} find in the setting of random feature models on isotropic data.

\subsubsection{Overparameterized Case}
We can decompose the full deep structured random feature model generalization error into bias and variance terms as follows:

\begin{equation}\label{eq:LRF_fine_grained_overparam}
    \underbrace{- \kappa_2^2 \tf_1'}_{\text{Bias}^2} \underbrace{- \kappa_2^2 \tf_1' \frac{\gamma^{(0)}}{1 - \gamma^{(0)}}}_{\text{Var}_{\X}}  + \underbrace{\kappa_2 \tf_1   \sum_{\ell=1}^L \frac{\gamma^{(\ell)}}{1 - \gamma^{(\ell)}}}_{\text{Var}_{\F} + \text{Var}_{\X, \F} }  + \underbrace{\frac{\gamma^{(0)}}{1 - \gamma^{(0)}}}_{\text{Var}_{\X, \e}} +  \underbrace{\sigma_\epsilon^2 \sum_{\ell=1}^L \frac{\gamma^{(\ell)}}{1-\gamma^{(\ell)}}}_{\text{Var}_{\X, \F, \e}}.
\end{equation}
\begin{equation}
\begin{aligned}
    \mathrm{Var}_{\F}
    &= \kappa_2 \tf_2 \frac{1}{1 + \sum_{\ell=0}^L \frac{\gamma^{(\ell)}}{1-\gamma^{(\ell)}}}.
\end{aligned}
\end{equation}
The remaining term is:
\begin{equation}
    \mathrm{Var}_{\X,\F} = \kappa_2 \tf_1 \sum_{\ell=1}^L \frac{\gamma^{(\ell)}}{1-\gamma^{(\ell)}} -  \frac{\kappa_2 \tf_2}{1 + \sum_{\ell=0}^L \frac{\gamma^{(\ell)}}{1-\gamma^{(\ell)}}}.
\end{equation}
Note that the model-wise double descent peak that occurs when any of the $\gamma^{\ell} = 1$ for $\ell \geq 1$ is due entirely to the variances $\mathrm{Var}_{\X, \e}, \mathrm{Var}_{\X, \F, \e}$. The sample-wise double descent peak on the other hand is due to only to $\mathrm{Var}_{\X}, \mathrm{Var}_{\X, \epsilon}$.

\subsubsection{Bottlenecked Case}
Noting $\frac{d \log \kappa_2}{d \log \kappa_1} = 0$, we get:
\begin{equation}
    \mathrm{Bias}^2 = -\kappa_2^2 \tf_1' \quad \mathrm{Var}_{\X} = \frac{\gamma^{(\ell)}}{1 - \gamma^{(\ell)}} [-\kappa_2^2 \tf_1'], \quad 
    \mathrm{Var}_{\F} = \kappa_2 \tf_2.
\end{equation}
Here $\gamma^{(\ell)} = N_\ell/P$. The remaining term in the variance is then
\begin{equation}
     \mathrm{Var}_{\X, \F} = \frac{\gamma^{(\ell)}}{1 - \gamma^{(\ell)}} \kappa_2 \tf_2.
\end{equation}
\subsubsection{Underparameterized Case}
Because $E_g$ depends only on $\sigma_\epsilon^2$ we have that all of $\mathrm{Var}_{\X}, \mathrm{Var}_{\F}, \mathrm{Var}_{\X, \F}$ vanish. The only nontrivial variance is the noise term, $\mathrm{Var}_{\X, \e}$.

\subsection{\texorpdfstring{Scaling Laws in $P$ and $N$}{Scaling Laws in P and N}} \label{sec:lrf_scaling}

As in the kernel setting, we take $\sigma_\epsilon^2 = 0$ and study the generalization performance for power-law distributed data $\eta_k \sim k^{-\alpha}$. In Section \ref{sec:lrf_teacher_avg} we will average over teachers, connecting to results of \citet{maloney2022solvable} and reproducing phenomena observed in \citet{bahri2021explaining}. In section  \ref{sec:lrf_general_teacher} we do not average over $\bar \w$ and instead take $\bar w_k^2 \eta_k \sim k^{-(1+2 r \alpha)}$.  We get a refinement of the scaling laws and observe different exponents in the overparameterized and underparameterized regime. As in Section \ref{sec:kernel_scaling}, $\alpha, r$ are the capacity and source exponents respectively. 

In Section \ref{sec:LRF_varlim}, we find a new scaling law in the overparameterized regime where finite width effects change the leading order scaling behavior and hurt generalization without fully bottlenecking the model. This is related to the variance-dominated behavior studied by the first and third authors with colleagues in \citet{atanasov2022onset}. 

\subsubsection{Target-Averaged Results} \label{sec:lrf_teacher_avg}

We can reproduce the results of \citet{maloney2022solvable} for general random feature models. There, the teacher vector $\bar \w$ was averaged over. In this case, using that $\mathbb E_{\bm w} \tf_1 = \df_1$, we get that in the zero noise limit of Equation \eqref{eq:Eg_LRF_deep_structured}:
\begin{equation}
    E_g \simeq \begin{cases}
        0, & P > D, N\\
        \displaystyle\frac{\kappa_2 \df_1}{1- N_\ell/P} & N < D, P\\
        \displaystyle \kappa_2 \df_1 \left(1 + \sum_{\ell=1}^L \frac{\gamma^{(\ell)}}{1 - \gamma^{(\ell)}} \right)   & P < D, N.
    \end{cases}
\end{equation}
Unsurprisingly, in the underparameterized setting with no noise, there is no scaling law since $\bar \w$ is recovered exactly. In order to study the scaling properties of the bottlenecked and overparameterized settings, we need to know how $\kappa_2$ scales with $N, P$ respectively. Again, this can be easily seen through Equation \eqref{eq:S_RF} defining the renormalized ridge $\kappa_2$.

In the ridgeless limit, we either have a pole in $S_{\F \F^\top}(-\df_1)$ (bottlenecked) or in $S_{\W}(-\df_1)$ (overaparameterized). Even in the most general case of deep structured random features, this happens only when $D \df_1(\kappa_2)$ scales either as $P$ or $N$, respectively. On the other hand, from Section \ref{sec:kernel_scaling}, we know that 
\begin{equation}
    D \df_1(\kappa_2) = \int_{1}^\infty \frac{k^{-\alpha}}{\kappa_2 + k^{-\alpha}} dk \sim \kappa_2^{-1/\alpha}.
\end{equation}
Thus, in order for the $S$-transforms to have a pole, we need $\kappa_2 \sim N^{-\alpha}, P^{-\alpha}$ in the bottlenecked and overaparameterized settings respectively. Then $\df_1 = N/D, P/D$ in these respective cases, giving:
\begin{equation}
    E_g \sim \begin{cases}
        \displaystyle N^{1-\alpha} \frac{1}{1- N_\ell/P} & \text{bottlenecked}\\
        \displaystyle P^{1-\alpha} \left(1 + \sum_{\ell=1}^L \frac{\gamma^{(\ell)}}{1 - \gamma^{(\ell)}} \right)  & \text{overparameterized}
    \end{cases}
\end{equation}
In the case where the covariances of the features are white, we get $\gamma^{(\ell)} = P/N_\ell$. 
At $L=1$, this formula then simplifies to 
\begin{equation}
    E_g \sim \begin{cases}
        \displaystyle N^{1-\alpha} \frac{1}{1- N_\ell/P} & \text{bottlenecked}\\
        \displaystyle P^{1-\alpha} \frac{1}{1-P/N_\ell}  & \text{overparameterized},
    \end{cases}
\end{equation}
which reproduces the main scalings found by \citet{maloney2022solvable}. One can see both resolution-limited and variance-limited scaling exponents in these expressions \cite{bahri2021explaining}. The parameter that is the bottleneck ($N, P$ respectively) has a nontrivial scaling exponent, and scaling it up will continue decreasing the loss until a double descent peak is hit. This is the resolution-limited scaling.  The non-bottleneck parameter enters only with trivial exponent, and scales only the subleading terms in the expansion of the generalization error. This is the variance-limited scaling.

We now consider the case where the weights of layer $\ell$ are drawn from an anisotropic distribution with covariance  $\S_\ell$ having eigenvalues decaying like $\eta_k \sim k^{-\alpha_\ell}$. This setting was studied in section \ref{sec:deep_structured_lrf}. In the overparameterized ridgeless limit given by Equation \eqref{eq:Eg_LRF_deep_structured}, we have by definition of $\kappa_\ell$ that $\df_{\S_\ell}^1(\kappa_\ell)=P/N_\ell$, which gives that $\kappa_\ell \sim P^{-\alpha_\ell}$ assuming $\alpha_\ell > 1$ in a normalizable spectrum. This then gives $\gamma^{(\ell)} = \frac{N_\ell}{P} \df^2_{\S_\ell} \sim O_P(1)$ independent of $P$. In the case where the spectrum of the weight matrices is not normalizable we get $\kappa_\ell \sim N_\ell^{1-\alpha_\ell}/P, \gamma^{(\ell)} \sim (P/N_\ell)^{\min(1, (1-\alpha_\ell)/\alpha_\ell)}$ as in Section \ref{sec:non_normalizable}. In the window of $1/2 < \alpha < 1$, we get that the $N_\ell$ enters with nontrivial exponent. That is, the variance-limited exponents become nontrivial if the weight spectrum is non-normalizable, contrasting with previous works that have only considered the case of normalizable or isotropic weight spectra \cite{zavatone2023learning,maloney2022solvable}. Previous empirical works on feature-learning neural networks have encountered nontrivial scaling in $N_\ell$ \cite{vyas2024feature, guth2023rainbow}. However, it is not clear whether this arises due through the mechanism described here or through data-dependent correlations between the weights at different layers. Products of strongly-correlated matrices are not amenable to easy treatment using the tools of free probability.

\subsubsection{General Targets} \label{sec:lrf_general_teacher}

We can extend this scaling analysis to general power-law structured $\bar \w$ with coefficients decaying as $\eta_k \bar w_k^2 = k^{-(1+2 \alpha r)}$ with source exponent $r$, rather than averaging over the target weights. As noted in the prior section, in the ridgeless limit we have that $\kappa_2 \sim \min(P, N)^{-\alpha}$. This yields:
\begin{equation}\label{eq:tf_scalings}
\begin{aligned}
    \kappa_2 \tf_1(\kappa_2) &\sim \int_1^\infty \frac{k^{-(1 + 2 \alpha r)}}{1 + k^{-\alpha}/\kappa_2} \sim \min(P, N)^{-2 \alpha \min(r, 1/2)},
    \\ 
    -\kappa_2^2 \tf_1'(\kappa_2) &\sim \int_1^\infty \frac{k^{-(1 + 2 \alpha r)}}{(1 + k^{-\alpha}/\kappa_2)^2} \sim \min(P, N)^{-2  \alpha \min(r, 1)}.
\end{aligned}
\end{equation}
This gives the following scalings in the bottlenecked and overparameterized regimes:
\begin{equation}
    E_g \sim \begin{cases}
        \displaystyle N^{-2  \alpha  \min(r, 1/2)} \frac{1}{1- N/P}  + \sigma_\epsilon^2 \frac{N/P}{1 - N/P} \\
        \displaystyle P^{-2  \alpha  \min(r, 1)}  + P^{- 2 \alpha \min(r, 1/2)} \sum_{\ell=1}^L \frac{\gamma^{(\ell)}}{1 - \gamma^{(\ell)}} + \sigma_\epsilon^2 \sum_{\ell=0}^L \frac{\gamma^{(\ell)}}{1-\gamma^{(\ell)}}. 
    \end{cases}
\end{equation}
The teacher-averaged results correspond to setting $1+2\alpha r = \alpha$, or equivalently $r = \frac12 \frac{\alpha-1}{\alpha}$. We see that in this setting $r < 1/2$. This uniquely determines the scalings and recovers the results of Section \ref{sec:lrf_teacher_avg}.
We consider the general non-ridgeless case  with label noise in Section \ref{sec:lrf_all_scaling}.

\subsubsection{Variance-Dominated Scaling} \label{sec:LRF_varlim}

\begin{figure}[t]
    \centering
    \includegraphics[width=4in]{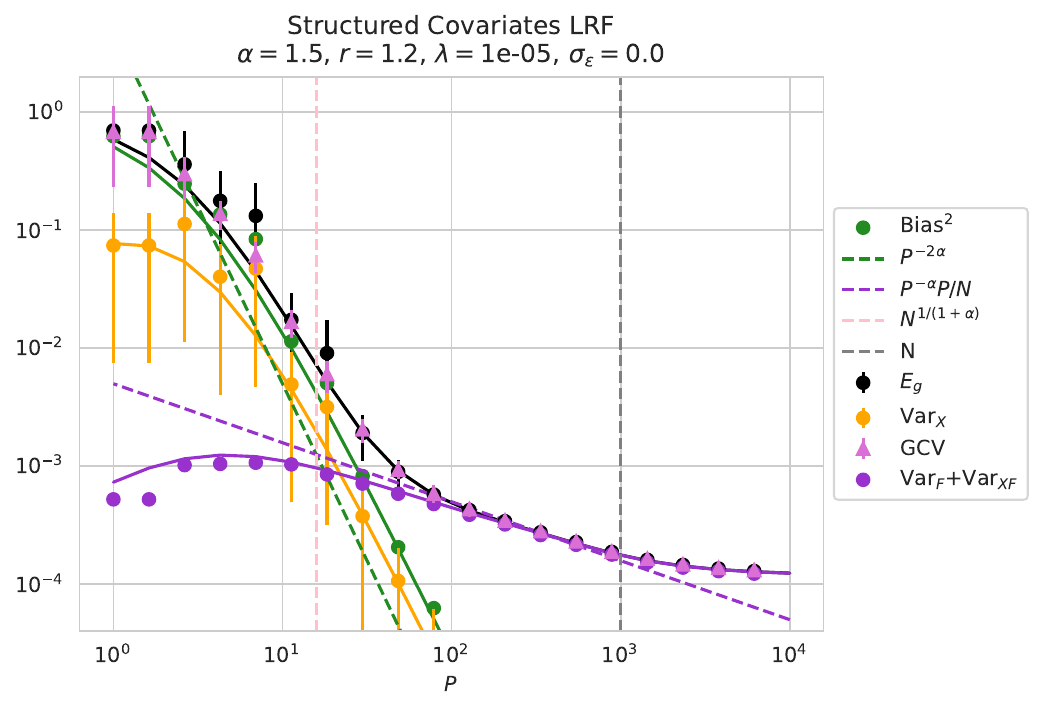}
    \includegraphics[width=4in]{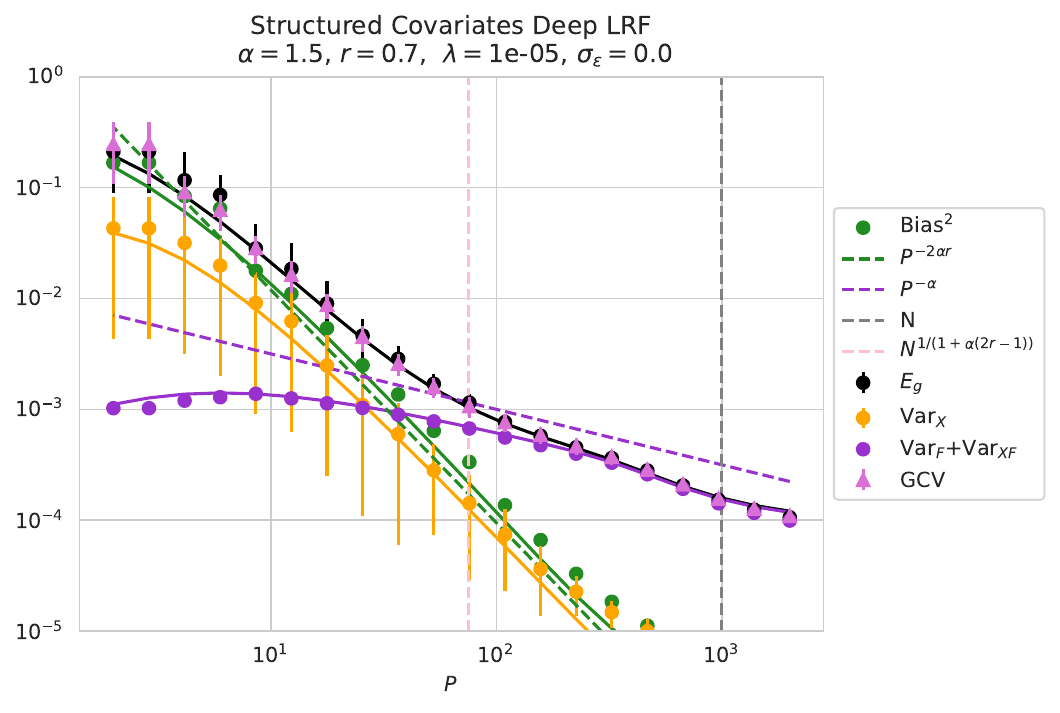}
    \caption{Left: Linear random feature model with input dimension $D=10000$ and hidden width $N=1000$ with unstructured weights but structured input. Right: Deep linear random feature model with input dimension $2000$ and two hidden layers of widths $N=1000$ and unstructured weights. The input dimension is $D=2000$. In both cases, the source exponent puts us in the regime where variance-dominated behavior can occur. Past a certain point (dashed pink), most of the limiting behavior of performance is due to variance over initializations (solid purple) which can thus be removed by ensembling. The ridge has been chosen to eliminate the double descent peak. We bag over 20 data seeds and ensemble over 20 initialization seeds.  }
    \label{fig:deep_LRF_OV}
\end{figure}

Several papers have found both theoretically and empirically that the leading order corrections of finite width in the overparameterized regime is to introduce an initialization-dependent variance that strictly hurts generalization \cite{geiger2020scaling, atanasov2022onset, bordelon2023fluctuations, zavatone2022contrasting, zv2022asymptotics}. By definition, this variance can be removed by ensembling networks over different initializations.
The authors in \citet{atanasov2022onset} also highlight that finite-width networks in the lazy regime can exhibit a large separation of scales in the overparameterized regime between the size of $P$ where this initialization-dependent variance begins to inhibit generalization and the interpolation threshold at $P = N$.  In that work, they studied a special type of nonlinear model to reproduce the behavior. Here, we show that this can happen also in linear random feature models. 

Using Equation \eqref{eq:LRF_fine_grained_overparam}, one can compute the following two terms in the overparameterized ridgeless setting:
\begin{equation}
\begin{aligned}
    \mathrm{Bias}^2 + \mathrm{Var}_{\X} &\sim P^{-2 \alpha \min(r, 1)}.\\
    \mathrm{Var}_{\F} + \mathrm{Var}_{\F, \X} &\sim  P^{-2 \alpha \min(r, 1/2)} \sum_{\ell=1}^L \frac{\gamma^{(\ell)}}{1 - \gamma^{(\ell)}}.
\end{aligned}
\end{equation}
When $\sigma_\epsilon^2 = 0$, the sum of these two terms gives the generalization error $E_g$. When over half of the generalization error is due to the variance term, we say that the scaling is \textbf{variance-dominated}. We will denote the value of $P$ where the scaling becomes variance-dominated by $P_{\F}$. In the above, if $r \leq 1/2$, then the scaling exponents of the $P$ factors in front agree. Assume for now that the features are isotropic. We have that $\gamma^{(\ell)} = P/N_{\ell}$. Consequently, we get that $P_{\F} \sim \frac{N}{1+L}$. Thus, for deep random feature models, the depth gives a linear separation between $P = P_{\F}$ and the interpolation threshold $P = N_\ell$. Unless $L$ is immense, this doesn't lead to a genuine scaling law.

We must therefore have $r >  1/2$ in order to have $\mathrm{Var}_{\F} + \mathrm{Var}_{\F \X}$ dominate $\mathrm{Bias}^2 + \mathrm{Var}_{\X}$ over an extended range of scales. The value of $P$ where this new scaling enters is at:
\begin{equation}
    P_{\F}^{-2 \alpha \min(r, 1)} \sim \frac{P_{\F}^{-(\alpha-1)}}{N} \Rightarrow P_{\F} \sim N^{\frac{1}{1 + 2 \alpha \min(r-1/2, 1/2)}}.
\end{equation}
This crossover determines when variance-dominated behavior emerges. 

The condition $r > 1/2$ has a clear interpretation in terms of the theory of kernels. Consider the $D$ dimensional input space as the reproducing kernel Hilbert space (RKHS) $\mathcal H$ of some kernel with eigenspectrum given by the eigenvalues $\eta_k$ of $\S$.  Having the target function $f(\x) = \bar \w \cdot \x$ be a normalizable element of $\mathcal H$ is equivalent to the two-norm $\|\w\|^2$ being finite. This in turn is equivalent to $r > 1/2$. Thus, if the target function is finite-norm in the original space, passing through random features can substantially hurt the scaling properties of $E_g$. 

Remaining in the overparameterized setting $P < N$, consider the case where a given $\S_\ell$ is anisotropic, with power law structure. That is, the eigenvalues of $\S_{\ell}$ decay as $k^{-\alpha_\ell}$. Then by the same analysis as in Section \ref{sec:non_normalizable}, we have that $\gamma^{(\ell)}$ scales as $(P/N_\ell)^{c_\ell}$ where $c_{\ell} = \min(0, \max(1, \frac{1-\alpha_\ell}{\alpha_\ell}))$. 
There, as long as $r > 1/2$, 
\begin{equation}\label{eq:OV_defn}
    P_{\F} \sim N^{\frac{c}{c + 2 \alpha \min(r-1/2, 1/2)}}.
\end{equation}
In particular when $r > 1/2$ and $\alpha_\ell \geq 1$ we get that this term always dominates. 

\subsubsection{Effects of Weight Structure}

\begin{figure}[t]
    \centering
    \includegraphics[scale=0.6]{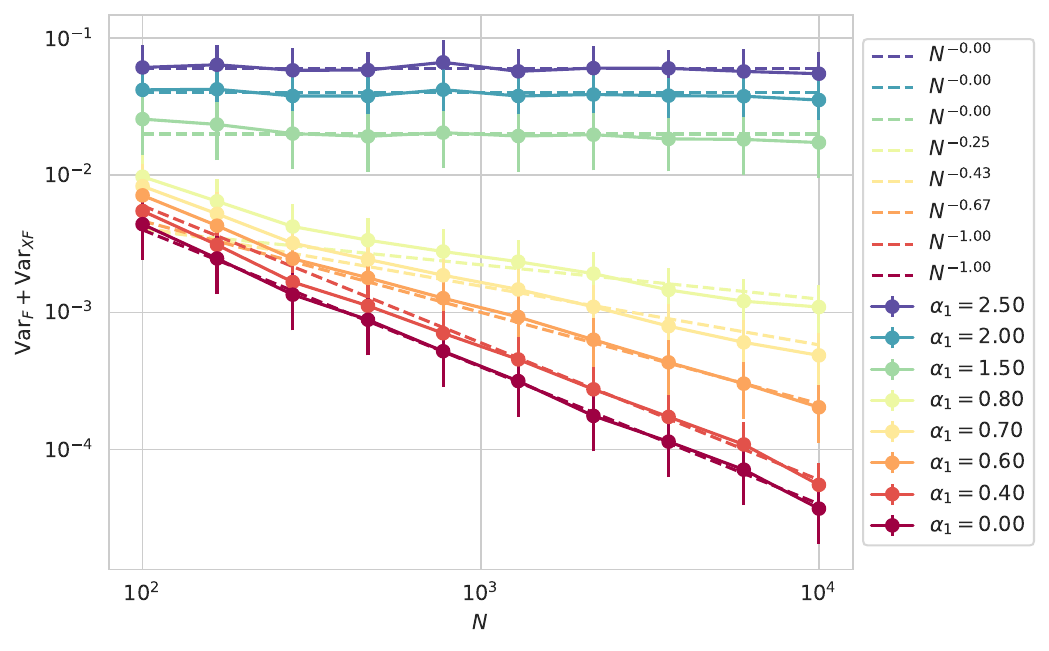}
    \caption{Scaling of finite $N$ corrections to a shallow linear random feature model when $P = 10, D=10000$. Dashed lines are pure power laws. For $0.5 < \alpha_\ell < 1$ one observes nontrivial scaling laws with the width. For $\alpha_\ell > 1$ one observes a constant scaling, and taking the infinite width limit does not get rid of $\mathrm{Var}_{\F}$. For $\alpha_\ell > 1$ the finite asymptotic values scale as $\alpha_\ell-1$, and are shown as dashed lines.}
    \label{fig:finite_width_scaling}
\end{figure}

In \citet{zavatone2023learning}, the second and third authors analyzed deep linear random feature models with structured Gaussian weights, showing that adding structure to weights generally hurts generalization. There, using the fact that each structured Gaussian can be interpreted as a product of an unstructured Gaussian matrix with the fixed weight covariance, this effect was interpreted in terms of the rotation-invariance of the unstructured Gaussian factors: there are no preferred directions into which variance in the weights should be shunted, so structure should not be beneficial. When studying scaling properties, \citet{zavatone2023learning} only considered the case of normalizable weight spectra $\alpha_\ell > 1$. 

Here, we offer a refined interpretation of why weight structure is harmful in terms of source-capacity conditions. Large exponents $\alpha_\ell$ yield rapidly-decaying weight spectra. This reduces the effective dimensionality of the hidden layers and limits the ability of signals to propagate through this channel. This induces a variance over initializations that becomes stronger as $\alpha_\ell$ is increased. For $\alpha_\ell > 1$, $\mathrm{Var}_{\F} + \mathrm{Var}_{\X\F}$ remains finite even as the hidden layer sizes go to infinity. This residual variance at infinite width can be seen from the approximation $\gamma^{(\ell)}/(1-\gamma^{(\ell)}) \approx \alpha_\ell-1 + 1/(N_\ell/P-1)$ with normalizable spectrum used in \citet{zavatone2023learning} and based on earlier results of \citet{maloney2022solvable}. We illustrate this effect in Figure \eqref{fig:finite_width_scaling}. 

The capacity-limiting effect of structured weights is related to the rotation-invariance of linear random feature models noted in \citet{zavatone2023learning}: even if the task is low-dimensional, meaning that only a low-dimensional signal needs to be propagated through the network, the lack of correlations between the layers means that this signal cannot be preserved through selective routing along large-variance dimensions. As a result, we suggest that the ability to coordinate signal propagation across layers is an important characteristic of feature learning in fully-trained deep networks. It would also be interesting to explore the connections between weight decay exponents and the exponents of finite-$N$ corrections in wide feature-learning networks.

We can also extend our analysis beyond the $\alpha_\ell > 1$ case. As in Section \eqref{sec:non_normalizable}, when $1/2 \alpha_\ell < 1$, we have that $\gamma^{(\ell)}$ scales nontrivially with $N_\ell$ as $(N_\ell/P)^c$, with $c = (1-\alpha_\ell)/\alpha_\ell$. In the language of \citet{bahri2021explaining}, this gives an example of nontrivial variance-limited scaling, that is, there is nontrivial scaling with respect to the bottleneck parameter $N_\ell$.

\subsubsection{Characterization of All Scaling Regimes}\label{sec:lrf_all_scaling}

We now consider the scaling regimes in the case of general $\lambda, \sigma_\epsilon^2$ in the case of a deep structured linear random feature model, as considered in Section \ref{sec:deep_structured_lrf}. We will take the spectrum of $\S$ to be normalizable. At finite ridge we need $\lambda > \min(P, N)^{-\alpha}$ so that $\kappa_2 \sim \lambda$, otherwise $\kappa_2$ will go as $\min(P, N)^{-\alpha}$ and the situation becomes equivalent to the ridgeless setting. If $\lambda$ exceeds this threshold, we have
\begin{equation}
\begin{aligned}
        -\kappa_2^2 \tf_1' &\sim \lambda^{2 \min(r, 1)}, \quad \kappa_2 \tf_1 \sim \lambda^{2 \min(r, 1/2)}\\
    \quad D \df_1 &\sim  D \df_2 \sim \lambda^{-1/\alpha}, \quad \gamma_2 \sim \frac{\lambda^{-1/\alpha}}{P}.
\end{aligned}
\end{equation}
Then for general structured random features from Equation \eqref{eq:dlogs_to_S}
\begin{equation}
    \frac{d \log \kappa_1}{d \log \kappa_2} =  1 + \frac{\df_1-\df_2}{\df_1} \sum_{\ell=1}^L \frac{\df_{\S_\ell}^2(\kappa_\ell)}{\df_{\S_\ell}^1(\kappa_\ell) - \df_{\S_\ell}^2(\kappa_\ell)}.
\end{equation}
We have by definition of $\kappa_\ell$ that $\df_{\S_\ell}^1(\kappa_\ell) = \frac{D}{N_\ell} \df_1 \sim \lambda^{-1/\alpha}/N_\ell$.
Assuming $\S_\ell$ has a power law spectrum with exponent $\alpha_\ell$, let $c_\ell$ be $\min(\max(\frac{1-\alpha_\ell}{\alpha_\ell}, 1), 0)$. Then, taking $N = \min(\{ N_\ell \}_{\ell=1}^L)$ to be the smallest width and $c$ the corresponding $c_\ell$:
\begin{equation}
\begin{aligned}
\frac{d \log S}{d \log \df_1} &\sim \left(\frac{\lambda^{-1/\alpha}}{N}\right)^{c},\\
    \frac{d \log \kappa_2}{d \log \kappa_1} \sim 1, \quad & 1- \frac{d \log \kappa_2}{d \log \kappa_1} \sim \left(\frac{\lambda^{-1/\alpha}}{N}\right)^{c}.
\end{aligned}
\end{equation}
We have used the fact that $\df_1 \sim \df_2$ when $\S$ has normalizable spectrum. Finally from Equation \eqref{eq:gamma2} we have $\gamma_1 \sim \lambda^{-1/\alpha}/P$. Together this gives:
\begin{equation}\label{eq:LRF_lambda_scaling}
    E_g \sim \lambda^{2 \min(r, 1)} + \lambda^{2 \min(r, 1/2)} \left(\frac{\lambda^{-1/\alpha}}{N}\right)^{c} + \sigma_\epsilon^2 \frac{\lambda^{-1/\alpha}}{P}.
\end{equation}

If we take the ridge to scale as $\lambda \sim P^{-l} + N^{-l}$ then in the overparameterized regime this is effectively $P^{-l}$ and in the bottlenecked regime this is effectively $N^{-l}$. As $N \to \infty$ this recovers the ridge scaling considered in Section \ref{sec:kernel_scaling}. If $l < \alpha$ then $\kappa_2 \sim \min(P, N)^{-\ell}$. If $l > \alpha$ then we achieve the ridgeless scaling limit $\kappa_2 \sim \min(P, N)^{-\alpha}$. 

Using Equations \eqref{eq:tf_scalings}, in the bottlenecked regime, $N < P$ we get
\begin{equation}\label{eq:LRF_N_scaling}
    E_g \sim  \frac{ N^{-2  \min(\alpha, l)  \min(r, 1/2)}}{1- N/P}  + \sigma_\epsilon^2 \frac{N^{\min(1, l/\alpha)}}{P}.
\end{equation}
This gives the following scaling regimes in $N$ (resolution limited) and $P$ (variance limited):
\begin{equation}\label{eq:LRF_all_scalings1}
    E_g \sim \begin{cases}
        \displaystyle \frac{ N^{-2  \alpha  \min(r, 1/2)}}{1- N/P}, \quad \alpha < l; \, N^{-2 \alpha \min(r, 1/2)} \gg \sigma_\epsilon^2 N/P & \text{Signal dominated}\\
        \displaystyle \frac{ N^{-2  l  \min(r, 1/2)}}{1- N/P}, \quad l < \alpha; \, N^{-2 l \min(r, 1/2)} \gg \sigma_\epsilon^2 N^{l/\alpha}/P  & \text{Ridge dominated}\\
        \displaystyle \sigma_\epsilon^2 \frac{N}{P} \hspace{0.75in} \alpha < l;\, N^{-2 \min(\alpha, l) \min(r, 1/2)} \ll \sigma_\epsilon^2 N/P  &  \text{Noise dominated}\\
        \displaystyle \sigma_\epsilon^2 N^{l/\alpha}/P \hspace{0.45in} l < \alpha;\, N^{-2 \min(\alpha, l) \min(r, 1/2)} \ll \sigma_\epsilon^2 N^{l/\alpha}/P &  \text{Noise mitigated}
    \end{cases}
\end{equation}
The resolution-limited exponents are similar but not identical to those in the linear regression  setting \eqref{eq:LR_Eg_scaling_full}. The variance-limited exponents in $P$ are always trivial.

\begin{figure}[t]
    \centering
    \includegraphics[width=4in]{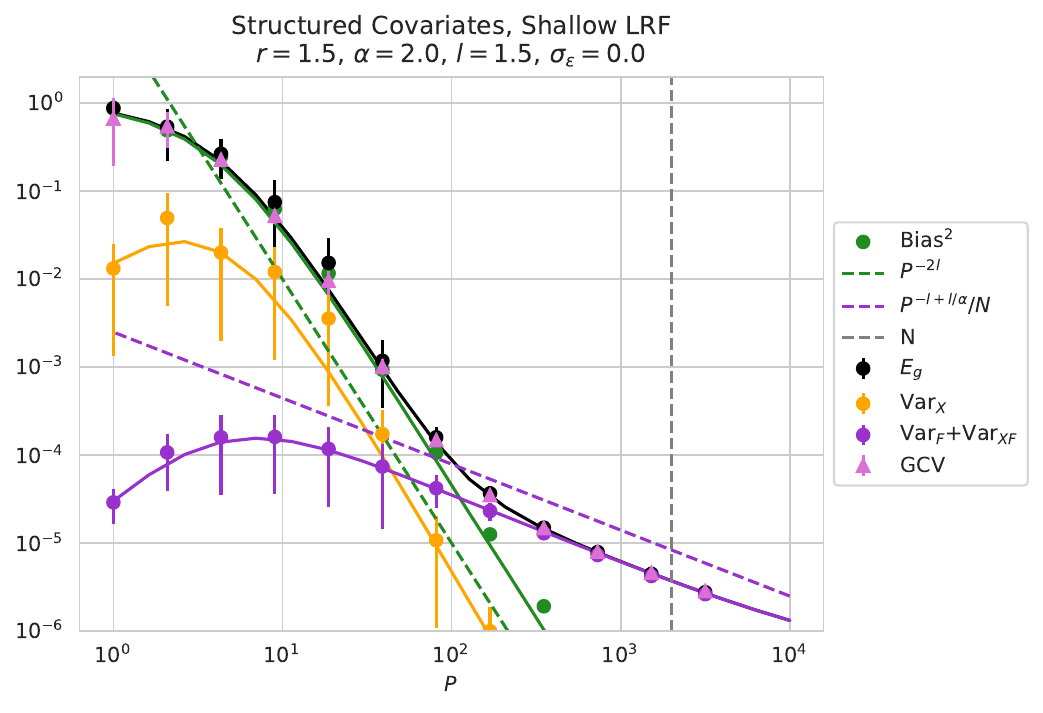}
    
    \includegraphics[width=4in]{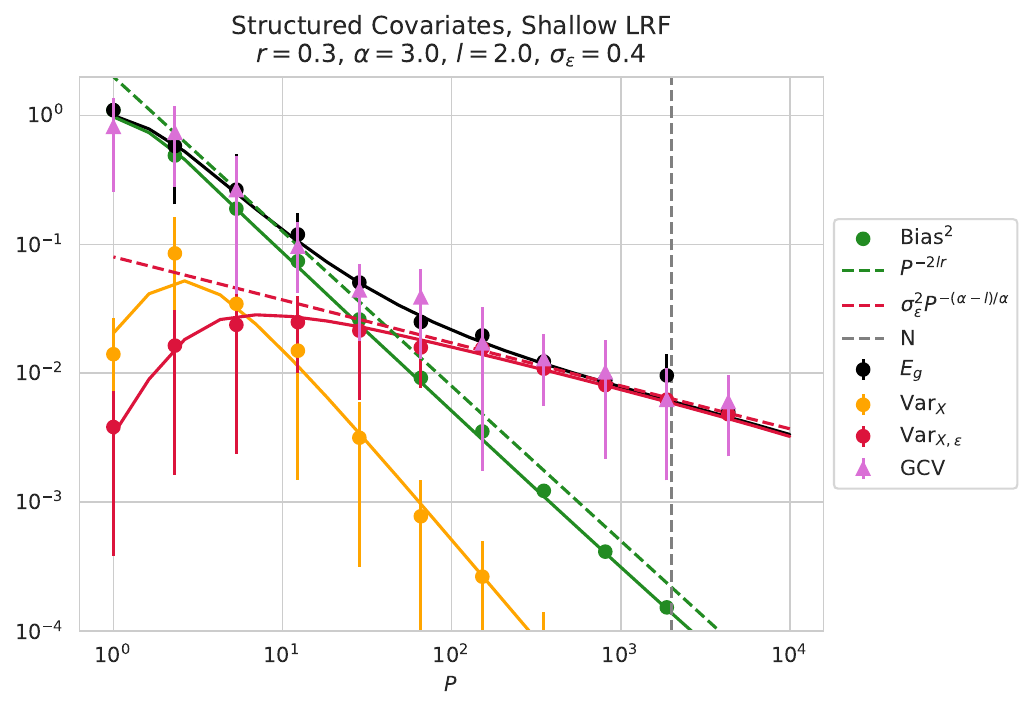}
    \caption{Shallow linear random feature model with $D=10000$, $N=2000$ and isotropic weights exhibiting multiple scaling regimes. Dashed lines are exact power laws for reference. Left:  exhibiting the transition from ridge dominated to joint variance and ridge dominated scaling. Solid curves are theory and dots are empirical results. Right: Shallow linear random feature model exhibiting the transition from ridge dominated to noise mitigating behavior. Relevant variances are plotted. In both cases, the double descent peak at $P=N$ is eliminated.}
    \label{fig:deep_LRF_scaling2}
\end{figure}

In the overparameterized regime, $P < N$ we have
\begin{equation}\label{eq:LRF_P_scaling}
\begin{aligned}
    E_g &\sim 
        \displaystyle P^{-2   \min(\alpha, l)   \min(r, 1)}  + P^{- 2  \min(\alpha, l)  \min(r, 1/2)} \left(\frac{P^{\min(1, l/\alpha)}}{N} \right)^c  + \sigma_\epsilon^2 P^{-1 + \min(1, l/\alpha)}. 
    \end{aligned}
\end{equation}
This gives the following scaling regimes:
\begin{equation}\label{eq:LRF_all_scalings2}
     E_g \sim \begin{cases}
        \displaystyle P^{-2  \alpha  \min(r, 1)}, \hspace{0.3in} \alpha < l;\,  P \ll P_{\e};\, r \leq 1/2 \text{ or } P \ll P_{\F}  & \text{Signal dominated}\\
        \displaystyle  P^{- \alpha} \left(\frac{P}{N}\right)^c,   \hspace{0.2in} \alpha < l;\,  P \ll P_{\e}; r > 1/2; P \gg P_{\F}  & \text{Var$_{\F}$ dominated} \\
        \displaystyle P^{-2  l  \min(r, 1)}, \hspace{0.35in} l < \alpha;\,  P \ll P_{\e};\, r \leq 1/2 \text{ or } P \ll P_{\F} & \text{Ridge dominated}\\
        \displaystyle  P^{- l } \left(\frac{P^{l/\alpha}}{N}\right)^c, \hspace{0.2in} l < \alpha;\,  P \ll P_{\e}; r > 1/2; P \gg P_{\F} & \text{Ridge \& Var$_{\F}$ dominated} \\
        \displaystyle \sigma_\epsilon^2 P^{0}, \hspace{0.8in} \alpha < l; P \gg P_{\e}, & \text{Noise dominated}\\
        \displaystyle \sigma_\epsilon^2 P^{-\frac{\alpha-l}{\alpha}},\hspace{0.6in} l < \alpha; P \gg P_{\e} & \text{Noise mitigated}\\
    \end{cases}
\end{equation}
Here,
\begin{equation}
    P_{\F} \sim N^{\frac{c}{c + 2 \min(\alpha, l) \min(r-1/2, 1/2)}}
\end{equation}
and $P_{\e}$ is defined to be the value of $P$ where either of the last two scalings become comparable in size to the first four: 
\begin{equation}
    \min(P_{\e}^{-2 \min(\alpha, l) \min(r, 1)}, P_{\e}^{-2 \min(\alpha, l) \min(r, 1/2)} P_{\e}^{c\,  \max(1, l/\alpha) }/N^c) = \sigma_\epsilon^2 P_{\e}^{-\min(0, \frac{\alpha-l}{\alpha})}.
\end{equation}

\subsubsection{Comparison with Defillippis, Loureiro, and Misiakiewicz}

\begin{figure}[t]
    \centering
    \includegraphics[scale=0.6]{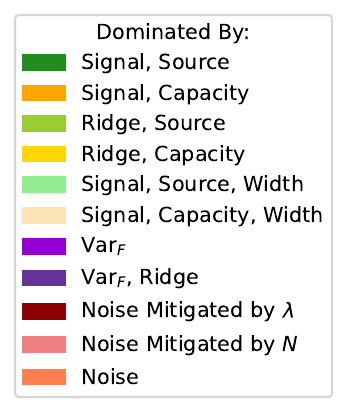} \includegraphics[scale=0.6]{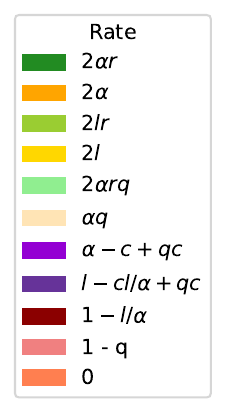}\\
    \includegraphics[width=0.3\linewidth]{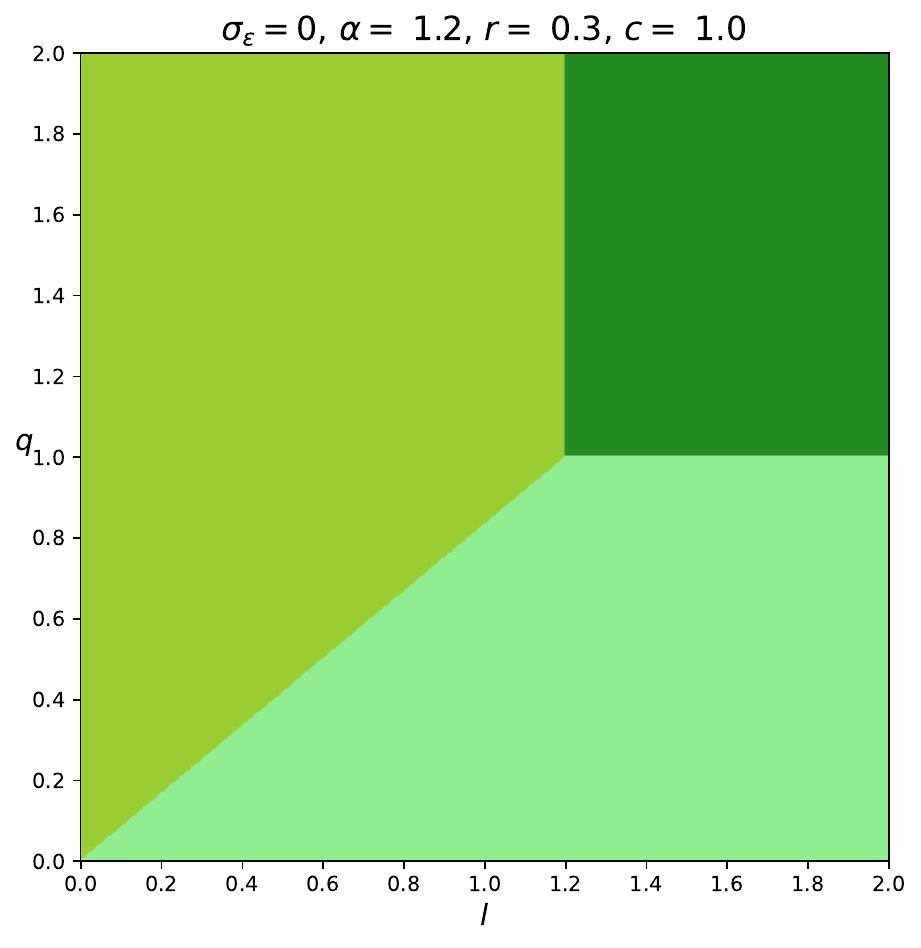}
    \includegraphics[width=0.3\linewidth]{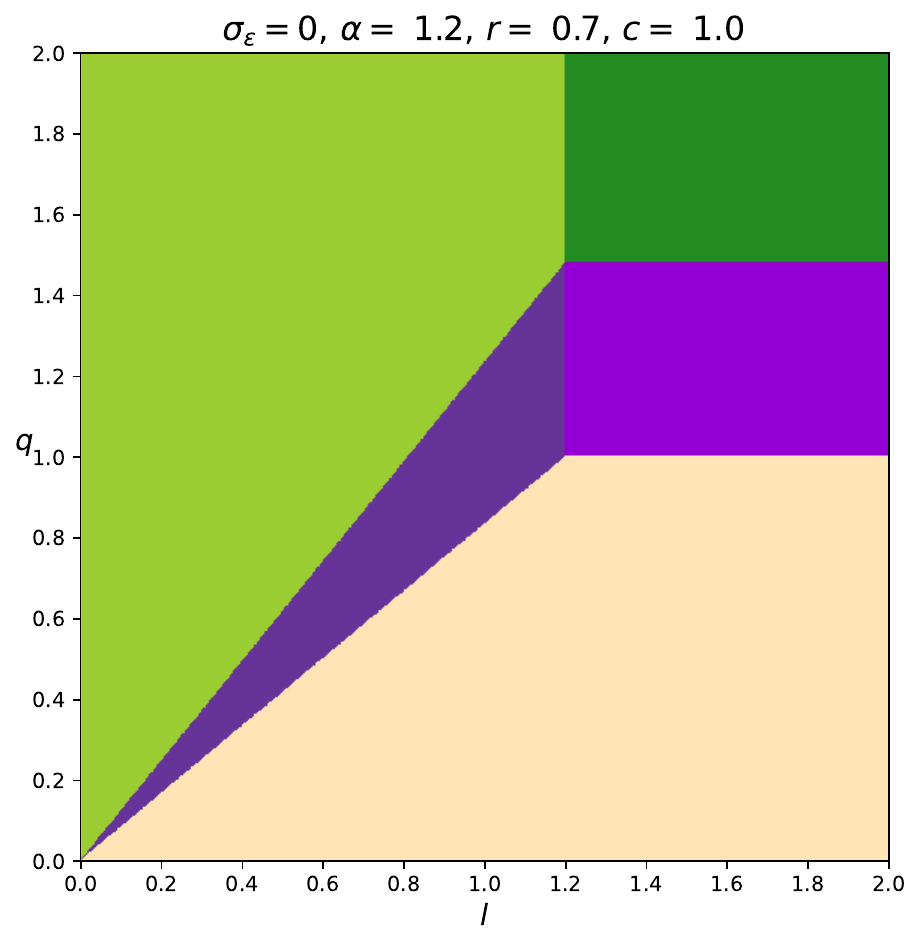}
    \includegraphics[width=0.3\linewidth]{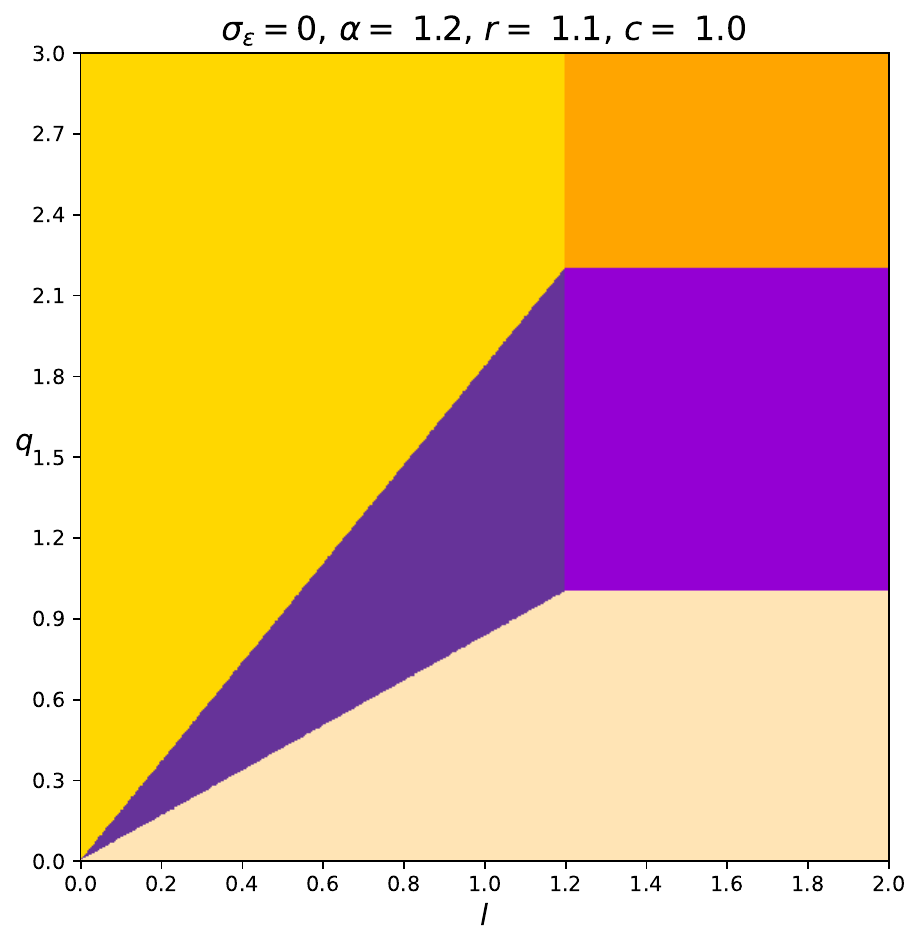}
    \includegraphics[width=0.3\linewidth]{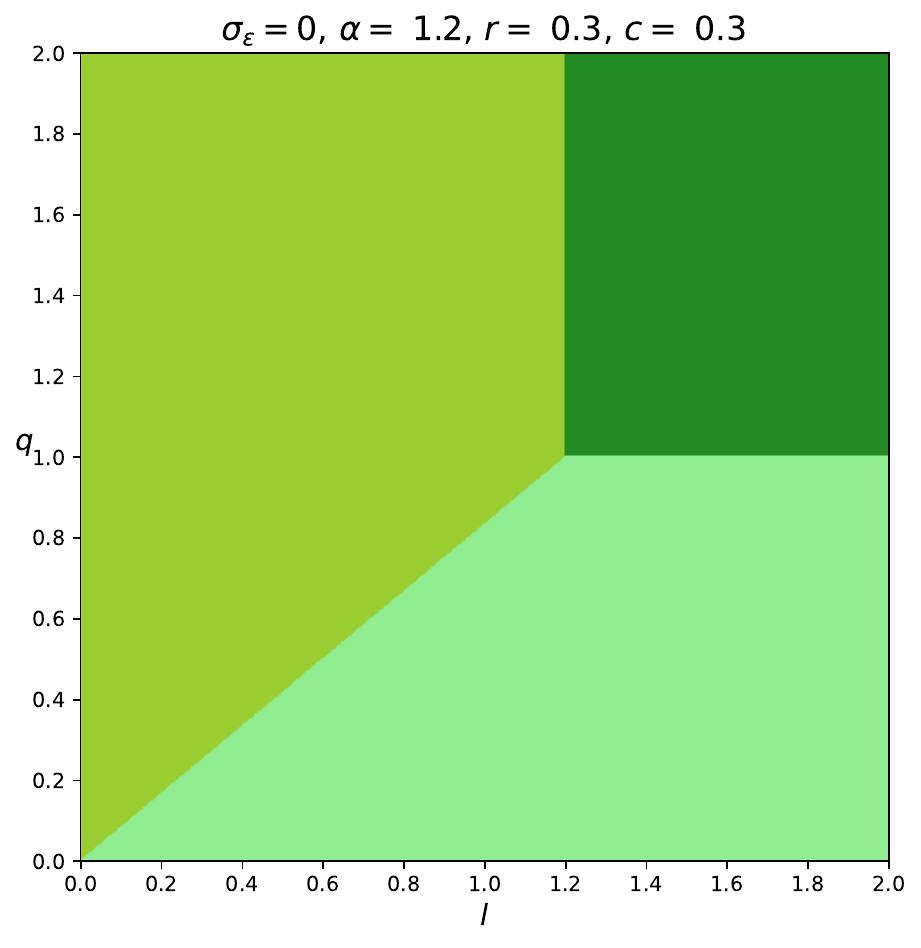}
    \includegraphics[width=0.3\linewidth]{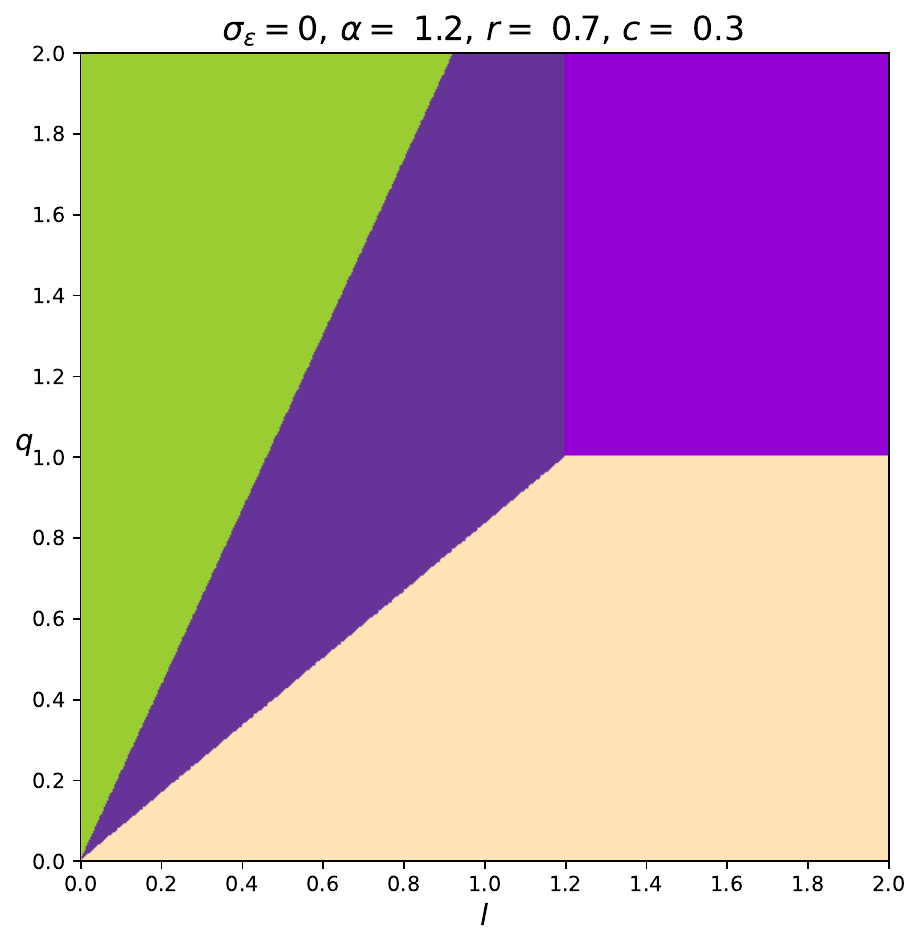}
    \includegraphics[width=0.3\linewidth]{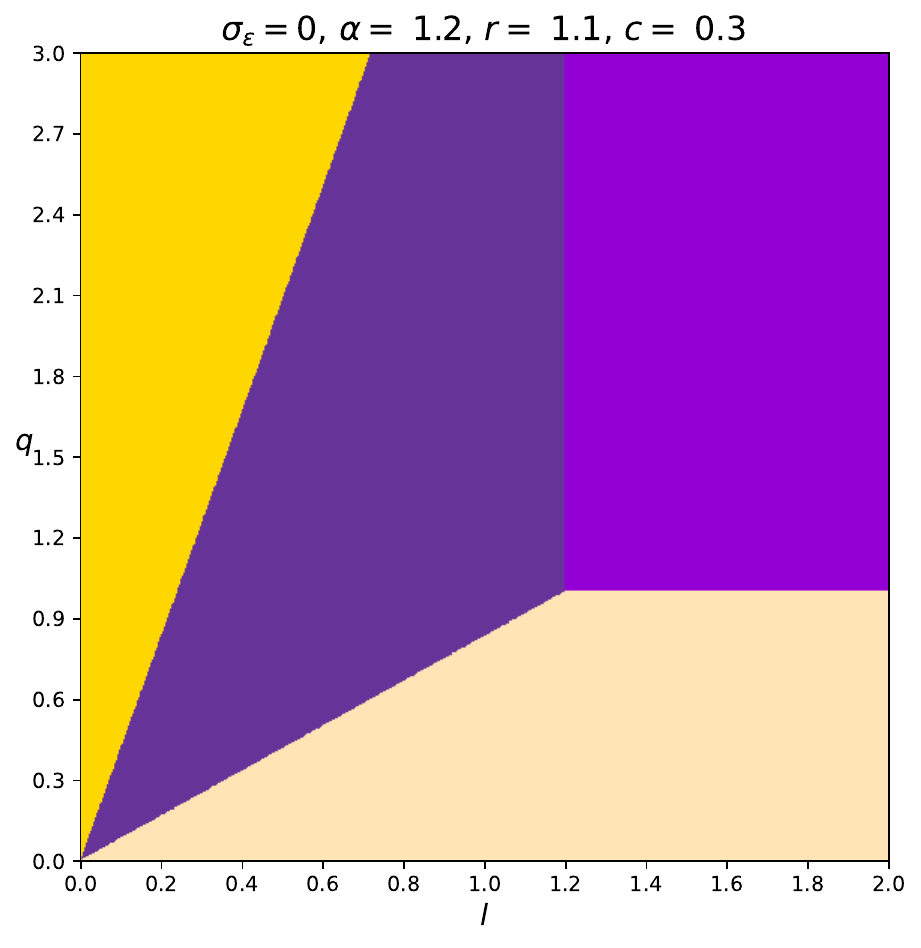}
    \includegraphics[width=0.3\linewidth]{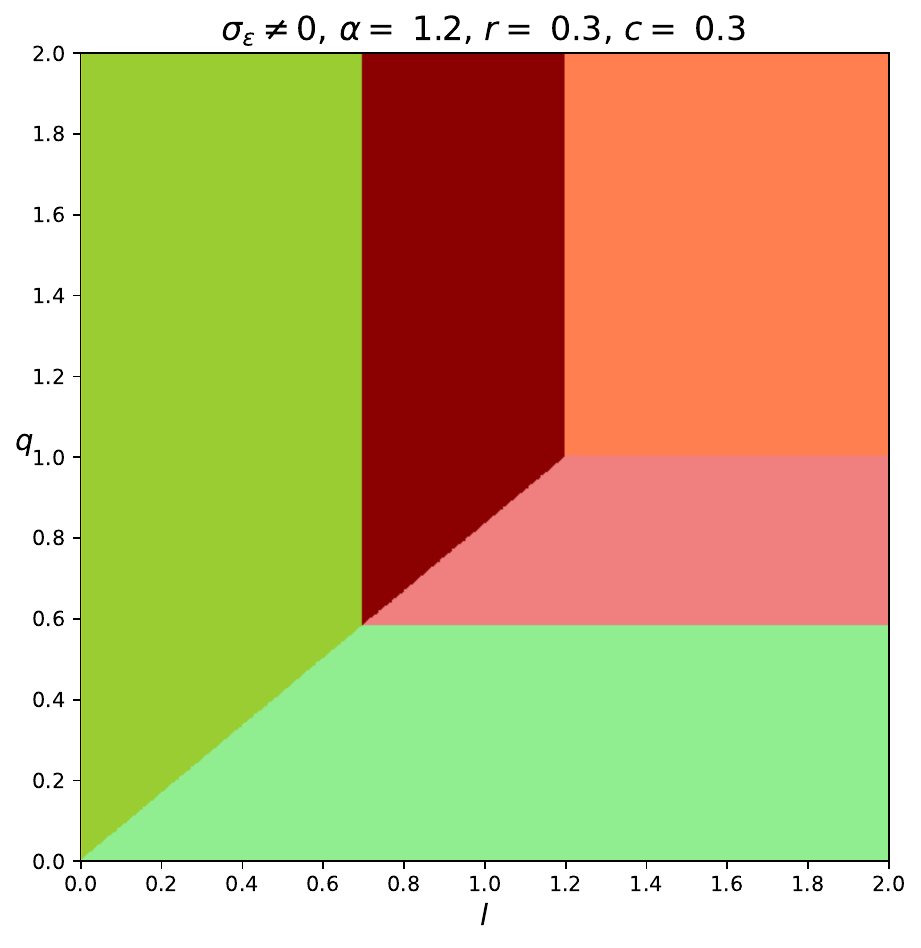}
    \includegraphics[width=0.3\linewidth]{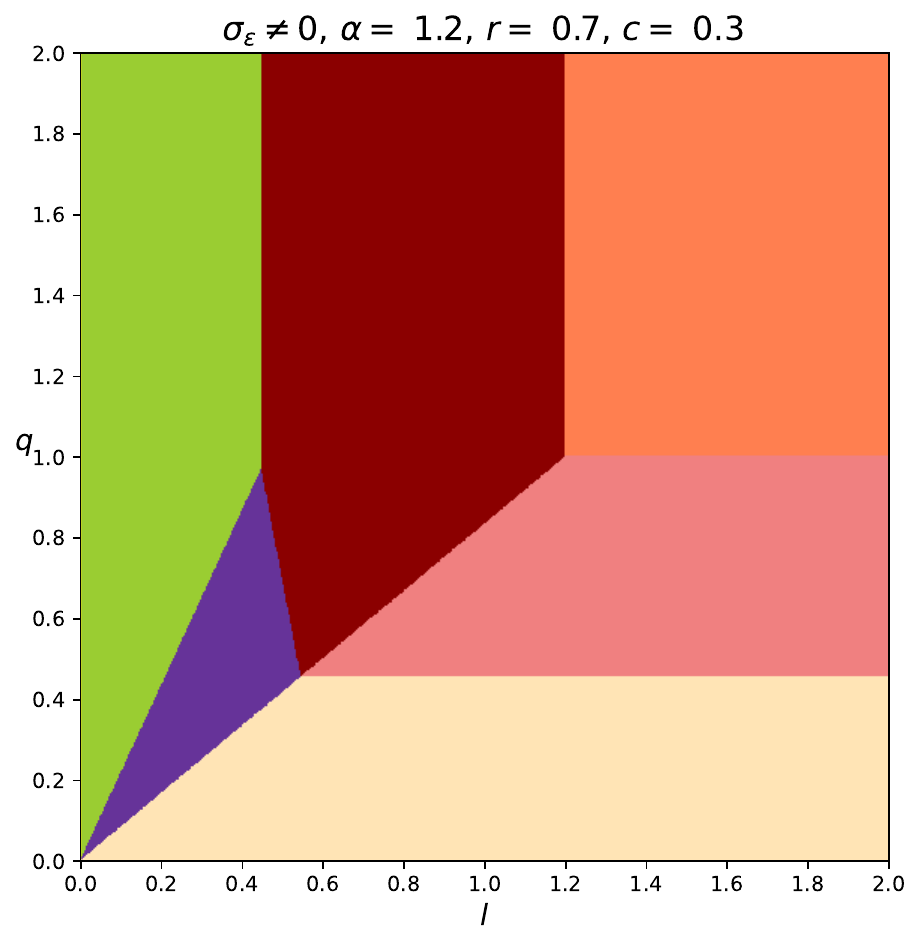}
    \includegraphics[width=0.3\linewidth]{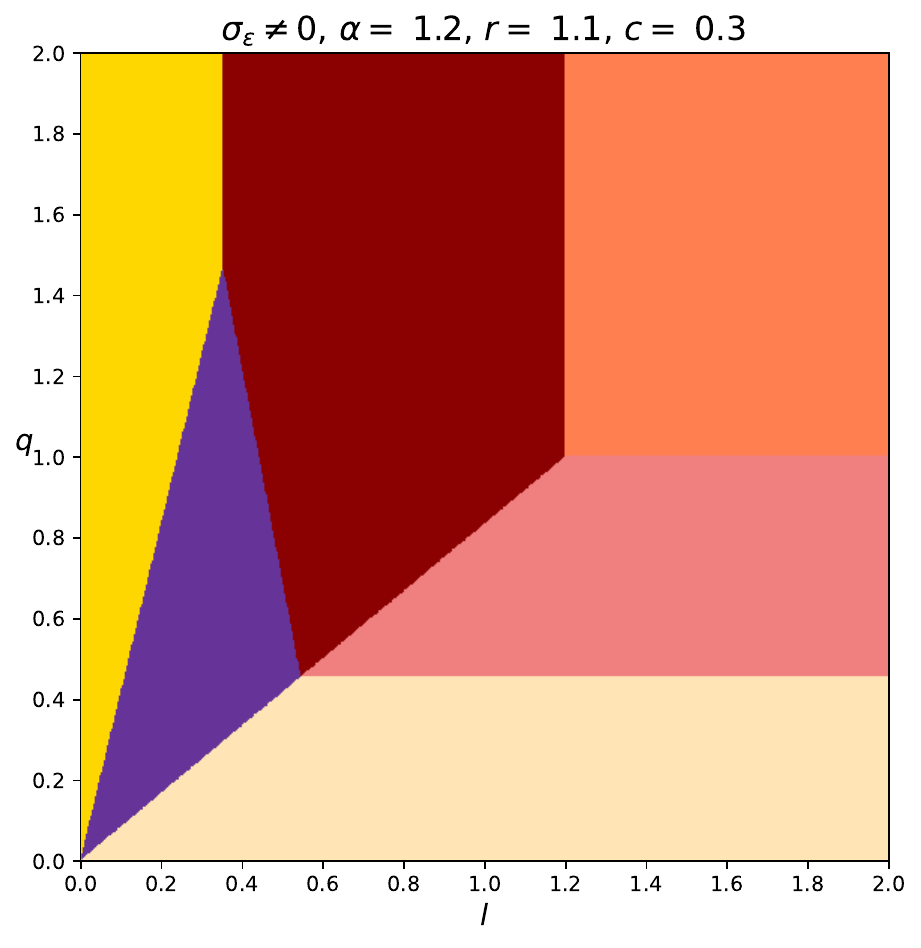}
    
    \caption{Phase plots for the asymptotic rate in the $l, q$ plane at different values of $\alpha, r, c$, inspired by \citet{defilippis2024dimension}. The colors are chosen to match the palette of the other plots in this section, and specifically the fine-grained bias-variance decomposition in Figure \ref{fig:LRF_BV}. The $\mathrm{Var}_{\F}$-dominated region does not appear when $r < 1/2$ and these plots are insensitive to the value of $c$. When $r > 1/2$, smaller values of $c$ expand the $\mathrm{Var}_{\F}$-dominated regime.}
    
    \label{fig:rate_plots}
\end{figure}

Shortly after the initial release of this work on arXiv, \citet{defilippis2024dimension} posted a very nice paper in which they examined a one-layer random feature model. In our notation, they considered the scaling $N \sim P^{q}$ and $\lambda \sim P^{-l}$. Our results and theirs are compatible. We consider Equations \eqref{eq:LRF_lambda_scaling}, \eqref{eq:LRF_N_scaling}, and \eqref{eq:LRF_P_scaling} under the replacement $N = P^{q}$. Further, we exclude the previously considered case of $\lambda \sim N^{-l}$ as this is accounted for by taking $\lambda \sim P^{-l}$ given that $N$ scales with $P$. One then obtains the following conditional expression for the asymptotic decay rate as $P \to \infty$  when $\sigma_\epsilon=0$:
\begin{equation}\label{eq:all_cases_noiseless}
    - \frac{\log E_g}{\log P} \sim \min\left[\underbrace{2 \alpha   \min(r, 1) \min \left(1, \frac l \alpha \right)}_{\mathrm{Bias}^2 + \mathrm{Var}_{\X}}, \; \underbrace{ 2 \alpha q \min\left(r, \frac12 \right)}_{\mathrm{Bias}^2 + \mathrm{Var}_{\F}}, \; \underbrace{\left(\alpha - c \right)\min\left(1,\frac l\alpha\right)   + q c}_{\mathrm{Var}_{\F} + \mathrm{Var}_{\X, \F}}\right].
\end{equation}
Here we have under-braced the cases to highlight which sources of variance lead to the scaling behavior observed. If $\sigma_\epsilon \neq 0$, one obtains an additional case:
\begin{equation}\label{eq:all_cases_noisy}
    - \frac{\log E_g}{\log P} \sim \min\left[\underbrace{2 \alpha   \min(r, 1) \min \left(1, \frac l \alpha \right)}_{\mathrm{Bias}^2 + \mathrm{Var}_{\X}}, \; \underbrace{ 2 \alpha q \min\left(r, \frac12 \right)}_{\mathrm{Bias}^2 + \mathrm{Var}_{\F}}, \; \underbrace{\left(\alpha - c \right)\min\left(1,\frac l\alpha\right)   + q c}_{\mathrm{Var}_{\F} + \mathrm{Var}_{\X, \F}}, \; \underbrace{1 - \min\left(1, \frac l\alpha, q\right)}_{\mathrm{Var}_{\X, \e} + \mathrm{Var}_{\X, \F, \e}} \right].
\end{equation}
In the case of $c = 1$, namely when the feature weights have power law structure decaying slower than $k^{-1/2}$, this recovers the rates obtained by \citet{defilippis2024dimension}. Increasing the power law decay of the random features amounts to decreasing $c$, which expands the region over which $\mathrm{Var}_{\F}$-related scaling dominates. We highlight several phase plots of these asymptotic rates in Figure \ref{fig:rate_plots}. 

We stress that although these expressions capture the final rate achieved when $P \to \infty$ with $N = P^q$, there can be many different scaling regimes that the loss curves can pass through before they reach the asymptotic rate. 

\clearpage

\section{Models with Additive Feature Noise}
\label{sec:NLRF}

\subsection{Setup and Motivation}

In this section, we turn our attention to a model in which the true latent features are not only randomly projected, but also corrupted by additive noise. Concretely, we consider a model where the targets are generated as
\begin{align}
    y_{\mu} = \bar{\w} \cdot \x_{\mu} + \epsilon_{\mu},
\end{align}
while the student has access only to features that are both projected by a matrix $\bm{F} \in \mathbb{R}^{D \times N}$ and corrupted by additive noise $\bm{\xi} \in \mathbb{R}^{N}$ that is independent and identically distributed for each sample. As before, the entries in $\F$ have variance $1/D$ while the entries in $\bm \xi$ are order $1$. Using the same setup as in Equation \eqref{eq:RF_dfn}, we have
\begin{align}\label{eq:learned_NLRF}
    f(\x) =  (\x^\top \F  + \bm{\xi}^\top ) \v.
\end{align}
Here $\v$ are the trainable weights. We take the latent features and additive noise to be jointly Gaussian and independent:
\begin{align}
    \begin{pmatrix}
        \x \\
        \bm \xi
    \end{pmatrix} \sim \mathcal{N}\left( \bm 0_{D+N},  \begin{pmatrix} \Sig & \bm{0} \\ \bm{0} & \Sig_{\xi} \end{pmatrix} \right).
\end{align}
This model has been prominently studied in several prior works. It was first explicitly solved by \citet{mei2022generalization}. There, the authors considered a random feature model $f(\x) = \sigma(\x^\top \F) \v$ where $\sigma$ is a nonlinearity applied element-wise and $\v$ is trainable. $\F$ has random entries with mean zero and variance $1/D$. Mei and Montanari highlighted that for a random feature model where features $\F$ are mapped through a nonlinearity $\sigma$ with
\begin{equation}
    \mu_0 = \mathbb E_{x \sim \mathcal N(0, 1)} [\sigma(x)], \quad \mu_1 =  \mathbb E_{x \sim \mathcal N(0, 1)} [x \sigma(x)], \quad \mu_{\star} =  \mathbb E_{x \sim \mathcal N(0, 1)} [\sigma(x)^2] - \mu_0^2 - \mu_1^2
\end{equation}
the asymptotic generalization error is equal to that for a Gaussian equivalent model. The Gaussian equivalent makes the replacement
\begin{equation}
    \sigma(\x^\top \F) \simeq \mu_0 \bm{1} + \mu_1 \x^\top \F + \bm \xi, \quad \bm \xi \sim \mathcal N(0, \mu_{\star} \mathbf I).
\end{equation}
Here $\mathbf{1}$ is the vector of ones. Taking $\mu_0 = 0, \mu_1 = 1$ we recover Equation \eqref{eq:learned_NLRF} in the case where the elements of $\bm \xi$ are independent and normally distributed for each sample. Equivalences of random features passed through nonlinearities in the proportional limit have also been studied in \citet{pennington2017nonlinear, dhifallah2020precise, hu2022universality}. Scalings beyond the linear regime have been studied in \citet{lu2022equivalence,hu2024asymptotics}.

An alternative reason to study random features corrupted by additive noise is an extension of the perspective taken in \citet{maloney2022solvable} for linear random features. There, one takes $D \gg N, P$. The $D$-dimensional space can be viewed as an analogue of the infinite-width NTK features, while $N$ is viewed as the number of parameters. A linear random feature model is thus similar to doing regression with a random feature approximation to the NTK. This is similar to the finite-width NTK (also known as the \textbf{empirical neural tangent kernel} or eNTK). However, it is known that the entries of the finite-width eNTK also have initialization-dependent variance going as $1/n$, where $n$ is the width of the network \cite{dyer2019asymptotics}.\footnote{This is distinct from $N$ in the random feature model, which we have also been calling width.} This enters at a different scale than the number of parameters $N$. The authors in \citet{atanasov2022onset} use this additive noise to model eNTK fluctuations. This leads to a performance decrease, driven primarily by initialization variance at relatively small values of $P$.

\subsection{Averaging Over Data}

Let $\X \in \mathbb R^{P \times D}$ be the design matrix on the train set, with $\X_{\mu i} = [\x_\mu]_i$. Let $\bm \Xi \in \mathbb R^{P \times N}$ be the feature noise matrix on the train set, with $\bm \Xi_{\mu i} = [\bm \xi_\mu]_i$.
Define the matrices $\overline \X$ and $\overline \F$ to be
\begin{align}
    \overline \X \equiv \begin{pmatrix} \Z_1 & \bm \Z_2 \end{pmatrix} \in \mathbb R^{P \times (D+N)}, \quad
    \overline \F \equiv \begin{pmatrix} \S^{1/2} \F \\ \S^{1/2}_\xi \end{pmatrix} \in \mathbb R^{(D+N)\times N}.
\end{align}
Here $\Z_1, \Z_2$ are both unstructured Gaussian matrices. All structure is added by the features $\overline \F$. Then $\overline \X, \overline \F$ are free of one another and we can apply deterministic equivalence. Moreover, $f(\X) = \overline \X\, \overline \F \v$ corresponds to a linear random feature model, as studied in the previous section. We also define the extended teacher vector:
\begin{align}
    \bar \w_{D+N} \equiv \begin{pmatrix} \S^{1/2} \bar{\w} \\ \bm{0}_N \end{pmatrix} \in \mathbb R^{D + N}.
\end{align}
This accounts for the fact that the target labels do not depend on the noise $\bm{\xi}$.

We can now directly apply the formulas for $E_g$ in the linear random feature case from the prior section. 
\begin{align}
    E_g &= - \frac{\kappa_1^2}{1-\gamma_1} \partial_{\kappa_1} \bar \w^\top_{D+N}  (\overline \F \overline \F^\top + \kappa_1 \mathbf I)^{-1}  \bar \w_{D+N} + \sigma_\epsilon^2 \frac{\gamma_1}{1-\gamma_1},
\end{align}
\begin{equation}
    \kappa_1 = \frac{\lambda}{1- \frac{N}{P} \df^1_{\overline \F^\top \overline \F}(\kappa_1)}, \quad
    \gamma_1 = \frac{N}{P} \df^2_{\overline \F^\top \overline \F} (\kappa_1),
\end{equation}
\begin{align}
    \overline \F^\top \overline \F &=  \F^\top \S \F + \S_{\xi}\\
    \overline \F  \overline \F^\top &= 
    \begin{pmatrix}
        \S^{1/2} \bm{F}\bm{F}^{\top} \S^{1/2} & \S^{1/2} \bm{F} \S_{\xi}^{1/2} \\ 
        \S_{\xi}^{1/2} \bm{F}^{\top} \S^{1/2} & \S_{\xi} 
    \end{pmatrix}.
\end{align}
Because of the structure of $\overline \w_{D+N}$, we care only about the top left block in:
\begin{align}
    (\overline \F \overline \F^\top + \kappa_1 \mathbf I )^{-1} &= \begin{pmatrix}
        \kappa_1 \mathbf I + \S^{1/2} \bm{F}\bm{F}^{\top} \S^{1/2} & \S^{1/2} \bm{F} \S_{\xi}^{1/2} \\ 
        \S_{\xi}^{1/2} \bm{F}^{\top} \S^{1/2} & \kappa_1 \mathbf I +  \S_{\xi} 
    \end{pmatrix}^{-1}
    \equiv
    \begin{pmatrix}
        \bm{M}_{11} & \bm{M}_{12} \\ \bm{M}_{12}^{\top} & \bm{M}_{22}
    \end{pmatrix}.
\end{align}
By applying the Schur complement formula \cite{horn2012matrix}, this can be written compactly as:
\begin{align}
    \bm{M}_{11}
    &= \left[\kappa_1 \mathbf I_D + \kappa_1 \tfrac1D \S^{1/2} \bm{F} (\S_{\xi} + \kappa_1 \mathbf I_N)^{-1} \bm{F}^{\top} \S^{1/2}  \right]^{-1}\\
    &=  \frac{1}{\kappa_{1}} \left[\mathbf I_D - \S^{1/2} \bm{F} ( \kappa_{1} \mathbf I_N + \bm{\Sigma}_{\xi} + \bm{F}^{\top} \S \bm{F} )^{-1} \bm{F}^{\top} \S^{1/2}  \right].
\end{align}
In the last line we have used the Woodbury matrix inversion identity \cite{horn2012matrix}. Upon taking the appropriate $\kappa_1$ derivatives, the signal term reproduces the formula for the model studied in \citet{atanasov2022onset}. This is also equivalent to the very general Gaussian model studied in \citet{loureiro2021learning}.

At this point, we will specialize to the case of isotropic noise $\S_\xi = \sigma^2_\xi \mathbf I$. 
This further simplifies the signal term to:
\begin{equation}
    -\frac{\kappa_1^2}{1-\gamma_1} \partial_{\kappa_1} \left[\frac{\kappa_1+\sigma^2_\xi}{\kappa_1} \bar \w^\top \S^{1/2} (\S^{1/2} \F \F^\top \S^{1/2} +  (\kappa_1 + \sigma_\xi^2) \mathbf I)^{-1} \S^{1/2} \bar \w \right].
\end{equation}

\subsection{Averaging Over Isotropic Features}

Under the assumption that $\F^\top \F$ is distributed as a white Wishart matrix, we get:
\begin{equation}
    \kappa_2 = \frac{\kappa_1 + \sigma^2_\xi}{\tfrac{N}{D} - \df^1_{\S}} \Rightarrow \kappa_2 \left(\frac{N}{D} - \df^1_{\S} - \frac{\sigma^2_\xi}{\kappa_2} \right) = \kappa_1.
\end{equation}
Because of the additive shift, $\df_{\overline \F^T \overline \F}^1 = \df_{\F^\top \S \F + \sigma_\xi^2}^1$ is related to $\df_{\S}^1$ as follows:
\begin{equation}
\begin{aligned}
    \df^1_{\F^\top \S \F + \sigma^2_\xi}(\kappa_1) &= \df^1_{\F^\top \S \F}(\kappa_1+\sigma^2_\xi) + \sigma^2_\xi \frac{1 - \df^1_{\S_{\F^\top \S \F}}(\kappa_1+\sigma^2_\xi)}{\kappa_1 + \sigma^2_\xi}\\
    &= \frac{D}{N} \df^1_{\S}(\kappa_2) +  \sigma^2_\xi \frac{1 - \tfrac{D}{N} \df^1_{\S}(\kappa_2)}{\kappa_1 + \sigma^2_\xi}
    \\
    &= \frac{D}{N} \underbrace{\left[\df^1_{\S}(\kappa_2) + \frac{\sigma^2_\xi}{\kappa_2}\right]}_{\overline{\df}_1}.
\end{aligned}
\end{equation}
Here we have defined $\overline \df_1$. The final expressions simplify dramatically in terms of this quantity.  Then:
\begin{equation}
    \overline{\df}_2 = \partial_{\kappa_2}[\kappa_2 \overline{\df}_1] =  \df_2
\end{equation}
\begin{equation}
    \gamma_1 =  \frac{N}{P} \frac{d}{d\kappa_1} [\kappa_1 \df_{\overline \F^\top \overline \F}^1(\kappa_1)] = \frac{N}{P} \overline{\df}_1 \left[1 - \frac{d \log \kappa_2}{d \log \kappa_1} \frac{ \overline{\df}_1 - \df_2}{ \overline{\df}_1}  \right],
\end{equation}
\begin{equation}
    \frac{d \log \kappa_1}{d \log \kappa_2}  = 1 + \frac{1}{N/D - \overline{\df}_1} (\overline{\df}_1 - \df_2).
\end{equation}
Writing $\tf_1 = \tf^1_{\S}$, the generalization error then takes an identical form to the linear random feature case, with the only difference being the replacement $\df_1 \to \overline{\df}_1$ in the self-consistency equation for $\kappa_2$:
\begin{equation}
     \boxed{E_g = -\frac{\kappa_2^2 \tf_1'}{1-\gamma_1}  \frac{d \log \kappa_2}{d \log \kappa_1} + \frac{\kappa_2 \tf_1}{1-\gamma_1}  \left[ 1 -  \frac{d \log \kappa_2}{d \log \kappa_1}\right] + \frac{\gamma_1}{1- \gamma_1} \sigma_\epsilon^2.}
\end{equation}
In the ridgeless limit, we have two behaviors depending on whether $\kappa_1 = 0$ or $\kappa_1 \neq 0$. Note that $\kappa_2$ always stays nonzero in this setting. This highlights that the input dimension $D$ drops out from determining whether the model is overparameterized or underparameterized. The relevant quantities to compare are $N$ and $P$. We have:
\begin{itemize}
    \item $N < P$, underparameterized:

    Then $\kappa_1 = 0$ and $\overline{\df}_1 \to 1$, giving $\gamma_1 = N/P$. Our final formula simplifies to
    \begin{equation}
        E_g = \frac{\kappa_2 \tf_1}{1-N/P} + \sigma_\epsilon^2 \frac{N/P}{1-N/P}.
    \end{equation}
    Here, $\kappa_2$ satisfies the equation
    \begin{equation}
        \frac{N}{D} = \overline{\df}_1 = \df^1_{\S}(\kappa_2) + \frac{\sigma_\xi^2}{\kappa_2}.
    \end{equation}

    \item $P < N$, overparameterized:

    Then $\overline \df_1 \to P/D$ and we get:
    \begin{equation}
        E_g = -\frac{\kappa_2^2 \tf_1'(\kappa_2)}{1 - \frac DP \df_2}  + \frac{\kappa_2 \tf_1 P/N}{1-P/N} + \sigma_\epsilon^2 \left[\frac{\frac{D}{P} \df_2}{1 - \frac{D}{P} \df_2} + \frac{P/N}{1 - P/N} \right].
    \end{equation}
    Here, $\kappa_2$ satisfies the equation
    \begin{equation}
        \frac{P}{D} = \overline{\df}_1 = \df^1_{\S}(\kappa_2) + \frac{\sigma_\xi^2}{\kappa_2}.
    \end{equation}
\end{itemize}
In both cases, these appear identical to the forms of the linear random feature model. Moreover, these expressions recover the results of \citet{mel2021anisotropic, atanasov2022onset}. We leave the extensions of this analysis to deep nonlinear random features with structured weights to future work. 

\subsection{An Interesting Equivalence}

We have seen that we can safely replace $N \df_{\F^\top \S \F + \sigma^2_\xi}(\kappa_1)$ with $D \overline{\df}_1(\kappa_2)$. Moreover, similar to Equation \eqref{eq:multiple_descent_kappa}, we can interpret $D \overline{\df}_1 = \Tr[\tilde \S (\tilde \S + \kappa_2)]$. Here $\tilde \S$ is a covariance matrix having the same spectrum as $\S$ with an additional $\tilde N$ eigenvalues with value $\tilde \sigma_{\xi}^2 \equiv \sigma_\xi^2/\tilde N$ and $\tilde N \to \infty$. Then, $\tilde \sigma_\xi^2 (\tilde \sigma_\xi^2 + \kappa_2)^{-1} \to \tilde \sigma_{\xi}^2/\kappa_2$. Since there are $\tilde N$ of them, the total contribution will yield $\sigma_\xi^2 / \kappa_2$.  These eigenvalues will always remain below the level of resolution given by $\kappa_2$ and thus be un-learnable. Thus, when they are passed through the linear random feature matrix, they act as additive feature noise.  This is analogous to how the higher-order unlearned modes in Section \ref{sec:multiple_descent} act as effective noise.

\subsection{Example: Nonlinear Random Features with Isotropic Covariates}

\begin{figure}[t]
    \centering
    \includegraphics[width=4in]{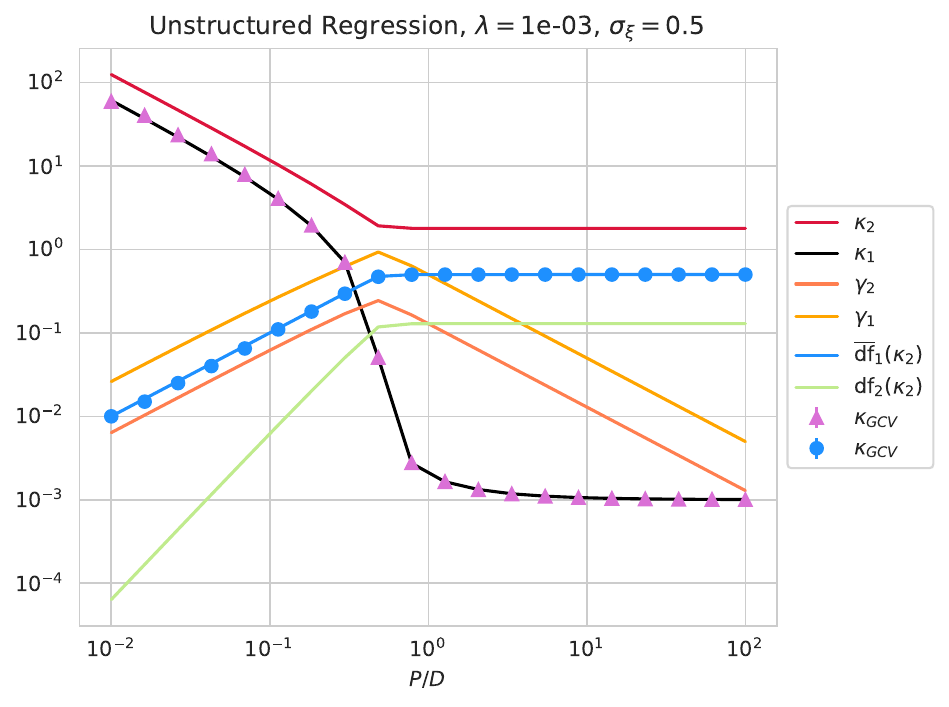}
    \includegraphics[width=4in]{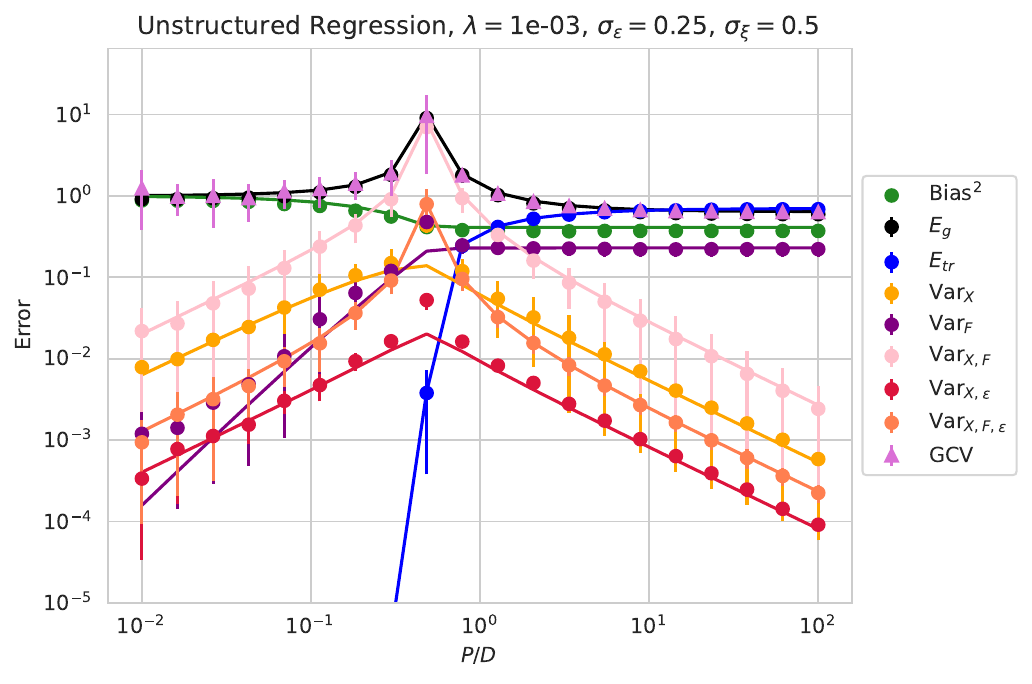}
    \caption{1-layer nonlinear linear random features with unstructured covariates, \textit{i.e.} $\S = \mathbf I$. Left: We plot theory (solid lines) for the various quantities of interest: $\kappa_1, \kappa_2, \gamma_1, \gamma_2$ as well as $\overline{\df}_1(\kappa_2), \df_2(\kappa_2)$. We also plot the estimate of $\kappa_1$ using the training set and find excellent agreement. Right: We plot the training and generalization (blue, black respectively) as well as the bias (green) and variances (orange, purple, pink, red, coral) due to all relevant quantities in the regression. Dots and error bars indicate empirical simulations over 40 seeds over training set and 40 seeds over random feature initializations. Solid curves show theory. We see strong agreement for all relevant quantities. The GCV estimator is plotted as orchid triangles and again we find excellent agreement with the generalization error.   }
    \label{fig:unstructured_NLRF_linspace}
\end{figure}

Specializing to the case where $\S = \mathbf I$ we can obtain the results for the random feature model studied in \citet{mei2022generalization, adlam2020understanding}:
\begin{equation}
    \kappa_2 = \frac{\lambda}{(\tfrac{N}{D} - \overline{\df}_1)(1 - \frac{D}{P} \overline{\df}_1)}, \quad \overline{\df}_1 = \frac{1}{1+\kappa_2} + \frac{\sigma_\xi^2}{\kappa_2}.
\end{equation}
One can solve these equations self-consistently for $\kappa_2$. In the ridgeless limit, this gives:
\begin{equation}
    \kappa_2 = \frac{1+ \sigma_\xi^2 - \psi - \sqrt{(1+\sigma_\xi^2-\psi)^2+4 \psi \sigma_\xi^2}}{2 \psi},
\end{equation}
where $\psi = \min(P,N)/D$. Using that $\tf_1 = \df_1 = (1+\kappa_2)^{-1}$ and $\df_2 = (1+\kappa_2)^{-2}$ we recover the ridgeless expressions in \cite{adlam2020understanding, mei2022generalization}:
\begin{itemize}
    \item Underparameterized
    \begin{equation}
        E_g = \frac{1 - \frac{N}{D} -\sigma_\xi^2 + \sqrt{(1 - \frac{N}{D} + \sigma_\xi^2)^2 + 4 \frac{N}{D} \sigma_\xi^2}}{2 (1-N/P)} + \sigma_\epsilon^2 \frac{N/P}{1-N/P}.
    \end{equation}
   
    \item Overparameterized
    \begin{equation}
    \begin{aligned}
          E_g 
          &=  \frac{1 - \frac{P}{D} -\sigma_\xi^2 + \sqrt{(1 - \frac{P}{D} + \sigma_\xi^2)^2 + 4 \frac{P}{D} \sigma_\xi^2}}{2 (1-P/N)} 
          \\&\quad  + (\sigma_\epsilon^2 - \sigma_\xi^2) \left[\frac{1+\tfrac{P}{D} + \sigma_\xi^2 - \sqrt{(1-\tfrac{P}{D} + \sigma_\xi^2 )^2 + 4 \frac{P}{D} \sigma_\xi^2}}{2  \sqrt{(1-\tfrac{P}{D} + \sigma_\xi^2 )^2 + 4 \frac{P}{D} \sigma_\xi^2}}\right] + \sigma_\epsilon^2 \frac{P/N}{1-P/N}.
    \end{aligned}
    \end{equation}
\end{itemize}
We illustrate these solutions in Figure \ref{fig:unstructured_NLRF_linspace}.

\subsection{Fine-Grained Bias-Variance Decomposition}

We conclude with a fine-grained bias-variance decomposition of nonlinear random feature models in the case of isotropic features and feature noise, and structured input data. This extends work by \citet{adlam2020understanding}, who derived this decomposition for isotropic input data. Again, using the technology we've developed so far, these can be derived in a few lines of algebra, and straightforwardly interpreted. 

Averaging over the dataset involves an average over both $\X$ and $\bm \Xi$. This is the same as averaging $\overline \X$ in the linear random feature description. Thus, the equations of the prior section apply.  For a test point prediction, one has
\begin{equation}\label{eq:NLRF_bag_ens}
\begin{aligned}
    \mathbb E_{\overline \X, \overline \F, \bm \xi} \hat y &= \mathbb E_{\overline \X, \overline \F, \bm \xi} (\x^\top \F + \bm \xi^\top) \hat{\v} = \mathbb E_{\overline \X, \overline \F} \x^\top \F \hat{\v} \\
    &= \mathbb E_{\overline \X, \F} \x^\top \F (\overline \F^\top \overline \X^\top \overline \X \overline \F + \lambda \mathbf I)^{-1} \overline \F^\top \overline \X^\top (\X \w + \e)\\
    &= \mathbb E_{\F} \x^\top \F (\F^\top \S \F + \sigma_\xi^2 \mathbf I + \kappa_1 \mathbf I)^{-1} \F^\top \S \bar \w\\
    &= \x^\top  (\S + \kappa_2 \mathbf I)^{-1} \S \bar \w.
\end{aligned}
\end{equation}
This gives us as before:
\begin{equation}
    \mathrm{Bias}^2 = -\kappa_2^2 \tf_1'(\kappa_2).
\end{equation}
Similarly, one can average over just the data. This is an average over both $\X$ and $\bm \Xi$ as $\bm \Xi$ carries a data index. This gives
\begin{equation}
\begin{aligned}
    \mathbb E_{\X, \Xi} \hat y &= \mathbb E_{\X, \bm \Xi} (\x^\top \F + \bm \xi^\top) \hat{\v} \\
    &= \mathbb E_{\overline \X} \overline \x^\top \overline \F (\overline \F^\top \overline \X^\top \overline \X \overline \F + \lambda \mathbf I)^{-1} \overline \F^\top \overline \X^\top (\overline \X \bar \w_{D+N} + \e)\\
    &= \overline \x^\top \overline \F  \overline \F^T (\overline \F  \overline \F^T  + \kappa_1 \mathbf I)^{-1} \bar \w_{D+N}.
\end{aligned}
\end{equation}
We note that the noise drops out as before, so $\mathrm{Var}_{\F, \e} = 0$. We thus get:
\begin{equation}
\begin{aligned}
    \mathrm{Bias}^2 + \mathrm{Var}_{\X} &= \kappa_1^2 \bar \w_{D+N}^\top (\overline \F \overline \F^\top+\kappa_1 \mathbf I)^{-2} \bar \w_{D+N}\\
    &= -\kappa_1^2 \partial_{\kappa_1} \left[\frac{\kappa_1+\sigma_\xi^2}{\kappa_1} \bar \w^\top \S^{1/2} (\F^\top \S \F + \kappa_1  \mathbf I + \sigma_\xi^2 \mathbf I)^{-1} \S^{1/2} \bar \w \right]\\
    &= -\kappa_1^2 \partial_{\kappa_1} \left[\frac{\kappa_2}{\kappa_1} \tf_1(\kappa_2) \right].
\end{aligned}
\end{equation}
This is as before and thus yields:
\begin{equation}
    \mathrm{Var}_{\F} = \left(1 - \tfrac{d \log\kappa_2}{d \log \kappa_1} \right) \kappa_2 \tf_2(\kappa_2).
\end{equation}

Averaging over features is more subtle, since both the $\F$ and $\bm \Xi$ matrices are averaged over. It is better to write:
\begin{equation}
    \overline{\bm X} = \begin{pmatrix}
        \X & \S_\xi^{1/2}
    \end{pmatrix}, \quad \overline{\bm F} = \begin{pmatrix}
        \F\\
        \Z
    \end{pmatrix} \sim \mathcal N(0, \mathbf I_{N+P}).
\end{equation}
In this case we still have $\hat{\v} = (\overline \F^\top \overline  \X^\top \overline  \X \overline  \F + \lambda \mathbf I)^{-1} \overline \F^\top \overline \X^\top  \X \w$.
One can then evaluate the feature-averaged test set prediction as follows: 
\begin{equation}
\begin{aligned}
    \mathbb E_{\overline \F, \bm \xi} \hat y &=  \mathbb E_{\overline \F, \bm \xi} [\x^\top \F + \bm \xi^\top] \hat{\v} = \mathbb E_{\overline \F} \x^\top \bm \F \hat{\v} \\
    &=   \mathbb E_{\overline \F} \bm \Pi_D \overline \F (\overline \F^\top \overline  \X^\top \overline  \X \overline  \F + \lambda \mathbf I)^{-1} \overline \F^\top \overline \X^\top  (\X \bar \w + \e) \\
    &= \bm \Pi_D  (\overline \X^\top \overline \X + \kappa_{\F} \mathbf I)^{-1} \overline \X^\top  (\X \bar \w + \e), \quad \kappa_{\F} = \lambda S_{\F \F^\top}\\
    &=  (\X^\top \X + \sigma_\xi^2 \mathbf I + \kappa_{\F} \mathbf I)^{-1} \X^\top (\X  \bar \w + \e).
\end{aligned}
\end{equation}
Here, in the second line we have written $\F = \bm \Pi_{D} \overline \F$ as the projection onto the first $D$ components of $\overline \F$. 
This is again just ridge regression without random features and with ridge parameter $\kappa_{\F} + \sigma_\xi^2$. As before, after averaging over $\X$ this ridge will get renormalized to $\kappa_2$. We thus get:
\begin{equation}
    \mathrm{Var}_{\X} = \frac{\gamma_2}{1-\gamma_2} [-\kappa_2^2 \tf_1'], \quad \mathrm{Var}_{\X, \e} =  \frac{\gamma_2}{1-\gamma_2} \sigma_\epsilon^2.
\end{equation}
This consequently gives:
\begin{equation}
     \mathrm{Var}_{\X, \e} =  \left[\frac{\gamma_1}{1-\gamma_1} - \frac{\gamma_2}{1-\gamma_2} \right] \sigma_\epsilon^2.
\end{equation}
We thus recover the exact same form of the decomposition as in the linear random feature model setting. See Figure \ref{fig:LRF_BV} for a schematic illustration.

\subsection{\texorpdfstring{Scaling Laws in $P$ and $N$}{Scaling Laws in P and N}}

As in prior scaling law subsections, we consider $\S$ to have eigenvalues decaying as $\eta_k \sim k^{-\alpha}$, with $\alpha$ the capacity exponent. We consider the scaling of $\kappa_2$ as a function of $P, N$ in the ridgeless limit $\lambda \to 0$. Because
\begin{equation}
    \kappa_2 = \frac{\lambda}{(\frac{N}{D} - \overline{\df}_1) (\frac{P}{D} - \overline{\df}_1)}, 
\end{equation}
we must have that $\overline{\df}_1 \to \frac{\min(P, N)}{D}$. This implies
\begin{equation}
    \frac{\min(P, N)}{D} = \df_{\S}(\kappa_2) + \frac{\sigma^2_\xi}{\kappa_2}.
\end{equation}
If $\sigma_\xi^2$ is negligible, we have that $\overline{\df}_1 \approx \df^1_{\S}$, giving $\kappa_2 \sim \min(P, N)^{-\alpha}$ as in Section \ref{sec:lrf_all_scaling}. Then, all of the results of that Section  apply. On the other hand, if the second term dominates, then $\kappa_2 \sim \frac{D}{\min(P, N)} \sigma_\xi^2$. 
Schematically, the transition from one behavior to the other will occur when:
\begin{equation}\label{eq:NLRF_varlim_condition}
    \frac{\sigma_\xi^2}{\kappa_2} \sim \frac{\min(P, N)}{D} \Rightarrow \min(P, N) \gg (\sigma_\xi^2 D)^{-1/(\alpha-1)}.
\end{equation}
One can consider scaling $\sigma_\xi$ with $D$ so that $ \tilde \sigma_\xi^2 \equiv \sigma_\xi^2 D$ is a constant. Under this scaling, when condition \eqref{eq:NLRF_varlim_condition} is met, we get that $\kappa_2 \sim \min(P, N)^{-1} \tilde \sigma_\xi^2$. 

\begin{figure}[t]
    \centering
    \subfigure[$\kappa_1, \kappa_2$ scaling]{\includegraphics[height=2.3in]{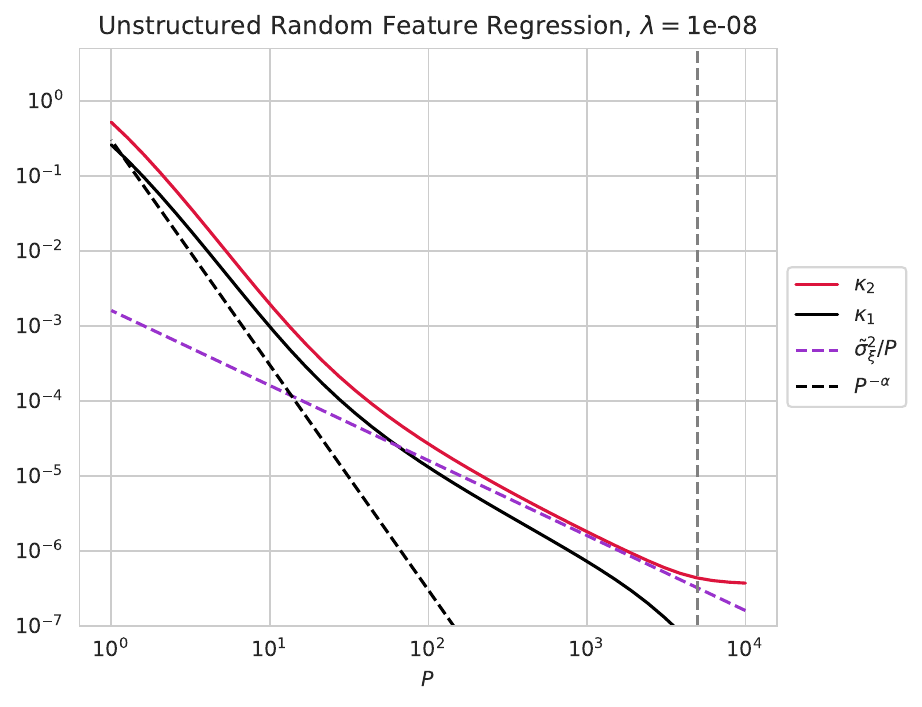}}
    \subfigure[$\bm \xi$-dominated]{\includegraphics[height=2.3in]{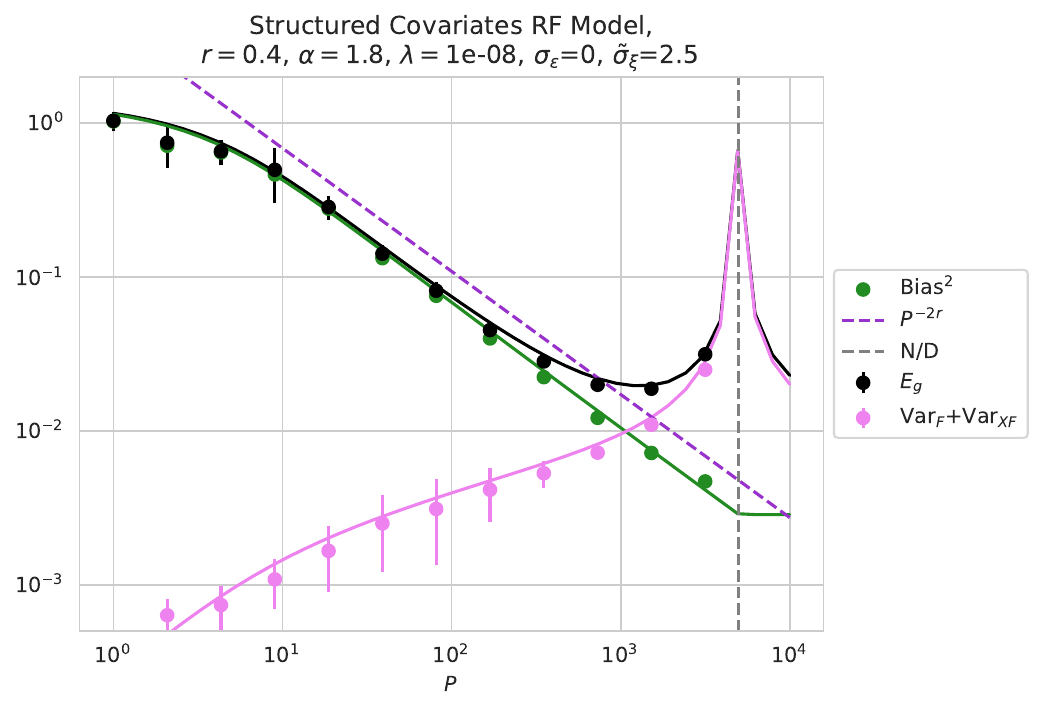}}
    \subfigure[$\bm \xi, \mathrm{Var}_{\F}$-dominated]{\includegraphics[height=2.3in]{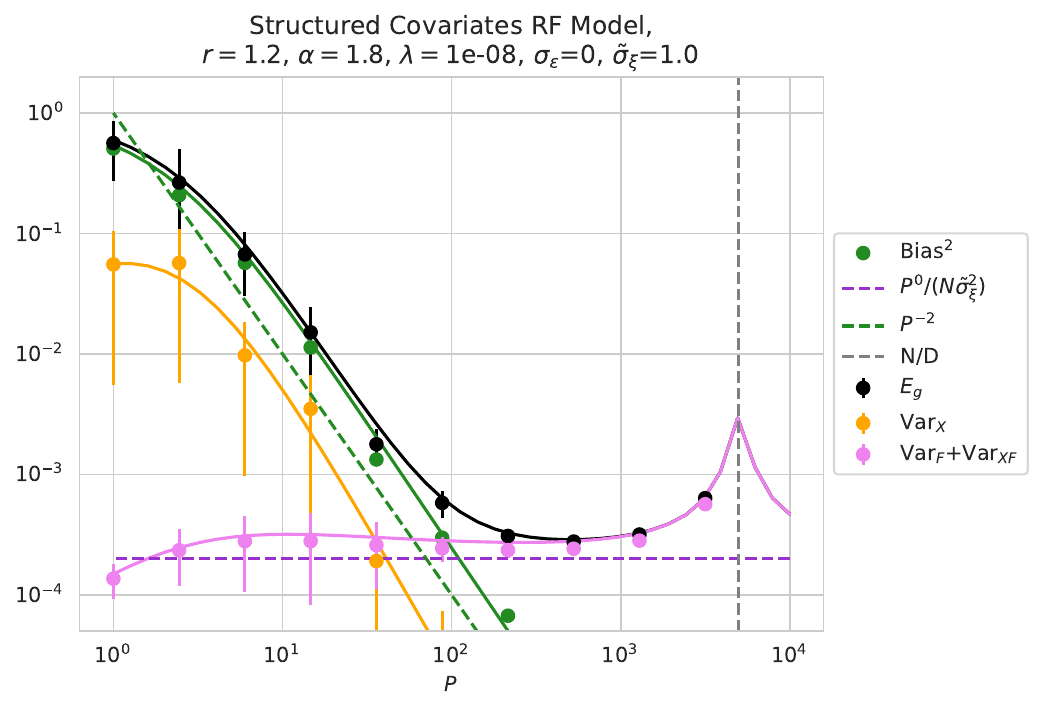}}
    \subfigure[Noise $\bm \xi$-mitigated]{\includegraphics[height=2.3in]{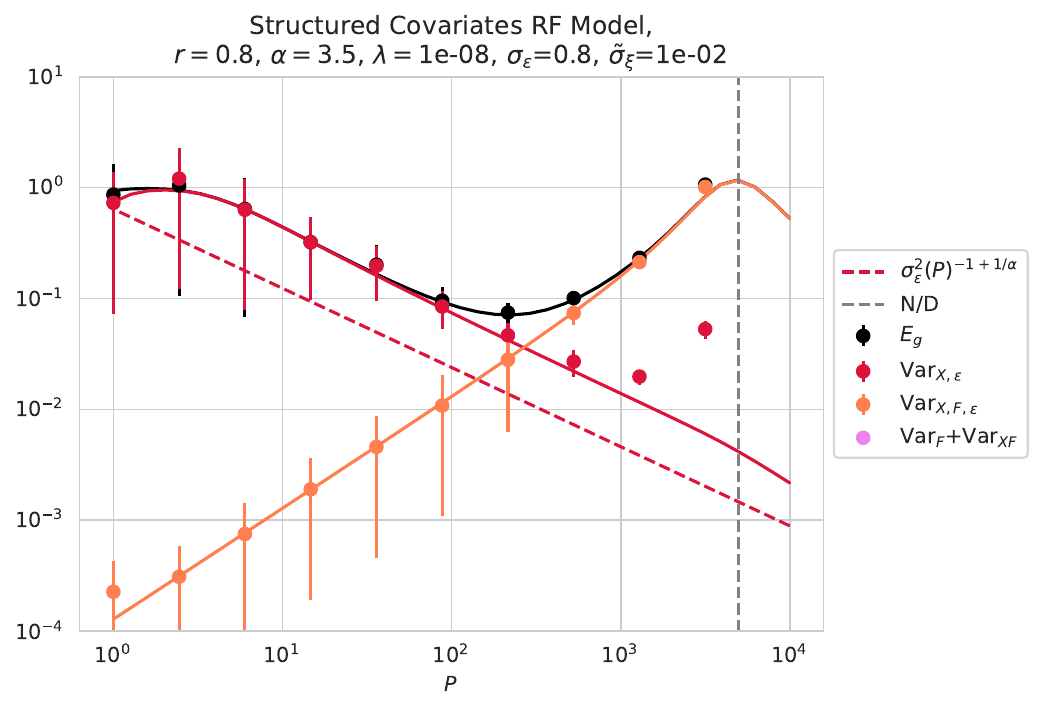}}
    
    \caption{a) The transition between $\kappa_2 \sim P^{-\alpha}$ and $\kappa_2 \sim \tilde \sigma_\xi^2/P$ in the overparameterized regime. b) Illustration of $\bm \xi$-dominated scaling. The dashed purple line is the power law exponent prediction. c) The transition from $\bm \xi$-dominated to joint $\bm \xi, \mathrm{Var}_{\F}$-dominated scaling when $r > 1/2$. We see a plateau, with an estimate given by the dashed purple line from scaling arguments. This is identical to the plateau studied in the model in \citet{atanasov2022onset}. d) The feature noise $\bm \xi$ can act as an effective ridge $\lambda \sim 1/P$ and thus mitigate the effect of noise. This gives a nontrivial scaling with $P$ in the presence of noise rather than a plateau. However, in the absence of explicit ridge, there is also a subsequent double descent peak. Our fine-grained bias-variance decomposition shows that this is due explicitly to the joint variance, $\mathrm{Var}_{\X \F \e}$. Near the double descent peak the empirics are less numerically stable, leading to slight deviation from theory curves. In all cases we take a width $N=5000$ random feature model and bag over 25 datasets and ensemble over 25 initializations.}
    \label{fig:NLRF_scalings}
\end{figure}
This then gives the following scalings in the underparameterized regime $N < P$: 
\begin{equation}\label{eq:NLRF_all_scalings}
     E_g \sim \begin{cases}
        \displaystyle \frac{ (N/\tilde \sigma_\xi^2)^{-2 \min(r, 1/2)}}{1- N/P}, \quad \, (N/\tilde \sigma_\xi^2)^{-2 \min(r, 1/2)} \gg \sigma_\epsilon^2 (N/\tilde \sigma_\xi^2)^{1/\alpha}/P  & \text{$\bm \xi$ dominated}\\
        \displaystyle \sigma_\epsilon^2 (N/\tilde \sigma_\xi^2)^{1/\alpha}/P \hspace{0.45in} \, (N/\tilde \sigma_\xi^2)^{-2 \min(r, 1/2)} \ll \sigma_\epsilon^2 (N/\tilde \sigma_\xi^2)^{1/\alpha}/P &  \text{Noise $\bm \xi$-mitigated}
        \end{cases}
\end{equation}
Similarly, in the overparameterized $P > N$ regime we have: 
\begin{equation}\label{eq:NLRF_all_scalings2}
     E_g \sim \begin{cases}
        \left(P/\tilde \sigma_\xi^2 \right)^{-2 \min(r, 1)}, \hspace{0.1in} P \ll P_{\e};\, r \leq 1/2 \text{ or } P \ll P_{\F}  & \text{$\bm \xi$ dominated}\\
        \displaystyle  P^0/(N \tilde \sigma_\xi^2), \hspace{0.55in}  P \ll P_{\e}; r > 1/2; P \gg P_{\F} & \text{Joint $\bm \xi, \mathrm{Var}_{\F}$ dominated} \\
        \displaystyle \sigma_\epsilon^2 \left(P/\tilde \sigma_\xi^2\right)^{-\frac{\alpha-1}{\alpha}},\hspace{0.2in}  P \gg P_{\e} & \text{Noise $\bm \xi$-mitigated}\\
    \end{cases}
\end{equation}
We demonstrate examples of these scalings in Figure \ref{fig:NLRF_scalings}.

\clearpage
\section{Conclusion}

By using $S$-transform subordination relations, we have given compact derivations for the generalization error, training error, and fine-grained bias-variance decomposition across a variety of high-dimensional regression models. These include linear regression, kernel methods, linear random feature models, and nonlinear random feature models. We also studied the scaling properties of these models in the setting where the input covariates and target weights had power law structure. We derived novel formulas for the generalization error of a very generic class of random feature models and for all the sources of variance in that setting. 

These results culminated in the enumeration of possible scaling regimes for deep linear random feature models in Section \ref{sec:lrf_all_scaling}. As illustrated in the phase diagrams in Figure \ref{fig:rate_plots}, this gives rise to a rich portrait of which sources of bias and variance generate particular scaling laws given particular structure in the task and random feature weights. This allowed us to interpret a novel scaling regime found in overparameterized random feature models as due to the limiting behavior of parameter variance. We extended this analysis to shallow nonlinear random feature models with structured input data though Gaussian equivalence principles. Thus, assuming Gaussian universality, the only class of random feature models whose possible scaling regimes are not enumerated here are deep nonlinear random feature models with structured weights. We leave this to future work. 

How does this diversity of scaling regimes in linear models relate to those observed in deep neural networks? Transitions between regimes with trivial and distinct non-trivial scaling exponents with increasing dataset and model size have been observed in a variety of deep networks \cite{vyas2024feature,atanasov2022onset}. In particular, past works have documented the existence of variance-dominated scaling in deep networks \cite{atanasov2022onset}, which we show here occurs ubiquitously in both linear and nonlinear random feature models. A broader feature of the study of scaling laws in linear models is that non-trivial scaling exponents are not universal; rather, they depend strongly on the structure of the data and of the target function. This is broadly consistent with observations in deep networks, where scaling exponents vary across language and vision tasks \cite{kaplan2020scaling,anwar2024foundational,hestness2017deep,hoffmann2022training,besiroglu2024chinchilla,muennighoff2024scaling,alabdulmohsin2024getting,zhai2022scaling,bachmann2024scaling}. 

The multiplicative property of the $S$-transform makes it a particularly powerful tool for analyzing the structure of covariances given by passing data through layers of features. It allows for most formulae in the literature on random feature models to be derived in a succinct, unified fashion. Beyond the proportional regime, or in a feature-learning regime where features in all layers become correlated with themselves and with the data, the free probability assumptions necessary to apply the $S$-transform almost certainly break down. It will be interesting to investigate to what extent methods in random matrix theory and free probability can still be adapted to this setting, and what additional technology will need to be developed to study scaling laws in the feature learning regime. 

\section*{Acknowledgments}

We thank Blake Bordelon, Hamza Chaudhry, and Paul Masset for inspiring conversations. We also thank Blake Bordelon, Hamza Chaudhry, Sabarish Sainathan, and especially Benjamin Ruben for helpful comments on earlier versions of this manuscript. 
ABA is grateful to Bruno Loureiro, Alex Maloney, and Jamie Simon for helpful discussions on random matrices, deterministic equivalence, and diagrammatics at the Aspen Center for Theoretical Physics Winter Program on Theoretical Physics for Machine Learning. ABA also thanks Galit Anikeeva for useful discussions on bias-variance decompositions. Finally, we thank Bruno Loureiro and Courtney Paquette for useful discussions regarding scaling regimes at the DIMACS Workshop on Modeling Randomness in Neural Network Training, held at the DIMACS Center at Rutgers University.

ABA is supported by the Professor Yaser S. Abu-Mostafa Fellowship from the Fannie and John Hertz Foundation. JAZV and CP were supported by NSF Award DMS-2134157 and NSF CAREER Award IIS-2239780. JAZV is presently supported by a Junior Fellowship from the Harvard Society of Fellows. CP is further supported by a Sloan Research Fellowship. This work has been made possible in part by a gift from the Chan Zuckerberg Initiative Foundation to establish the Kempner Institute for the Study of Natural and Artificial Intelligence. This research was supported in part by grants NSF PHY-1748958 and PHY-2309135 to the Kavli Institute for Theoretical Physics (KITP), through the authors' participation in the Fall 2023 program ``Deep Learning from the Perspective of Physics and Neuroscience.''

\clearpage 

\appendix

\section{Diagrammatic Derivations of Subordination Relations} \label{sec:diagrams}

In this Appendix, we give a self-contained derivation of the subordination relations \eqref{eq:subordination1}, \eqref{eq:subordination2}, and \eqref{eq:subordination3}, along with a brief overview of the aspects of free probability theory as applied to random matrices that we use in this paper. For the interested reader, there are many more extensive introductory texts, including \citet{voiculescu1997free,nica2006lectures, mingo2017free, potters2020first}. 

\subsection{Definition of Freedom}

Free probability studies non-commutative random variables. The simplest statistic that distinguishes free probability from commutative probability is the joint fourth moment of two random variables. Consider two random matrices $\A_1$, $\A_2$ with $\tr[\A_i] \simeq 0$ in the limit $N \to \infty$. If $\A_1, \A_2$ are \textbf{free} of one another, one consequence is that
\begin{equation}
    \mathbb E_{\A_1, \A_2} \tr[\A_1 \A_2 \A_1 \A_2] \simeq 0.
\end{equation}
Note that this fourth moment certainly would not vanish for nonzero commutative random variables. In free probability, when two mean zero random variables $\A, \B$ are free of one another, their alternating moments will vanish. One consequence of this is that a sum of free random variables has lower kurtosis. This is a reason why the Wigner semicircle law (the analog of the Gaussian in free random matrix theory; see \sectionsymbol\ref{sec:wigner}) has lower kurtosis than the Gaussian and is in fact compactly supported. 

We now formally define what it means for a collection of random variables to be \textbf{jointly free}. Though the theory of free probability extends to more general algebras \cite{voiculescu1997free}, here we will focus only on the case of \textbf{asymptotically free random matrices},\footnote{The reader should distinguish this from the notion of asymptotic freedom in gauge theory. } i.e., $N \times N$ matrices which behave as free random variables in the limit $N \to \infty$, as that is the setting which is relevant for the present work \cite{mingo2017free}. As we work in the $N \to \infty$ limit throughout, we will frequently drop the qualifier ``asymptotic'' and simply state that certain random matrices are free. 

Joint (asymptotic) freedom of a set of $n$ random matrices $\{\A_i\}_{i=1}^n$ of size $N \times N$ is defined by considering all mixed moments of these random variables in the limit $N \to \infty$. Take a set of $m$ polynomials $\{p_{k}\}_{k=1}^{m}$ and a labeling $\{i_k\}_{k=1}^n$ with each $i_k \in \{1, \dots , n\}$ so that $i_{k} \neq i_{k+1}$ for all $k$. Let each $p_k$ have the property that 
\begin{align}
    \tr[ p_k(\A_{i_k}) ] \simeq 0.
\end{align}
Then $\{\A_i\}_{i=1}^n$ are jointly asymptotically free if and only if
\begin{equation}
    \tr \left[ p_1(\A_{i_1}) \cdots p_m(\A_{i_m})  \right] \simeq 0
\end{equation}
for any $m$ and labeling $\{i_k\}$ and set of polynomials $\{p_k\}_{k=1}^{m}$ satisfying the mean zero property above. Independent draws from the classical random matrix ensembles we consider are all jointly asymptotically free, as they can be randomly rotated relative to one another. The normalized traces concentrate to deterministic values for all ensembles we consider.

\subsection{\texorpdfstring{$R$-Transform Subordination}{R-Transform Subordination}}\label{sec:R_diagrams}

In this Appendix, we give a self-contained diagrammatic derivation of the $R$-transform subordination relation 
\begin{align}
     \mathbb E_{\B} \bm G_{\A + \B}(z) &\simeq \G_{\A} (z- R_{\B}(g_{\A + \B}(z))) ,
\end{align}
listed as \eqref{eq:subordination1} in the main text. We will consider the case where $\A$ is deterministic and $\B$ is random and drawn from a rotation-invariant distribution. In the large $N$ limit, the spectra of both $\A, \B$ will be deterministic. We then have $\B = \O \B' \O^\top$ where $\B'$ is a deterministic diagonal matrix. Then, to average over $\B$, we only need to evaluate the average over the relative rotation matrix $\O$.
 
We perform the following expansion of $\G_{\A + \B}$:
\begin{equation}
    \mathbb E_{\O} \bm G_{\A + \O \B' \O^\top}(z) = \mathbb E_{\O} \left[ \G_{\A}(z) + \G_{\A}(z) \O \B' \O^\top \G_{\A}(z)  +\cdots   \right].
\end{equation}
We use solid dots to denote insertions of $\O \B' \O^\top$ and solid lines to denote contraction with $\G_{\A}(z)$. A general term in this series will look like:
\begin{center}
\begin{tikzpicture}
    \def\spacing{3cm}
    \def\linewidth{0.5mm} %
  \foreach \x in {0,...,2}
    \filldraw[black] (\x*\spacing,0) circle (3pt) node[below] {$\O \B' \O^\top$};

  \foreach \x in {-1,...,2}
    \draw[line width=\linewidth] (\x*\spacing,0) -- node[below] {$\G_{\A}$} (\x*\spacing+\spacing,0);

  \node[below] at (-1*\spacing,0) {};
  \node[below] at (3*\spacing,0) {};
\end{tikzpicture}
\end{center}
We now perform the average over $\O$. 
We will not have to do any explicit calculations. Rather, we observe the following facts:
\begin{enumerate}
    \item Because the entries of an orthogonal matrix have average size $N^{-1/2}$, a correlator of $2n$ $\O$ matrices has the scaling:
    \begin{equation}
        \mathbb E_{\O} [\O_{i_1 j_1} \cdots \O_{i_{2n} j_{2n}}] \sim O(N^{-n}).
    \end{equation}
    \item At leading order in $N$, the $\O$ behave like matrices with independent Gaussian entries. This allows us to compute averages by Wick contractions, also known as Isserlis' theorem: 
    \begin{equation}\label{eq:isserlis}
        \mathbb E[\O_{i_1 j_1} \dots \O_{i_{2n} j_{2n}}] = N^{-n} \sum_{\text{pairings } P} \prod_{(k, k') \in P} \delta_{i_{k} i_{k'}} \delta_{j_k, j_{k'}} + \text{subleading terms}.
    \end{equation}
    Here the $i$ and $j$ indices have the same pairing in each term. The subleading terms contributing to higher cumulants are known as Weingarten contributions, which have been the subject of considerable past study \cite{weingarten1978asymptotic, brouwer1996diagrammatic, collins2009some, banica2010orthogonal}. Although they will enter into our calculations, we will not need to know precise details about their forms. See Chapter 12 of \citet{potters2020first} for details.
\end{enumerate}
We will denote the expectation of this over $\O$ by dashed lines. Consider first the quantity $\O \B' \O^\top$:
\begin{equation}
    \mathbb E_{\O} [\O  \B' \O^\top] \equiv \hspace{0.2cm} \begin{tikzpicture}[baseline=-1.0ex]
    \def\spacing{3cm}
   \filldraw[black] (0 ,0) circle (3pt) node[below] {$\O \B' \O^\top$};
  \draw[dashed] (0,0) arc[start angle=-90, end angle=270, radius=\spacing/4];
\end{tikzpicture}
\end{equation}
One can evaluate this expectation by appeal to symmetry alone. First, because the distribution of $\O$ is invariant under an orthogonal transformation $\O \mapsto \bm{U}_{L} \O$ for any orthogonal matrix $\bm{U}_{L}$, the final result must be rotationally invariant, and therefore proportional to the identity matrix $\mathbf{I}$. Second, because the distribution of $\O$ is invariant under an orthogonal transformation $\O \mapsto \O \bm{U}_{R}$ for any orthogonal matrix $\bm{U}_{R}$, the expectation must depend only on the eigenvalues of $\B'$, and the dependence must be linear. Finally, when $\B' = \mathbf I$, it is equal to $\mathbf I$. This uniquely determines this quantity to be:
\begin{equation}
     \C_1 \equiv \mathbb E_{\O} [\O  \B' \O^\top] = \tr[\B] \mathbf I.
\end{equation}
This agrees with just directly applying Equation \eqref{eq:isserlis}. However, the  above argument is true for all $N$, not just at leading order. 
We now make an observation about traces:
\begin{enumerate}
  \setcounter{enumi}{2}
    \item Each loop in the diagrams corresponds to a free index that is traced over. Converting from a trace to a normalized trace (which is order $1$) leaves over a factor of $N$.
\end{enumerate}
Thus, to get an $O(1)$ contribution from a correlator of $2n$ matrices $\O$, we need diagrams with $n$ loops to contribute $n$ factors of $N$ to cancel out the $N^{-n}$ scaling. Diagrams with fewer loops will be suppressed in the large $N$ limit. This will mean that crossing diagrams are not counted.

Next, using shorthand $\B = \O \B' \O^\top$, consider the second moment $\mathbb E_{\O} [\B \G_{\A} \B]$. We can write this as two pieces:
\begin{equation}
\begin{aligned}
    \mathbb E_{\O} [\B \G_{\A} \B] = \left( \mathbb E_{\O} [\B \G_{\A} \B] - \mathbb E_{\O} [\B] \G_{\A} \mathbb E_{\O}[\B]  \right)  +  \mathbb E_{\O} [\B] \G_{\A} \mathbb E_{\O}[\B]
\end{aligned}
\end{equation}
We call the first term this \textbf{connected} term and the second term the \textbf{disconnected} term. Graphically, we will write this as
\begin{equation}
\mathbb E_{\O} [\B \G_{\A} \B] = 
\begin{tikzpicture}[scale=0.8, baseline=-1.0ex]
    \def\spacing{3cm}
    \def\shift{0.1cm}
    \def\linewidth{0.5mm} %

  \fill[gray!30] (1*\spacing +  \shift,0) arc[start angle=0, end angle=180, radius=\spacing/2 + \shift];

    \fill[white] (1*\spacing -  \shift,0) arc[start angle=0, end angle=180, radius=\spacing/2 - \shift];

  \draw[dashed] (1*\spacing - \shift,0) arc[start angle=0, end angle=180, radius=\spacing/2- \shift];
  \draw[dashed] (1*\spacing +  \shift,0 ) arc[start angle=0, end angle=180, radius=\spacing/2+\shift];

\foreach \x in {0,...,1}
    \filldraw[black] (\x*\spacing,0) circle (3pt) node[below] {$\O \B' \O^\top$};

  \foreach \x in {0}
    \draw[line width=\linewidth] (\x*\spacing,0) -- node[below] {$\G_{\A}$} (\x*\spacing+\spacing,0);
\end{tikzpicture}
+
\begin{tikzpicture}[scale=0.8, baseline=-1.0ex]
    \def\spacing{3cm}
    \def\shift{0.1cm}
    \def\linewidth{0.5mm} %
  \draw[dashed] (0, 0) arc[start angle=-90, end angle=270, radius=\spacing/4];
  \draw[dashed] (\spacing,0) arc[start angle=-90, end angle=270, radius=\spacing/4];
  
\foreach \x in {0,...,1}
    \filldraw[black] (\x*\spacing,0) circle (3pt) node[below] {$\O \B' \O^\top$};

  \foreach \x in {0}
    \draw[line width=\linewidth] (\x*\spacing,0) -- node[below] {$\G_{\A}$} (\x*\spacing+\spacing,0);
  
\end{tikzpicture}
\end{equation}
This is analogous to how a moment is equal to a given cumulant plus contributions from lower order cumulants. Here, we have shaded the first diagram to highlight that it includes both the Wick contraction as well as a potential contribution from the fourth cumulant of orthogonal matrices:
\begin{equation}\label{eq:2pt_fc}
\C_2[\G_{\A}] \equiv \hspace{-.05in}
    \begin{tikzpicture}[scale=0.8, baseline=-1.0ex]
    \def\spacing{3cm}
    \def\shift{0.1cm}
    \def\linewidth{0.5mm} %

  \fill[gray!30] (1*\spacing +  \shift,0) arc[start angle=0, end angle=180, radius=\spacing/2 + \shift];

    \fill[white] (1*\spacing -  \shift,0) arc[start angle=0, end angle=180, radius=\spacing/2 - \shift];

  \draw[dashed] (1*\spacing - \shift,0) arc[start angle=0, end angle=180, radius=\spacing/2- \shift];
  \draw[dashed] (1*\spacing +  \shift,0 ) arc[start angle=0, end angle=180, radius=\spacing/2+\shift];

\foreach \x in {0,...,1}
    \filldraw[black] (\x*\spacing,0) circle (3pt) node[below] {$\O \B' \O^\top$};

  \foreach \x in {0}
    \draw[line width=\linewidth] (\x*\spacing,0) -- node[below] {$\G_{\A}$} (\x*\spacing+\spacing,0);
\end{tikzpicture} \hspace{-.05in} = \hspace{-.05in}
\begin{tikzpicture}[scale=0.8, baseline=-1.0ex]
    \def\spacing{3cm}
    \def\shift{0.1cm}
    \def\linewidth{0.5mm} %

  \draw[dashed] (1*\spacing - \shift,0) arc[start angle=0, end angle=180, radius=\spacing/2- \shift];
  \draw[dashed] (1*\spacing +  \shift,0 ) arc[start angle=0, end angle=180, radius=\spacing/2+\shift];

\foreach \x in {0,...,1}
    \filldraw[black] (\x*\spacing,0) circle (3pt) node[below] {$\O \B' \O^\top$};

  \foreach \x in {0}
    \draw[line width=\linewidth] (\x*\spacing,0) -- node[below] {$\G_{\A}$} (\x*\spacing+\spacing,0);
\end{tikzpicture}
\hspace{-.05in}
+ 
\hspace{-.05in}
\begin{tikzpicture}[scale=0.8, baseline=-1.0ex]
    \def\spacing{3cm}
    \def\shift{0.1cm}
    \def\linewidth{0.5mm} %

  \draw[dashed] (1*\spacing - \shift,0) arc[start angle=0, end angle=180, radius=\spacing/2- \shift];
  \draw[dashed] (1*\spacing +  \shift,0 ) arc[start angle=0, end angle=180, radius=\spacing/2+\shift];

\foreach \x in {0,...,1}
    \filldraw[black] (\x*\spacing,0) circle (3pt) node[below] {$\O \B' \O^\top$};

  \foreach \x in {0}
    \draw[line width=\linewidth] (\x*\spacing,0) -- node[below] {$\G_{\A}$} (\x*\spacing+\spacing,0);

    \node[draw, circle, minimum size=15pt, inner sep=0pt, fill=gray!30] at (0.5*\spacing, \spacing/2) {4};

\end{tikzpicture}
\end{equation}
Here, the first term is a Wick contraction, giving a term proportional to $\tr[\G_{\A}] \tr[\B^2]$. We have not included the crossing Wick contraction because it will not contribute at large $N$, as discussed above. The second term corresponds the fourth cumulant of the $\O$s. This is a subleading Weingarten term in Equation \eqref{eq:isserlis}. The only way that it might contribute is if it has at least $3$ traces. Thus, if it enters, it must enter as $\tr[\G_{\A}] \tr[\B]^2$.

At third order we will have several terms involving connected and disconnected components. 
One such term is:
\begin{equation}\label{eq:3pt_red}
\C_2[\G_{\A}] \G_{\A} \C_1 = 
\begin{tikzpicture}[baseline=-1.0ex]
    \def\spacing{3cm}
    \def\shift{0.1cm}
    \def\linewidth{0.5mm} %

  \fill[gray!30] (1*\spacing +  \shift,0) arc[start angle=0, end angle=180, radius=\spacing/2 + \shift];

    \fill[white] (1*\spacing -  \shift,0) arc[start angle=0, end angle=180, radius=\spacing/2 - \shift];

  \draw[dashed] (1*\spacing - \shift,0) arc[start angle=0, end angle=180, radius=\spacing/2- \shift];
  \draw[dashed] (1*\spacing +  \shift,0 ) arc[start angle=0, end angle=180, radius=\spacing/2+\shift];
  \draw[dashed] (2*\spacing,0) arc[start angle=-90, end angle=270, radius=\spacing/4];

\foreach \x in {0,...,2}
    \filldraw[black] (\x*\spacing,0) circle (3pt) node[below] {$\O \B' \O^\top$};

  \foreach \x in {0,...,1}
    \draw[line width=\linewidth] (\x*\spacing,0) -- node[below] {$\G_{\A}$} (\x*\spacing+\spacing,0);
\end{tikzpicture}
\end{equation}
Another such term is
\begin{equation}\label{eq:3pt_irred}
\C_2[\G_{\A} \C_1 \G_{\A}] =
\begin{tikzpicture}[scale=0.8][baseline=-1.0ex]
    \def\spacing{3cm}
    \def\shift{0.1cm}
    \def\linewidth{0.5mm} %

  \fill[gray!30] (2*\spacing +  \shift,0) arc[start angle=0, end angle=180, radius=\spacing + \shift];

    \fill[white] (2*\spacing -  \shift,0) arc[start angle=0, end angle=180, radius=\spacing - \shift];

  \draw[dashed] (2*\spacing - \shift,0) arc[start angle=0, end angle=180, radius=\spacing- \shift];
  \draw[dashed] (2*\spacing +  \shift,0 ) arc[start angle=0, end angle=180, radius=\spacing+\shift];
  \draw[dashed] (1*\spacing,0) arc[start angle=-90, end angle=270, radius=\spacing/4];
\foreach \x in {0,...,2}
    \filldraw[black] (\x*\spacing,0) circle (3pt) node[below] {$\O \B \O^\top$};

  \foreach \x in {0,...,1}
    \draw[line width=\linewidth] (\x*\spacing,0) -- node[below] {$\G_{\A}$} (\x*\spacing+\spacing,0);
\end{tikzpicture}
\end{equation}
The fully connected term is denoted by $\C_{3}[\G_{\A}, \G_{\A}]$ with
\begin{equation}\label{eq:3pt_fc}
\C_3[\A_1, \A_2] \equiv
\begin{tikzpicture}[scale=0.8, baseline=-1.0ex]
    \def\spacing{3cm}
    \def\shift{0.1cm}
    \def\linewidth{0.5mm} %

    \fill[gray!30] (2*\spacing + \shift,0) arc[start angle=0, end angle=180, radius=\spacing + \shift];

     \fill[white]  (2*\spacing -  \shift,0 ) arc[start angle=0, end angle=180, radius=\spacing/2 - \shift];
     \fill[white]  (\spacing -  \shift,0 ) arc[start angle=0, end angle=180, radius=\spacing/2 - \shift];

  \foreach \x in {0,...,2}
    \filldraw[black] (\x*\spacing,0) circle (3pt) node[below] {$\O \B' \O^\top$};

  \foreach \x in {1,...,2}
    \draw[line width=\linewidth] (\x*\spacing-\spacing,0) -- node[below] {$\A_{\x}$} (\x*\spacing,0);

  \draw[dashed] (2*\spacing + \shift,0) arc[start angle=0, end angle=180, radius=\spacing + \shift];
  \draw[dashed] (2*\spacing -  \shift,0 ) arc[start angle=0, end angle=180, radius=\spacing/2 - \shift];
  \draw[dashed] (\spacing -  \shift,0 ) arc[start angle=0, end angle=180, radius=\spacing/2 - \shift];
\end{tikzpicture}
\end{equation}
This will be the sum of the Wick contractions, plus the fourth cumulant contributions that correlate together at least one  $\O$ from each $\O \B' \O^\top$ insertion, plus the potential sixth cumulant contributions. Again, because the only way these subleading cumulants can contribute is by introducing additional traces, we'll have that this quantity will depend on $\A_1, \A_2$ only through $\tr[\A_1] \tr[\A_2]$.

We will call diagrams that cannot be reduced to two independently-taken averages \textbf{irreducible}. Diagrams \eqref{eq:2pt_fc}, \eqref{eq:3pt_irred}, \eqref{eq:3pt_fc} are all irreducible while \eqref{eq:3pt_red} is not. 
We will call the diagrams corresponding to $\C_1, \C_2, \C_3$ etc \textbf{fully connected}. Diagram \eqref{eq:3pt_irred} is irreducible but not fully connected.  We denote $n$-point fully connected diagram by $\C_n[\A_1, \dots, \A_{n-1}]$. The $\A_{i}$ are the matrices that appear below the arcs.  For our purposes, it is enough to know the following facts:
\begin{enumerate}
\setcounter{enumi}{3}
    \item The $n$-point fully connected diagram depends on the $\A_i$ only through the product of their traces $\prod_{i=1}^{n-1} \tr[\A_i]$. At the level of Wick contractions this is clear, where $\C_n$ goes as $\tr[\B^n] \prod_{i=1}^{n-1} \tr[\A_i]$. Subleading terms will only serve to further split $\tr[\B^n]$ into additional traces over $\B$.

    \item Dually, by tracing $\C_{n}$ against a test matrix $\A_n$, we have that this can depend only on $\A_n$ through $\tr[\A_n]$. This implies that $\C_n \propto \mathbf I$. 
    Together with iv), this implies:
        \begin{equation}\label{eq:R_principles}
        \C_n[\A_1, \dots, \A_{n-1}] = \kappa_{\B}^{(n)} \tr[\A_1] \cdots \tr[\A_{n-1}] \mathbf I
    \end{equation}
    for some constant $\kappa_{\B}^{(n)}$ that depends only on $\B$ which we call the \textbf{$n$th free cumulant} of $\B$. The reasons for this will become clear shortly.
\end{enumerate}

Because crossing diagrams do not contribute, we can notice a pattern. Each term in the series can be broken up into a string of irreducible diagrams connected together by $\G_{\A}$. 
\begin{center}
\begin{tikzpicture}[scale=0.8]
    \def\spacing{3cm}
    \def\shift{0.1cm}
    \def\linewidth{0.5mm} %

    \fill[gray!40] (1*\spacing,0) arc[start angle=0, end angle=180, radius=\spacing/2] -- cycle;
   \fill[gray!40] (3*\spacing, 0) arc[start angle=0, end angle=180, radius=\spacing/2] -- cycle;
   
  \draw[dashed] (1*\spacing +  \shift,0 ) arc[start angle=0, end angle=180, radius=\spacing/2+\shift];

   \draw[dashed] (3*\spacing +  \shift,0 ) arc[start angle=0, end angle=180, radius=\spacing/2+\shift];
   
   \node at ($(0*\spacing,0)!0.5!(1*\spacing,0)+(0,\spacing/5)$) {$\R$};
   \node at ($(2*\spacing,0)!0.5!(3*\spacing,0)+(0,\spacing/5)$) {$\R$};
   
  \foreach \x in {0,...,3}
    \filldraw[black] (\x*\spacing,0) circle (3pt) node[below] {};

  \foreach \x in {-1,1,3}
    \draw[line width=\linewidth] (\x*\spacing,0) -- node[below] {$\G_{\A}$} (\x*\spacing+\spacing,0);
  \foreach \x in {0, 2}
    \draw[line width=\linewidth] (\x*\spacing,0) -- node[below] {} (\x*\spacing+\spacing,0);

  \node[below] at (-1*\spacing,0) {};
  \node[below] at (3*\spacing,0) {};
\end{tikzpicture}
\vspace{-.2cm}
\end{center}
The matrix $\bm R$ 
is analogous to the \textbf{1 Particle Irreducible} diagrams or \textbf{Self-Energy} in physics that contribute to a mass shift.  We can then resum this series:
\begin{equation}
\begin{aligned}
    \mathbb E_{\O} \bm G_{\A + \O \B \O^\top}(z) &\simeq \G_{\A}(z) +  \G_{\A}(z) \R  \G_{\A}(z) + \G_{\A}(z) \R \G_{\A}(z) \R \G_{\A}(z) + \dots\\
    &= (z \mathbf I - \A - \R)^{-1}.
\end{aligned}
\end{equation}
It remains to compute $\R$. We get the following sum over fully-connected diagrams: \vspace{-.2cm}
\begin{equation}\label{eq:diagram_sum}
\begin{aligned}
    \R = & \hspace{-1.5cm}
    \begin{tikzpicture}[baseline=-0.65ex, scale=0.8]
        \def\spacing{3cm}
        \def\shift{0.1cm}
        \def\linewidth{0.5mm}
        \foreach \x in {0}
            \filldraw[black] (\x*\spacing,0) circle (3pt) node[below] {$\O \B' \O^\top$};

        \draw[dashed] (0 ,0 ) arc[start angle=-90, end angle=270, radius=\spacing/4 ];

        \node[below] at (-1*\spacing,0) {};
        \node[below] at (3*\spacing,0) {};
    \end{tikzpicture}
      \hspace{-6.5cm} +  \hspace{-2cm}
    \begin{tikzpicture}[baseline=-0.65ex, scale=0.9]
        \def\spacing{3cm}
        \def\shift{0.1cm}
        \def\linewidth{0.5mm} 
         \fill[gray!30] (1*\spacing +  \shift,0) arc[start angle=0, end angle=180, radius=\spacing/2 + \shift];
        
            \fill[white] (1*\spacing -  \shift,0) arc[start angle=0, end angle=180, radius=\spacing/2 - \shift];
                
        \foreach \x in {0,...,1}
            \filldraw[black] (\x*\spacing,0) circle (3pt) node[below] {$\O \B' \O^\top$};

        \foreach \x in {0}
            \draw[line width=\linewidth] (\x*\spacing,0) -- node[below] {$\G_{\A+\B}$} (\x*\spacing+\spacing,0);

        \draw[dashed] (\spacing + \shift,0 ) arc[start angle=0, end angle=180, radius=\spacing/2 + \shift];
        \draw[dashed] (\spacing -  \shift,0 ) arc[start angle=0, end angle=180, radius=\spacing/2 - \shift];
        
        \node[below] at (-1*\spacing,0) {};
        \node[below] at (3*\spacing,0) {};
    \end{tikzpicture}
    \hspace{-4.8cm} +  \hspace{-2.2cm}
    \begin{tikzpicture}[baseline=-0.65ex,scale=0.9]
        \def\spacing{3cm}
        \def\shift{0.1cm}
        \def
\linewidth{0.5mm} %
\foreach \x in {0,...,2}
\filldraw[black] (\x*\spacing,0) circle (3pt) node[below] {$\O \B' \O^\top$};
    
    \fill[gray!30] (2*\spacing + \shift,0) arc[start angle=0, end angle=180, radius=\spacing + \shift];

     \fill[white]  (2*\spacing -  \shift,0 ) arc[start angle=0, end angle=180, radius=\spacing/2 - \shift];
     \fill[white]  (\spacing -  \shift,0 ) arc[start angle=0, end angle=180, radius=\spacing/2 - \shift];
     
    \foreach \x in {0,...,1}
        \draw[line width=\linewidth] (\x*\spacing,0) -- node[below] {$\G_{\A+\B}$} (\x*\spacing+\spacing,0);

    \draw[dashed] (2*\spacing + \shift,0) arc[start angle=0, end angle=180, radius=\spacing + \shift];
    \draw[dashed] (2*\spacing -  \shift,0 ) arc[start angle=0, end angle=180, radius=\spacing/2 - \shift];
    \draw[dashed] (\spacing -  \shift,0 ) arc[start angle=0, end angle=180, radius=\spacing/2 - \shift];

    \node[below] at (-1*\spacing,0) {};
    \node[below] at (3*\spacing,0) {};
\end{tikzpicture} \hspace{-2.2cm} + \dots
\end{aligned}
\end{equation}
Note we are using $\G_{\A + \B}$ rather than $\G_{\A}$ to perform the contractions beneath each arc. Because of that, we don't need to include terms corresponding to configurations of ``arcs within arcs'', as they are already accounted for. That is, we don't need to explicitly include irreducible diagrams that aren't fully-connected. For example, the following contribution is already included for in the second term of Equation \eqref{eq:diagram_sum} above.
\begin{center}
\begin{tikzpicture}[scale=0.5]
    \def\spacing{3cm}
    \def\shift{0.1cm}
    \def\linewidth{0.5mm}

     \fill[gray!30] (3*\spacing +  \shift,0) arc[start angle=0, end angle=180, radius=3*\spacing/2 + \shift];
    \fill[white] (3*\spacing -  \shift,0) arc[start angle=0, end angle=180, radius=3*\spacing/2 - \shift];

    \fill[gray!30] (2*\spacing +  \shift,0) arc[start angle=0, end angle=180, radius=\spacing/2 + \shift];
    \fill[white] (2*\spacing -  \shift,0) arc[start angle=0, end angle=180, radius=\spacing/2 - \shift];
  
  \foreach \x in {0,...,3}
    \filldraw[black] (\x*\spacing,0) circle (3pt) node[below] {$\B$};

  \foreach \x in {0,...,2}
    \draw[line width=\linewidth] (\x*\spacing,0) -- node[below] {$\G_{\A}$} (\x*\spacing+\spacing,0);

  \draw[dashed] (3*\spacing + \shift,0) arc[start angle=0, end angle=180, radius=3/2*\spacing + \shift];
  \draw[dashed] (3*\spacing - \shift,0) arc[start angle=0, end angle=180, radius=3/2*\spacing - \shift];
  \draw[dashed] (2*\spacing +  \shift,0 ) arc[start angle=0, end angle=180, radius=\spacing/2 + \shift];
  \draw[dashed] (2*\spacing -  \shift,0 ) arc[start angle=0, end angle=180, radius=\spacing/2 - \shift];

  \node[below] at (-1*\spacing,0) {};
  \node[below] at (3*\spacing,0) {};
\end{tikzpicture}
\end{center}
When we average over $\O$ in Equation \eqref{eq:diagram_sum}, all the appearances of $\G_{\A+\B}(z)$ will be traced over. 
Using Equation \eqref{eq:R_principles} together with the fact that $g_{\A+\B}$ concentrates over $\O$
we can write:
\begin{equation}
    \R \simeq \sum_{n=0}^\infty \kappa^{(n)}_{\B} g_{\A + \B}(z)^{n-1}  \mathbf I.
\end{equation}
We now define $R_{\B}$ by 
\begin{equation}
    R_{\B}(g) = \sum_{n=0}^\infty \kappa_{\B}^{(n)} g^{n-1}.
\end{equation}
We thus arrive at the desired subordination relation:
\begin{equation}
\begin{aligned}
    \mathbb E_{\O} \bm G_{\A + \B}(z) &\simeq  \G_{\A}(z - R_{\B}(g_{\A+\B}(z))).
\end{aligned}
\end{equation}
Taking a trace and setting $\A = \bm{0}$, we recover the definition of the $R$-transform given in Section \ref{sec:rmt}. As discussed at the start of Section \ref{sec:diagrams}, from this relation, one obtains the additivity of the $R$-transform. This further implies that
\begin{equation}
    \kappa_{\A + \B}^{(n)} \simeq \kappa_{\A}^{(n)} + \kappa_{\B}^{(n)}.
\end{equation}
This justifies the term ``free cumulants'' for the $\kappa^{(n)}_{\A}$. The free cumulants of a sum of two relatively free random matrices just add. This is analogous to how cumulants of independent random variables are additive in classical probability. 

\subsection{\texorpdfstring{$S$-Transform Subordination}{S-Transform Subordination}}\label{sec:S_diagrams}

The proof for the $S$-transform subordination relation 
\begin{align}
   \mathbb E_{\B} \bm T_{\A \B}(z) &\simeq \bm T_{\A} (z S_{\B}(t_{\A \B}(z))), 
\end{align}
listed as \eqref{eq:subordination2} in the main text, is very similar. Recall that we want to compute
\begin{align}
    \mathbb{E}_{\B} \bm{T}_{\A\B}(z) 
    = \mathbb{E}_{\B} \A\B(z-\A\B)^{-1}
    = \A \mathbb{E}_{\B} \B(z-\A\B)^{-1} .
\end{align}
for fixed $\A$. We again take $\B = \O \B' \O$ with $\A$, $\B'$ deterministic and perform the $\O$ average. This time, we expand in powers of $\B/z$: 
\begin{equation}
    \mathbb E_{\O} \bm T_{\A \O \B' \O^\top}(z) = \A \, \mathbb E_{\O} \left[ \frac{1}{z} \O \B' \O^\top  + \frac{1}{z^2} \O \B' \O^\top \A  \O \B' \O^\top  + \dots    \right].
\end{equation}
We again resum in terms of irreducible diagrams:
\begin{center}
\begin{tikzpicture}[scale=0.8]
    \def\spacing{3cm}
    \def\shift{0.1cm}
    \def\linewidth{0.5mm} %

    \fill[gray!40] (1*\spacing,0) arc[start angle=0, end angle=180, radius=\spacing/2] -- cycle;
   \fill[gray!40] (3*\spacing, 0) arc[start angle=0, end angle=180, radius=\spacing/2] -- cycle;
   
  \draw[dashed] (1*\spacing +  \shift,0 ) arc[start angle=0, end angle=180, radius=\spacing/2+\shift];

   \draw[dashed] (3*\spacing +  \shift,0 ) arc[start angle=0, end angle=180, radius=\spacing/2+\shift];
   
   \node at ($(0*\spacing,0)!0.5!(1*\spacing,0)+(0,\spacing/5)$) {$\R'$};
   \node at ($(2*\spacing,0)!0.5!(3*\spacing,0)+(0,\spacing/5)$) {$\R'$};
   
  \foreach \x in {0,...,2}
    \filldraw[black] (\x*\spacing,0) circle (3pt) node[below] {};

  \foreach \x in {-1,1}
    \draw[line width=\linewidth] (\x*\spacing,0) -- node[below] {${\A/z}$} (\x*\spacing+\spacing,0);
  \foreach \x in {0, 2}
    \draw[line width=\linewidth] (\x*\spacing,0) -- node[below] {} (\x*\spacing+\spacing,0);

  \node[below] at (-1*\spacing,0) {};
  \node[below] at (3*\spacing,0) {};
\end{tikzpicture}
\vspace{-.2cm}
\end{center}
As before, because of the outer orthogonal average, $\R'$ is proportional to the identity. Calling the constant of proportionality $S^{-1}$ gives:\footnote{It is because of historical convention that this is denoted by $S^{-1}$ rather than $S$.}
\begin{equation}
\begin{aligned}
    \mathbb E_{\O} \bm T_{\A \O \B' \O^\top}(z) &\simeq  \frac{1}{z} \A S^{-1} + \frac{1}{z^2} \A  S^{-1} \A  S^{-1}  + \dots  \\
    &=  \A (z S \mathbf I - \A )^{-1} = \bm T_{\A}(z S).
\end{aligned}
\end{equation}
It remains to evaluate $S$. Expanding $\R'$ give the following terms
\begin{equation}
\begin{aligned}
    S^{-1} \mathbf I = & \hspace{-1.5cm}
    \begin{tikzpicture}[baseline=-0.65ex, scale=0.8]
        \def\spacing{3cm}
        \def\shift{0.1cm}
        \def\linewidth{0.5mm}
        \foreach \x in {0}
            \filldraw[black] (\x*\spacing,0) circle (3pt) node[below] {$\O \B' \O^\top$};

        \draw[dashed] (0 ,0 ) arc[start angle=-90, end angle=270, radius=\spacing/4 ];

        \node[below] at (-1*\spacing,0) {};
        \node[below] at (3*\spacing,0) {};
    \end{tikzpicture}
      \hspace{-6.5cm} +  \hspace{-1.9cm}
    \begin{tikzpicture}[baseline=-0.65ex, scale=0.8]
        \def\spacing{3cm}
        \def\shift{0.1cm}
        \def\linewidth{0.5mm} 
         \fill[gray!30] (1*\spacing +  \shift,0) arc[start angle=0, end angle=180, radius=\spacing/2 + \shift];
        
            \fill[white] (1*\spacing -  \shift,0) arc[start angle=0, end angle=180, radius=\spacing/2 - \shift];
                
        \foreach \x in {0,...,1}
            \filldraw[black] (\x*\spacing,0) circle (3pt) node[below] {$\O \B' \O^\top$};

        \foreach \x in {0}
            \draw[line width=\linewidth] (\x*\spacing,0) -- node[below] {$\bm G$} (\x*\spacing+\spacing,0);

        \draw[dashed] (\spacing + \shift,0 ) arc[start angle=0, end angle=180, radius=\spacing/2 + \shift];
        \draw[dashed] (\spacing -  \shift,0 ) arc[start angle=0, end angle=180, radius=\spacing/2 - \shift];
        
        \node[below] at (-1*\spacing,0) {};
        \node[below] at (3*\spacing,0) {};
    \end{tikzpicture}
    \hspace{-4.2cm} +  \hspace{-2cm}
    \begin{tikzpicture}[baseline=-0.65ex,scale=0.8]
        \def\spacing{3cm}
        \def\shift{0.1cm}
        \def
\linewidth{0.5mm} %
\foreach \x in {0,...,2}
\filldraw[black] (\x*\spacing,0) circle (3pt) node[below] {$\O \B' \O^\top$};
    
    \fill[gray!30] (2*\spacing + \shift,0) arc[start angle=0, end angle=180, radius=\spacing + \shift];

     \fill[white]  (2*\spacing -  \shift,0 ) arc[start angle=0, end angle=180, radius=\spacing/2 - \shift];
     \fill[white]  (\spacing -  \shift,0 ) arc[start angle=0, end angle=180, radius=\spacing/2 - \shift];
     
    \foreach \x in {0,...,1}
        \draw[line width=\linewidth] (\x*\spacing,0) -- node[below] {$\bm G$} (\x*\spacing+\spacing,0);

    \draw[dashed] (2*\spacing + \shift,0) arc[start angle=0, end angle=180, radius=\spacing + \shift];
    \draw[dashed] (2*\spacing -  \shift,0 ) arc[start angle=0, end angle=180, radius=\spacing/2 - \shift];
    \draw[dashed] (\spacing -  \shift,0 ) arc[start angle=0, end angle=180, radius=\spacing/2 - \shift];

    \node[below] at (-1*\spacing,0) {};
    \node[below] at (3*\spacing,0) {};
\end{tikzpicture} \hspace{-1.7cm} + \dots
\end{aligned}
\end{equation}
Here, the lines beneath each arc are:
\begin{equation}
    \bm G  =  \frac{1}{z} \A +  \frac{1}{z^2} \A \O \B' \O^\top \A + \dots = \G_{\A \B'} \A =  \frac{1}{z} \left(\mathbf I + \bm T_{\A \B} \right) \A, 
\end{equation}
where we have related the resolvent $\G_{\A \B}$ to $\bm T_{\A \B}$ using Equation \eqref{eq:tg_rel}. 
As before, by Equation \eqref{eq:R_principles}, $S^{-1}$ is equal to:
\begin{equation}
    S^{-1} \simeq \sum_{n=0}^\infty \frac{\kappa^{(n)}_{\B}}{z^{n-1}} \tr \left[ (\mathbf I + \bm T_{\A \B} (z)) \A \right]^{n-1} \simeq \sum_{n=0}^\infty \kappa^{(n)}_{\B} [S t_{\A}(z S)]^{n-1} = R_{\B}(S t_{\A \B}(z)).
\end{equation}
Here we have used the $\A, \B$ are free of one another and that $t_{\A \B}$ concentrates. Defining $S_{\B}$ through the self-consistent equation $1/S_{\B}(t) = R_{\B}(S_{\B}(t) t)$ gives us the desired subordination relation:
\begin{equation}
    \mathbb E_{\O} \bm T_{\A \O \B' \O^\top}(z) \simeq \bm T_{\A}(z S_{\B}(t_{\A \B}(z))).
\end{equation}
As discussed at the start of Section \ref{sec:subordination}, from this relation, one obtains the multiplicative property of the $S$-transform. We note that, as $\A$ is fixed, this also implies 
\begin{align} \label{eqn:t_tsfm_sub_no_a}
    \mathbb E_{\O} \O \B' \O^\top (z - \A \O\B' \O^{\top})^{-1}  \simeq (z S_{\B}(t_{\A \B}(z))- \A)^{-1}.
\end{align}

\clearpage

\section{\texorpdfstring{$R$ and $S$ Transforms of Important Ensembles}{R and S Transforms of Important Ensembles}}\label{sec:S_examples}

In this Appendix we derive the $R$- and $S$-transforms of a variety of useful random matrix ensembles. None of the final results are novel, but to the best of our knowledge some of the derivations are new. In particular, we are not aware of previous works that note how the $S$-transform of a Wishart matrix can be bootstrapped from the $S$-transform of a projection matrix. 

For a random matrix $\A$ we will write $S_{\A}(t)$ as a function of the $t$-transform to connect to the standard literature. In future sections, the results will be much more clearly expressible in terms of the degrees of freedom $\df_1(\lambda) \equiv -t(-\lambda)$. There, we will have $S_{\A}(t) = -S(-\df_1)$.

\subsection{Wigner}\label{sec:wigner}

As their elements are Gaussian, the sum of two matrices $\M_1, \M_2$ taken from Wigner distributions of variance $\sigma_1^2, \sigma_2^2$ will be a Wigner matrix of variance $\sigma_1^2 + \sigma_2^2$. Because the $R$-transform is additive, we get that $R_{\M_i}(g)$ must be proportional to $\sigma_i^2$. Further, by writing $R_{\M} = \sigma^2 f(g)$ and noting that $\alpha \M$ is a Wigner matrix with variance $\alpha^2 \sigma^2$, the scaling property in Equation \eqref{eq:R_scaling} gives that $\alpha^2 \sigma^2 f(g) = \alpha \sigma^2 f(\alpha g)$ from which we conclude that $f(g)$ must be linear. 

\begin{figure}[t]
    \centering
    \includegraphics[scale=0.6]{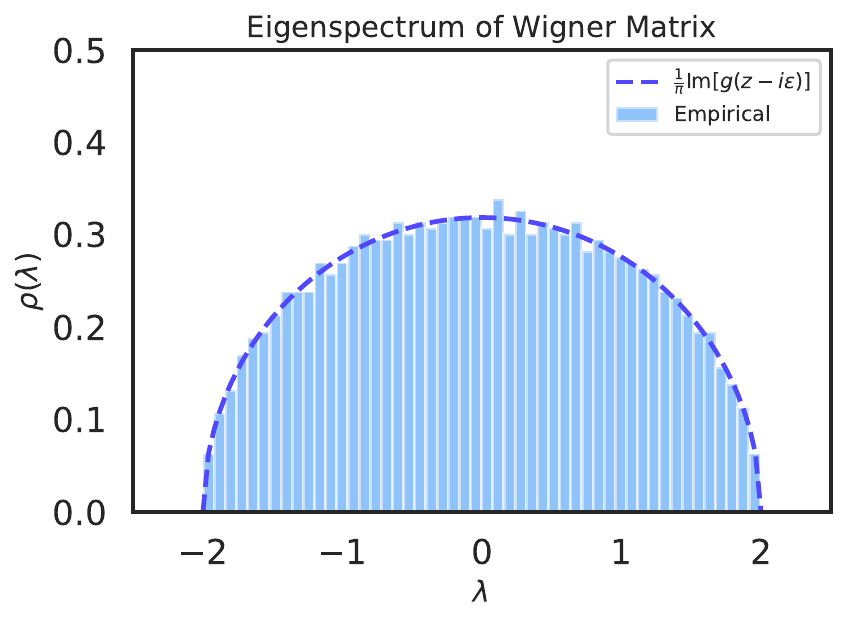}
    \caption{Empirical eigenspectrum of an $N \times N$ Wigner matrix when $N=2000$ (blue). Overlayed is the prediction of random matrix theory (dashed blue). }
    \label{fig:wigner}
\end{figure}

The constant can be fixed by considering the Laurent series expansion of $g_{\M}$:
\begin{equation}\label{eq:R_wig}
    g_{\M}(z) = \frac{1}{z} + \frac{1}{z^3} \tr[\M^2] + O(z^{-4}) = \frac{1}{z - \frac{\tr[\M^2]}{z}} + O(z^{-4}),
\end{equation}
which gives at leading order that $R_{\M} (g) = \tr[\M^2] g$. Because we've shown $R_{\M}$ is linear, this is exact. Using the fact that $\tr[\M^2] = \sigma^2$, we get that 
\begin{equation}
 R_{\M}(g) = \sigma^2 g   
\end{equation}
More generally, the above equality follows immediately from the fact that the $R$-transform is the free cumulant generating function and the only nonzero free cumulant of a Wigner matrix is its second.
As a consequence of Equation \eqref{eq:R_wig}, we get that
\begin{equation}
    g_{\M}(z) = \frac{1}{z-\sigma^2 g_{\M}(z)}.
\end{equation}
We can solve for $g$ as a function of $z$. We take the branch so that $g(z) \sim 1/z$ as $z \to \infty$ to obtain:
\begin{equation}
    g(z) = \frac{1}{2 \sigma^2} ( z - \sqrt{z^2 - 4 \sigma^2}),
\end{equation}
from which we can extract the density using equation \eqref{eq:inv_stiltjes}, yielding the famous \textbf{Wigner Semicircle Law}:
\begin{equation}
    \rho(\lambda) = \frac{\sqrt{4 \sigma^2 - \lambda^2}}{2 \pi \sigma^2}, \quad -2\sigma \leq \lambda \leq 2 \sigma.
\end{equation}
We illustrate the semicircle law in Fig. \ref{fig:wigner}. The Wigner distribution plays the role in free probability theory that the Gaussian distribution plays in ordinary probability theory: the spectral measure of properly normalized sums of free random matrices with independent and identically distributed elements converges to the Wigner distribution \cite{tao2012random}.

\subsection{Square Projections}

We consider symmetric square projection matrices $\P \in \mathbb R^{D \times D}$ onto $N$-dimensional subspaces of $\mathbb R^D$. $\P$ satisfies $\P = \P^2$. $\P$ thus has all eigenvalues either $0$ or $1$. We take $N$  out of $D$ eigenvalues to be unity and the rest to be zero. Defining the parameter $q = N/D$ we have that 
\begin{equation}
\begin{aligned}
    t_{\P}(z) = q t_{\mathbf I}(z) \Rightarrow \zeta_{\P}(t) &= \zeta_{\mathbf I}(t/q) = \frac{t/q}{t/q+1}. \\
\end{aligned}
\end{equation}
This directly yields the $S$-transform:
\begin{equation}\label{eq:S_square_proj}
    S_{\P}(t) = \frac{t+1}{t} \frac{t/q}{t/q+1} = \frac{t+1}{t+q}.
\end{equation}

\subsection{Rectangular Projections}

Often, one will encounter a projection matrix $\bm \Pi \in \mathbb R^{D \times N}$ mapping from $\mathbb R^D \to \mathbb R^N$. We call this a rectangular projection because $\bm \Pi$ is a rectangular matrix. Here, the directions in the null space are not included in the codomain of $\bm \Pi$. One can still calculate $S_{\bm \Pi^\top \A \bm \Pi}(t)$ in terms of $S_{\P * \A}(t) = S_{\P}(t) S_{\A}(t)$.

Let $\P \in \mathbb R^{D \times D}$ be  the square form of $\bm \Pi$. The trick is to relate $t_{\P * \A}(z)$ in $D$-dimensional space to $t_{\bm \Pi^\top \A \bm \Pi}(z)$ in $N$-dimensional space. Since we are keeping all the dimensions with nonzero eigenvalues, the unnormalized traces are the same, and we just need to account for the different normalizations. This means 
\begin{equation}
\begin{aligned}
    N t_{\bm \Pi^\top \A \bm \Pi}(z) &= D t_{\P * \A}(z) \\
    \Rightarrow  t_{\bm \Pi^\top \A \bm \Pi}(z) &= q^{-1} t_{\P * \A}(z)
    \\
    \Rightarrow \zeta_{\bm \Pi^\top \A \bm \Pi}(t) &= \zeta_{\P * \A}(q t).
\end{aligned}
\end{equation}
In terms of $S$-transforms, using Equation \eqref{eq:S_square_proj} this yields:
\begin{equation}\label{eq:rect_proj}
    S_{\bm \Pi^\top \A \bm \Pi}(t) = \frac{(t+1) q t}{t (q t + 1)} S_{\P}(q t) S_{\A}(q t) = S_{\A}(q t).
\end{equation}

\subsection{White Wishart}\label{sec:white_wish}

\begin{figure}
    \centering
    \includegraphics[scale=0.5]{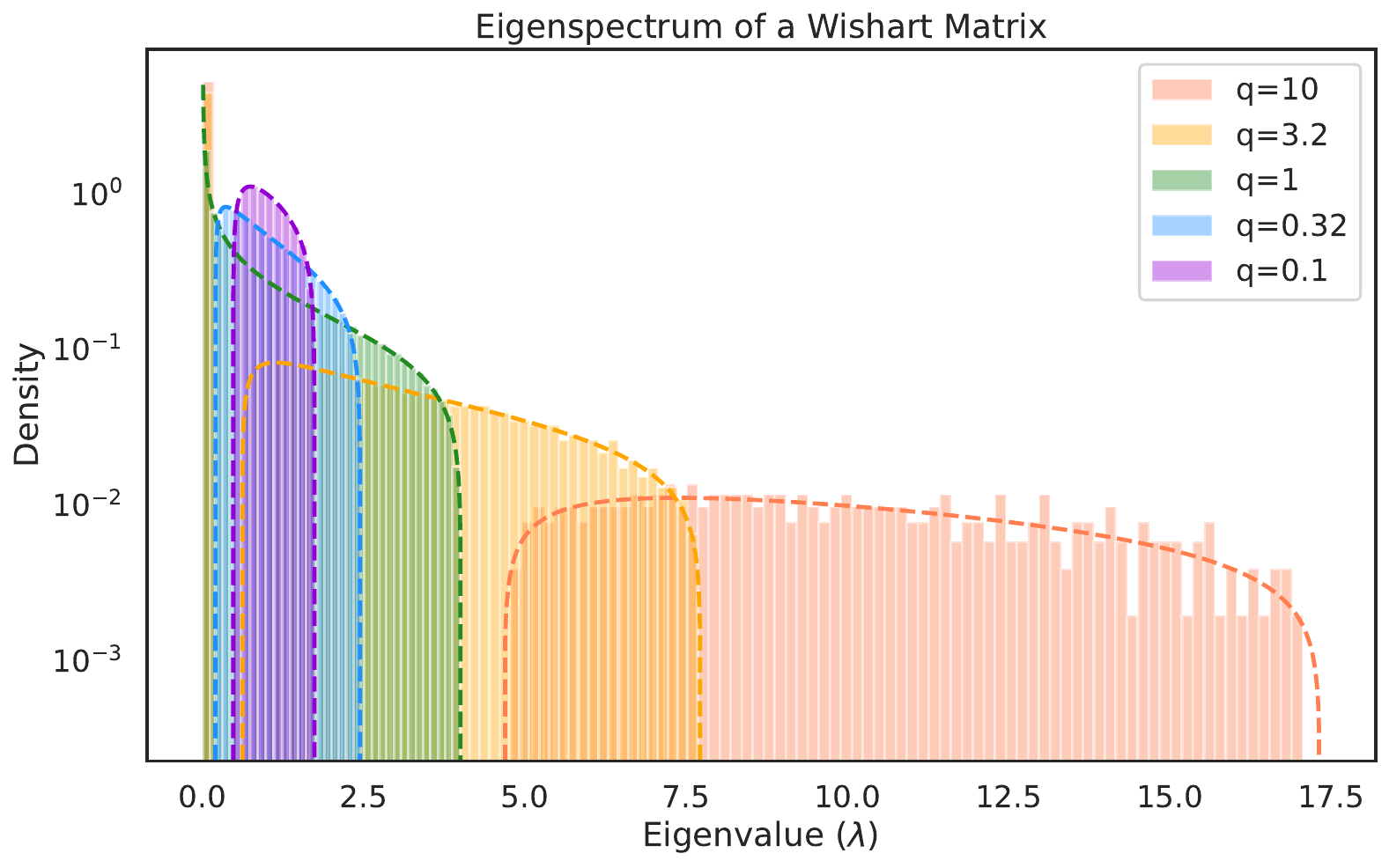}
    \caption{A series of empirical eigenspectra of unstructured Wishart matrices across different values of the overaparameterization ratio $q=D/P$. In all cases we have chosen $D = 1000$. The population covariance corresponds to a dirac delta function spike at $\lambda = 1$. The dashed lines are the prediction of random matrix theory, given by $\frac{1}{\pi} \mathrm{Im}\, g_{\W}(\lambda - i \epsilon)$ as $\epsilon \to 0$. We see as $q \to 0$ we get close to a delta function at $1$. For $q > 1$ we have some component that is a delta function at $0$ with weight $q-1$, separated from a bulk of eigenvalues. As $q$ increases above $q$, this gap grows. At $q = 1$ we have no gap. This is the key effect leading to double descent in linear regression, as was observed in \citet{advani2020high}. }
    \label{fig:white_wishart}
\end{figure}

The formula for the $S$-transform of a large Wishart matrix can be obtained by direct computation of $t_{\A}$, which is possible through a variety of methods (cavity, replica, etc). However, to demonstrate the manipulations that can be performed via the $S$-transform, we will derive this solely from knowing the $S$-transform of a projection as calculated in the preceding section.

In the large-$N$ limit, it will turn out that the spectral properties of a Wishart matrix depend only on the ratio of the number of dimensions to the number of data points. We therefore will view Wishart matrices as a one-parameter family of distributions. Concretely, for $\X \in \mathbb{R}^{P \times D}$ a data matrix with i.i.d. standard Gaussian entries, we therefore write $\W_{q} = \frac{1}{P} \X^{\top} \X$ for the corresponding empirical covariance, where $q = D/P$. 

Consider a $q=1$ Wishart matrix $\W_1 \in \mathbb R^{D \times D}$. The act of subsampling from $D$ down to $P$ points corresponds to taking a free product of $\W_{1}$ with $\frac{D}{P} \P$ where $\P \in \mathbb R^{D \times D}$ is a square projection onto a random $P$-dimensional space. Applying Equation \eqref{eq:S_square_proj} this for any ratio $D/P$ and using the fact that the resulting matrix has a Wishart distribution with $P$ degrees of freedom yields
\begin{equation}\label{eq:S_bootstrap1}
    S_{\W_{D/P}}(t) = S_{\frac{D}{P} \P} (t) S_{\W_1}(t) = \frac{1+t}{1 + \frac{D}{P} t} S_{\W_1}(t).
\end{equation}
Here, we have applied equation \eqref{eq:S_scaling} and recognized $\P$ as a projection with parameter $\frac{P}{D}$. 

In addition to subsampling, we can also project out features from $D$ to $N$. This involves mutliplying by a \textit{rectangular} projection with parameter $N/D$. Using equation \eqref{eq:rect_proj} we get:
\begin{equation}\label{eq:S_bootstrap2}
    S_{\W_{N/P}}(t) =  S_{\W_{D/P}}(t N/D) = \frac{1 + \frac{N}{D} t}{1 + \frac{N}{P} t} S_{\W_1}(t N / D).
\end{equation}
We now take $D$ much larger than $N, P$ so that $N / D \to 0$ and write $q = N/P$. By considering the $P \to \infty$ limit and noting that there, $\W_q \to \mathbf I$ and $S_{\mathbf I}(t) = 1$ we fix the normalization and obtain:
\begin{equation}\label{eq:Wishart_S}
    S_{\W_q}(t) = \frac{1}{1 + q t}.
\end{equation}
This is the most important $S$-transform for what follows. 

A consequence of this is that via equation \eqref{eq:RSinv}, we get
\begin{equation}
    R_{\W}(g) = \frac{1}{1- q g} \Rightarrow g_{\W}(z) = \frac{1}{z - \frac{1}{1-q g_{\W}(z)}}.
\end{equation}
This is a quadratic equation for $g_{\W}$, which can be solved exactly. Recalling that $g_{\W}(z)$ is the moment generating function in powers of $1/z$ and that $\tr[\W^0] = 1$, we must have $g_{\W}(z) \sim 1/z$ at large $z$. This fixes the root and yields:
\begin{equation}
    g_{\W}(z) = \frac{z + q - 1 - \sqrt{(z-\lambda_+) (z-\lambda_-)}}{2 q z}, \quad \lambda_{\pm} = (1 \pm \sqrt{q})^2.
\end{equation}
We can extract the spectrum using the equation \eqref{eq:inv_stiltjes}. This time we must be careful as $g_{\W}(z)$ has a pole with residue $q-1$ at $0$ if $q > 1$. This is due to $q>1$ Wishart matrices being non-invertible. We get:
\begin{equation}
    \rho(\lambda) = \frac{q-1}{q} \delta(\lambda) \bm 1_{q > 1}  + \frac{\sqrt{(\lambda_+ - \lambda) (\lambda - \lambda_-)} }{2\pi q \lambda} \bm{1}_{\lambda \in [\lambda_{-},\lambda_{+}]}.
\end{equation}
Here $\bm 1_{q > 1}$ is the indicator function that is $1$ when $q > 1$ and $0$ otherwise, and similarly $\bm{1}_{\lambda \in [\lambda_{-},\lambda_{+}]}$ is the indicator function that is $1$ when $\lambda \in [\lambda_{-},\lambda_{+}]$ and $0$ otherwise. The result is the well-known \textbf{Mar\caron{c}enko-Pastur} eigenvalue distribution \cite{marchenko1967distribution}. See Figure \ref{fig:white_wishart} for details. 

We note at small $q$ that this looks like a semicircle law of the identity matrix plus a Wigner matrix with entry noise having a standard deviation of $\sqrt{q}$. We noted that this is the leading order correction to covariance matrices in classical statistics in Section \ref{sec:RMT_examples} Example \ref{eg:Wigner}. 

We can also calculate the $S$-transform of the Gram matrix $\frac{1}{P} \X \X^\top \in \mathbb R^{P \times P}$ by recognizing it as $\frac{N}{P}$ times a Wishart with parameter $1/q$. Then using equation \eqref{eq:S_scaling}, we obtain another important $S$-transform:
\begin{equation}\label{eq:Gram_S}
    S_{\frac{1}{P} \X \X^\top}(t) = \frac{1}{q} \frac{1}{1 + t/q} = \frac{1}{q + t}.
\end{equation}

\subsection{Structured Wishart: Correlated Features} \label{sec:structured_wish_features}

We have considered the case where the features are not identically drawn from an isotropic distribution in Section \ref{sec:emp_covariance}, to motivate an application of the $S$-transform. Let us take $\Sh = \frac{1}{P} \X^\top \X$ where the rows of $\X$ are i.i.d. but drawn from a Gaussian with nontrivial covariance $\x_\mu \sim \mathcal N(\bm 0, \S)$. Having explicitly calculated the $S$-transform for a white Wishart matrix $\W$ with parameter $q = N/P$, we can now write:
\begin{equation}
    S_{\Sh}(t) = \frac{S_{\S}(t)}{1 + q t}.
\end{equation}
This lets us write 
\begin{equation}
\begin{aligned}
    &t_{\Sh}(z) = t_{\S}(\tilde z) \\
     \tilde z &= \frac{z}{1+ q t_{\S}(\tilde z)}.
\end{aligned}
\end{equation}
Given the spectrum of $\S$, this gives a self-consistent equation for $\tilde z$. We will use this equation (with $z = - \lambda, \tilde z = -\kappa, t = -\df_1$) very often in later sections.

\begin{figure}
    \centering
    \includegraphics[width=4in]{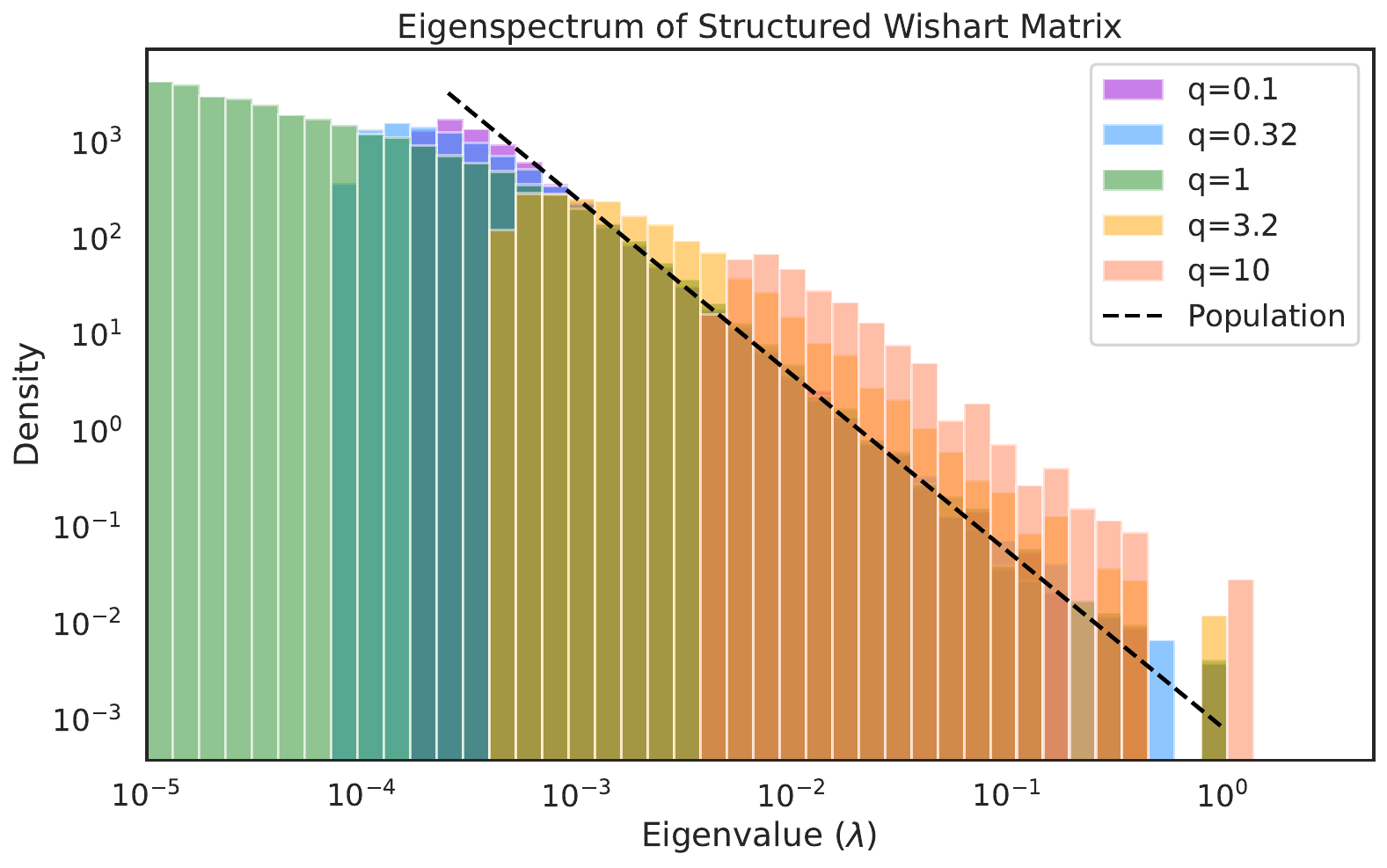}
    \caption{The eigenspectra of structured Wishart matrices as one varies the overparameterization parameter $q=D/P$. In all cases $D=1000$. The dashed black line is the eigenspectrum of the population covariance $\S$. Here $\S$ is chosen to have structure $\lambda_k = k^{-\alpha}$ for $k = \{1, \dots, D\}$ and $\alpha = 1.2$.}
    \label{fig:colored_wishart}
\end{figure}

\subsection{Structured Wishart: Correlated Samples} \label{sec:structured_wish_samples}

The converse problem is for the rows to be drawn from an isotropic (unstructured) Gaussian $x_\mu \sim \mathcal N(\bm 0, \mathbf I)$ but for different datapoints to be correlated. This corresponds to a matrix of the form
\begin{equation}
    \Sh = \frac{1}{P} \X^\top \K \X.
\end{equation}
We can calculate the $S$ transform of this as a rectangular projection with parameter $N/P$ of the free product $\K * \W_{1}$ where $\W_1 \in \mathbb R^{P \times P}$ is a white Wishart. This gives
\begin{equation}
    S_{\Sh}(t) = \frac{S_{\K}(q t)}{1 + q t} \Rightarrow \zeta_{\Sh}(t) = q (1 + t) \zeta_{\K}(q t)
\end{equation}
This implies that 
\begin{equation}
\begin{aligned}
    z &= \zeta_{\Sh}(t_{\Sh}(z)) = q (1+ t_{\Sh}(z)) \zeta_{\K}(q t_{\Sh}(z))\\
    &\Rightarrow q t_{\Sh}(z) = t_{\K}\left(\frac{z}{ q (1+ t_{\Sh}(z))} \right).
\end{aligned}
\end{equation}
Equivalently we can write this as
\begin{equation}\label{eq:S_gram}
    t_{\Sh}(z)  = q^{-1} t_{\K} (\tilde z), \quad \tilde z = \frac{z}{q + t_{\K}(\tilde z)}.
\end{equation}

\subsection{Structured Wishart: Correlated Features and Samples}\label{sec:full_structured_wish}

We now take the general case of a Wishart with correlations both between features and between samples. 
\begin{equation}
    \hat \S = \frac{1}{P} \S^{1/2} \X^\top \K \X \S^{1/2}.
\end{equation}
This gives us:
\begin{equation}\label{eq:S_full_structured_wish}
    S_{\hat \S}(t) = \frac{S_{\S}(t) S_{\K}(q t)}{1 + q t} \Rightarrow \zeta_{\Sh}(t) = q t \zeta_{\S}(t) \zeta_{\K}(q t).
\end{equation}
This implies that
\begin{equation}
\begin{aligned}
    z = \zeta_{\Sh}(t_{\Sh}(z)) = q t_{\Sh}(z) \, \zeta_{\S}(t_{\Sh}(z) ) \,\zeta_{\K}(q t_{\Sh}(z) )\\
    \Rightarrow t_{\Sh}(z) \simeq t_{\S} \left( \frac{z}{q t_{\Sh}(z) \zeta_{\K} (q t_{\Sh}(z)) } \right).
\end{aligned}
\end{equation}
Equivalently we can write this as:
\begin{equation}
    t = t_{\Sh}(z) \simeq t_{\S}(\tilde z), \quad \tilde z = \frac{z}{q t \zeta_{\K} (q t) }.
\end{equation}
This recovers the results obtained by \citet{burda2005moments}. 

\subsection{Shifted Wishart}\label{sec:shifted_wish}

Consider a white Wishart matrix $\W$ shifted by the identity, $\W + J \mathbf I$. Calculating the $S$-transform of this will be very helpful in the derivations that follow. 
One of the easiest ways to obtain this is to use equation \eqref{eq:SRinv} to relate the $S$ transform to the $R$ transform and then use equation \eqref{eq:Radd} to perform the shift. This gives:
\begin{equation}
    S_{\W + J \mathbf I}(t) = \frac{1}{R_{\bm W + J \mathbf I}(t S_{\W + J \mathbf I})} = \frac{1}{J + \frac{1}{1 - q t S_{\W + J \mathbf I}(t)}}.
\end{equation}
This can be solved exactly to give
\begin{equation}\label{eq:full_shifted_wish}
    S_{\W + J \mathbf I}(t) = \frac{2}{1 + J + q t + \sqrt{(1+ J + qt)^2 - 4 J q t}}.
\end{equation}
This is related to the generalization error of additively noised random features studied in Section \ref{sec:NLRF}. 
For our purposes in Section \ref{sec:LR}, we will only care about the leading order behavior in $J$, which can be written as:
\begin{equation}\label{eq:partial_shifted_wish}
    S_{\W + J \mathbf I}(t) = \frac{1}{1 + q t + \frac{J}{1 + q t}} + O(J^2)
\end{equation}

\subsection{Deep White Wishart Product} \label{sec:deep_white_wish}
\begin{figure}
    \centering
    \includegraphics[width=4in]{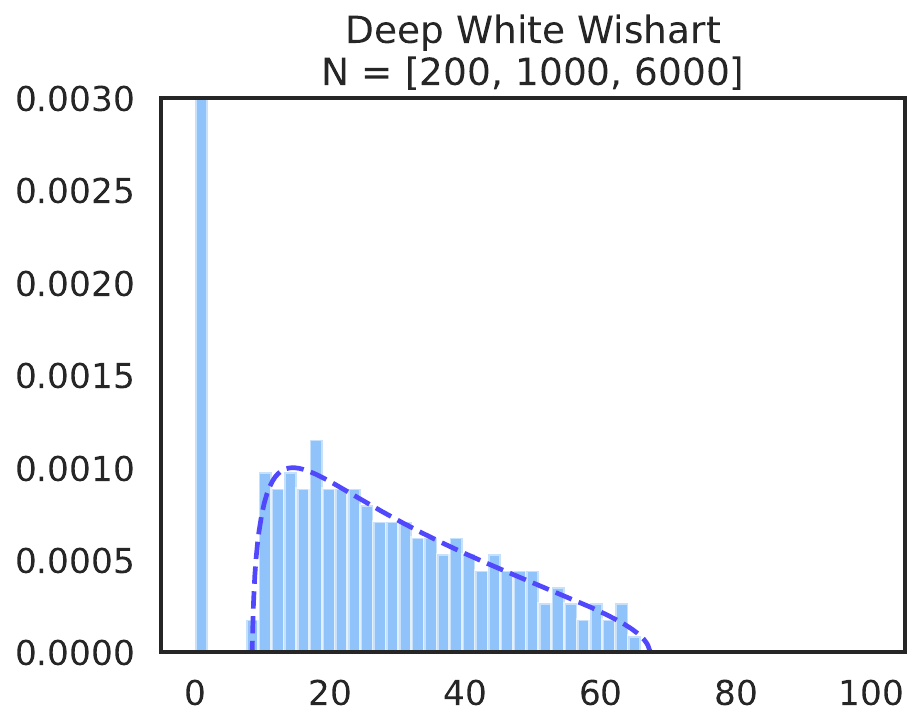}
    \includegraphics[width=4in]{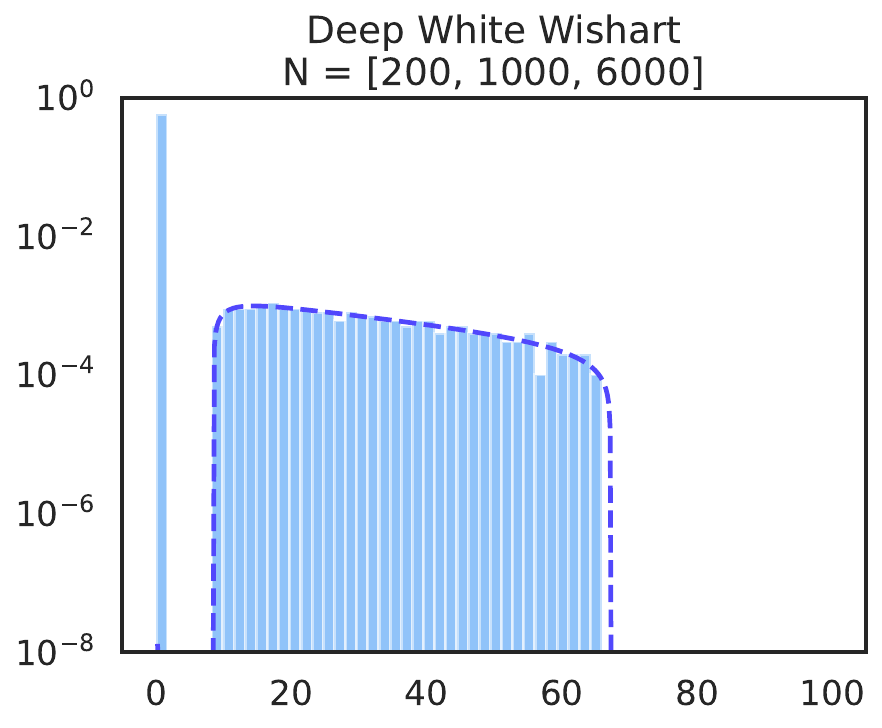}
    \caption{The eigenspectrum of a depth-2 Wishart product $\frac{1}{N_0 N_1} \X_{2}^{\top} \X_{1}^{\top} \X_{1} \X_{2}$, where $\X_{\ell} \in \mathbb{R}^{N_{\ell-1} \times N_{\ell}}$. Here, $N_{0} = 200$, $N_{1} = 1000$, and $N_{2} = 6000$, as indicated by the list of dimensions in the title of each panel. The dashed solid lines are given by the predictions of \eqref{eq:deep_white_wish} The left panel is linearly spaced on the y-axis while the right is logarithmically spaced. }
    \label{fig:deep_white_wishart}
\end{figure}

Consider a series of white Wishart matrices $\W_\ell = \frac{1}{N_{\ell-1}} \X_\ell^\top \X_\ell$ with $\X_\ell \in \mathbb R^{N_{\ell-1} \times N_{\ell}}$ having rows drawn i.i.d. from $\mathcal N(\bm 0, \mathbf I)$.
Consider the following matrix product, which we will call a deep Wishart product:
\begin{equation}
    \C_L = \frac{\X_L^\top \cdots \X_1^\top \X_1 \cdots \X_L}{N_0 \dots N_{L-1}}.
\end{equation}
By Equations \eqref{eq:Wishart_S} and \eqref{eq:Gram_S}, we have
\begin{equation}
\begin{aligned}
    S_{\frac{1}{N_{\ell-1}}\X_\ell^\top \X_\ell}(t) &= \frac{1}{1+\frac{N_\ell}{N_{\ell-1}} t},\\
    S_{\frac{1}{N_{\ell-1}} \X_\ell \X_\ell^\top}(t) &= \frac{1}{\frac{N_\ell}{N_{\ell-1}} + t}.
\end{aligned}
\end{equation}
At each step we look first at the free product
\begin{equation}
\begin{aligned}
    \tilde {\C}_\ell &\equiv \C_{\ell -1}* \left(\frac{1}{N_{\ell-1}} \X_\ell \X_\ell^\top\right) \in \mathbb R^{N_{\ell-1} \times N_{\ell-1}}\\
    \Rightarrow & S_{\tilde \C_\ell}(t) =  S_{\C_{\ell-1}}(t) \frac{1}{\frac{N_\ell}{N_{\ell-1}} + t}.
\end{aligned}
\end{equation}
Again, because the nonzero spectra of these matrices agree, their unnormalized traces are equal. Accounting for the different normalizations,  we have $t_{\C_\ell} = \frac{N_{\ell-1}}{N_\ell} t_{\tilde \C_\ell} \Rightarrow \zeta_{\C_\ell}(t) = \zeta_{\tilde \C_\ell}(t N_\ell / N_{\ell -1 })$. That means 
\begin{equation}
\begin{aligned}
    S_{\C_\ell}(t) &= \frac{t+1}{t+\frac{N_{\ell -1 }}{N_\ell}} S_{\tilde \C_\ell}(t N_\ell / N_{\ell -1 }) \\
    &= \frac{1}{1 + \frac{N_\ell}{N_{\ell-1}} t} S_{\C_{\ell-1}}(t N_\ell/N_{\ell - 1}).
\end{aligned}
\end{equation}
Expanding this full product recursively yields:
\begin{equation}\label{eq:deep_white_wish}
    S_{\C_L}(t) = \prod_{\ell=0}^{L-1} \frac{1}{1 + \frac{N_L}{N_{\ell}} t},
\end{equation}
consistent with the self-consistent equation derived in previous works \cite{muller2002asymptotic,zavatone2023replica,burda2010singular}. As shown in Figure \ref{fig:deep_white_wishart}, numerical solution of the resulting self-consistent equation yields an excellent match to numerical experiment.

One can apply the same recursive argument to the Gram matrices:
\begin{equation}
\begin{aligned}
    \K_L &= \frac{\X_1 \cdots \X_L \X_L^\top \cdots \X_1^\top}{N_0 \cdots N_{L-1}}.
\end{aligned}
\end{equation}
This yields:
\begin{equation}\label{eq:deep_white_gram}
    S_{\K_L}(t) = \prod_{\ell=1}^{L} \frac{1}{\frac{N_\ell}{N_0} +t}.
\end{equation}

\subsection{Deep Structured Wishart Product}\label{sec:deep_structured_wish}

Now, let us allow for arbitrary structure in the features of each Wishart matrix in the deep product. We write $\frac{1}{N_{\ell-1}} \M_\ell^\top \M_\ell = \W_\ell * \bm \Sigma_\ell$ for $\W_\ell$ a white Wishart and $\S_\ell$ the population covariance of the $\ell$-th Wishart matrix. We then get
\begin{equation}
    S_{\tilde \C_\ell}(t) =  S_{\C_{\ell-1}}(t) \frac{1+t}{\frac{N_\ell}{N_{\ell-1}}  + t} S_{\W_\ell}(t N_{\ell -1}/N_\ell) S_{\bm \Sigma_\ell}(t N_{\ell -1}/N_\ell)
\end{equation}
\begin{equation}
\begin{aligned}
    \Rightarrow S_{\C_\ell}(t) &= \frac{1+t}{\frac{N_{\ell-1}}{N_\ell}  + t} S_{\tilde \C_\ell}(t N_{\ell}/N_{\ell-1})\\
    &= \frac{1+t}{\frac{N_{\ell-1}}{N_\ell}  + t} \frac{1 + t \frac{N_\ell}{N_{\ell-1}}}{\frac{N_\ell}{N_{\ell-1}} + t \frac{N_\ell}{N_{\ell-1}}} S_{\bm W_\ell} (t) S_{\bm \Sigma_\ell} (t)  S_{\C_{\ell-1}} (t N_\ell/N_{\ell-1}) \\
    &= \frac{S_{\bm \Sigma_\ell}(t)}{1+\frac{N_{\ell}}{N_{\ell -1}} t}  S_{\C_{\ell-1}} (t N_\ell/N_{\ell-1}) .
\end{aligned}
\end{equation}
Expanding this recursively gives 
\begin{equation}
    S_{\C_L}(t) = \prod_{\ell=0}^{L-1} \frac{S_{\bm \Sigma_\ell}(\frac{N_{L}}{N_\ell} t)}{1 + \frac{N_L}{N_{\ell}} t}.
\end{equation}
In terms of the inverse functions $\zeta_{\bm \Sigma_\ell}$ we get:
\begin{equation}\label{eq:deep_structured_wish}
    \frac{1}{S_{\C_L}(t)} = \prod_{\ell=0}^{L-1} \frac{N_L}{N_\ell} t \zeta_{\bm \Sigma_\ell}\left(\frac{N_L}{N_\ell} t \right). 
\end{equation}
Similarly one gets:
\begin{equation}\label{eq:deep_structured_gram}
    \frac{1}{S_{\K_L}(t)} = \prod_{\ell=1}^{L} t \zeta_{\bm \Sigma_\ell}\left(\frac{N_0}{N_L} t \right). 
\end{equation}
This is consistent with the self-consistent equation derived in \citet{zavatone2023replica}, where it was shown that the resulting prediction for the spectral density gives good matches with numerical experiment. It is interesting to note that the order parameters in the replica computation of \citet{zavatone2023replica} correspond precisely to the $S$-transforms of partial products including only the first $\ell$ factors. 

\clearpage

\bibliography{references}

\end{document}